%% file: main.tex
\documentclass[preprint,3p]{elsarticle}

\usepackage[T1]{fontenc}
\usepackage[utf8]{inputenc}
\usepackage[english]{babel}

\usepackage{amsmath}
\usepackage{amssymb}
\usepackage{amsfonts}
\usepackage{bbm}
\usepackage{dsfont}

\usepackage{algorithm}
\usepackage{algpseudocode}

\usepackage{graphicx}
\usepackage{tikz}
\usepackage{pgfplots}
\usepackage{pgfplotstable}
\usepackage{subcaption}
\usepackage{caption}
\usetikzlibrary{
  positioning,
  arrows,
  arrows.meta,
  patterns,
  shapes.geometric,
  fit,
  matrix
}
\usepgfplotslibrary{fillbetween}
\pgfplotsset{compat=1.18} 

\usepackage{booktabs}
\usepackage[table]{xcolor}
\usepackage{multirow}
\usepackage{makecell}
\usepackage{longtable}
\usepackage{adjustbox}

\usepackage{modular}
\usepackage{listings}
\usepackage{titlesec}
\usepackage{zref-abspage}
\usepackage{csquotes}

\usepackage{siunitx}

\usepackage{hyperref}
\usepackage{cleveref}

\usepackage{pdfpages}

\DeclareMathAlphabet{\mathcal}{OMS}{cmsy}{m}{n}

\input{commands}
\input{symbols}

\newcommand{\new}[1]{#1}

\journal{Information Sciences}

\begin{document}

\begin{frontmatter}



\title{Benchmarking noisy label detection methods}


\author[ufsc]{Henrique Pickler}
\author[ufsc]{Jorge K. S. Kamassury} 
\author[ufsc]{Danilo Silva}

\affiliation[ufsc]{organization={Universidade Federal de Santa Catarina},
            addressline={R. Eng. Agronômico Andrei Cristian Ferreira},
            city={Florianópolis},
            postcode={88040-900}, 
            state={Santa Catarina},
            country={Brazil}}

\begin{abstract}
Label noise is a common problem in real-world datasets, affecting both model training and validation. Clean data are essential for achieving strong performance and ensuring reliable evaluation. While various techniques have been proposed to detect noisy labels (or label errors), there is no clear consensus on optimal approaches. We perform a comprehensive benchmark of detection methods by decomposing them into three fundamental components: gathering strategy (in-sample vs out-of-sample), label disagreement measure, and aggregation method. This decomposition can be applied to many existing detection methods, and enables systematic comparison across diverse approaches. To fairly compare methods, we propose a unified benchmark task: detecting a fraction of training samples equal to the dataset's noise rate. We also introduce a novel metric: the false negative rate at this fixed operating point. Our evaluation spans vision and tabular datasets under both synthetic and real-world noise conditions. We identify that in-sample gathering using average probability aggregation combined with the logit margin as the label disagreement measure achieves the best results across most scenarios. Our findings provide practical guidance for designing new detection methods and selecting techniques for specific applications.
\end{abstract}




\end{frontmatter}


\section{Introduction}

\input{01_introduction}
\input{02_background}
\input{03_detecting}
\input{04_methods}
\input{05_results}

\input{06_conclusion}

\appendix

\input{09_appendix}

\bibliographystyle{elsarticle-num} 
\bibliography{references}
\end{document}

%% file: commands.tex
\newcommand{\plotwithconfidenceinterval}[5][]{%
    \addplot+[mark=none, #1] table [x=#3, y=#4, col sep=comma] {#2};
    
    \pgfplotsset{cycle list shift=-1} 
    \addplot+[name path=upper,draw=none, mark=none, forget plot] table[x=#3, y expr=\thisrow{#4}+\thisrow{#5}, col sep=comma] {#2};
    \addplot+[name path=lower,draw=none, mark=none, forget plot] table[x=#3, y expr=\thisrow{#4}-\thisrow{#5}, col sep=comma] {#2};

    \addplot+[fill opacity=0.3, mark=none, #1, forget plot] fill between[of=upper and lower];
    \pgfplotsset{cycle list shift=0} 
}

\newcommand{\stdsize}{\footnotesize}

\makeatletter
\newcommand{\autopagecounter}{%
  \zref@labelbyprops{autopage}{abspage}%
  \setcounter{page}{\zref@extract{autopage}{abspage}}
}
\makeatother

\newcommand{\comment}[1]{}

%% file: symbols.tex
\DeclareMathOperator*{\argmax}{arg\,max}

\usepackage{dsfont}

\newcommand{\RR}{\mathbb{R}}

\newcommand{\calC}{\mathcal{C}}

\newcommand{\calN}{\mathcal{N}}

\newcommand{\vectorindex}[1]{_{#1}}
\newcommand{\vi}[1]{\vectorindex{#1}}

\newcommand{\realnumbers}{\mathbb{R}}
\newcommand{\softmax}{\text{softmax}}
\newcommand{\indicatorfunction}{\mathbbm{1}}

\newcommand{\probabilityop}{P}
\newcommand{\featurespace}{\mathcal{X}}
\newcommand{\samplefeature}{x}
\newcommand{\labelspace}{\mathcal{Y}}
\newcommand{\samplelabel}{y}
\newcommand{\samplelabelnoisy}{\tilde{y}}
\newcommand{\samplepair}[1]{(\samplefeature#1, \samplelabel#1)}
\newcommand{\samplepairnoisy}[1]{(\samplefeature#1, \samplelabelnoisy#1)}
\newcommand{\modelparameters}{\theta}
\newcommand{\modelfunction}{f}
\newcommand{\classifier}{h}
\newcommand{\lossfunction}{\ell}
\newcommand{\agreementf}{L}
\newcommand{\empiricalrisk}{\mathcal{R}}
\newcommand{\modellogits}{z}
\newcommand{\modelprobas}{\hat{p}}

\newcommand{\epochindex}[1]{^{[#1]}}
\newcommand{\sampleonehotlabel}{e}
\newcommand{\sampleonehotlabelnoisy}{\tilde{\sampleonehotlabel}}

\newcommand{\detectionfunction}{h_{\text{NL}}}
\newcommand{\scorefunction}{f_{\text{NL}}}

\newcommand{\cleanjoint}{\probabilityop}
\newcommand{\dataset}{\mathcal{D}}
\newcommand{\noisydataset}{\tilde{\dataset}}
\newcommand{\cleansubset}{\calC}
\newcommand{\noisysubset}{\calN}
\newcommand{\noiserate}{\eta}

\newcommand{\residualnoise}{\noiserate_{\text{res}}}
\newcommand{\normalizedresidualnoiserate}{\rho}
\newcommand{\remainingfnr}{\fnr}
\newcommand{\perfectfnr}{\fnr_\text{ideal}}

\newcommand{\numberofclasses}{K}
\newcommand{\numberofsamples}{N}
\newcommand{\epochs}{E}
\newcommand{\foreachsample}[1]{#1 \in \{1,\, \ldots \,, \numberofsamples \}}

\newcommand{\truepositive}{\text{TP}}

\newcommand{\falsenegative}{\text{FN}}
\newcommand{\fnr}{\text{FNR}}

\newcommand{\budget}{b}
\newcommand{\predictedasnoisy}{\hat{\noisysubset}}
\newcommand{\noisyscore}{S}
\newcommand{\detectionparameters}{\Phi}
\newcommand{\detectionthreshold}{\tau}

\newcommand{\confidentjoint}{C}

%% file: 01_introduction.tex
Most supervised learning methods assume a perfectly labeled dataset. However, training data often contain incorrectly labeled instances. Even large, standard benchmark datasets, such as CIFAR, ImageNet, and MS-COCO, are known to have noisy labels (interchangeably referred to as label errors) \cite{northcutt_confident_2021, northcutt_pervasive_2021}. Label noise can originate from various sources, including human error, inherent data ambiguity, and automated labeling systems \cite{frenay_survey}.

During training, label noise can significantly degrade model performance \cite{frenay_survey}, as models tend to memorize incorrect labels \cite{zhang_understanding}. To address this challenge, various methods have been proposed in the literature, including modifications to model architectures, the development of robust regularization techniques, and the design of noise-robust loss functions \cite{song_survey}. One interesting strategy is to identify mislabeled samples and leverage this information for training. For instance, removing samples identified with label noise can be advantageous even under imperfect detection \cite{huang_o2u-net_2019, han_co-teaching_nodate, northcutt_confident_2021}, as their detrimental effect often outweighs the benefits of having more data. Another approach is to avoid the labels of identified mislabeled samples by using semi-supervised techniques \cite{li_dividemix_2020, karim_unicon_2022, arazo_unsupervised_2019}, or self-supervised techniques \cite{tan_co-learning_2021}, mitigating the negative effect of label noise while maintaining some of the benefits of having more training data. A third approach involves (manually or automatically) relabeling these samples in order to have a cleaner dataset \cite{chen_clean_2022}, enabling further training without using specific methods designed for label errors. 

During evaluation, the presence of label noise in a validation or test dataset can lead to misleading conclusions. If the noise is biased, for instance, model assessments based on such data may inherit and reflect that bias. In this context, label error detection can be used to alleviate this issue, by identifying and helping correct mislabeled samples in the validation or test set.

While it is common to apply filtering (removing samples), semi-supervised learning (treating identified mislabeled samples as unlabeled), or relabeling (either automatically or manually) to achieve robust training and evaluation, an important observation is that the detection stage is an independent step that enables these strategies. As such, improving the criteria used for detection can enhance the overall effectiveness of these methods. 

Be it noise in the training, validation or test set, detecting mislabeled samples is an important task to enable working with noisy datasets. Despite its importance in both robust training and robust evaluation, the detection task itself often receives limited attention as the focus tends to be on robust training outcomes, where the classification performance is generally the main metric observed \cite{northcutt_confident_2021, karim_unicon_2022, zhao_centrality_2022, huang_o2u-net_2019, yue_ctrl_2023, pleiss_identifying_nodate, chen_clean_2022}. While some methods briefly compare their proposed detection methods with others \cite{huang_o2u-net_2019, pleiss_identifying_nodate, yue_ctrl_2023}, they lack strong baselines, often rely on multiple metrics to evaluate methods on different operating points, or do not use the appropriate detection procedure associated with the methods being compared.

\begin{figure}[h]
    \centering
    \input{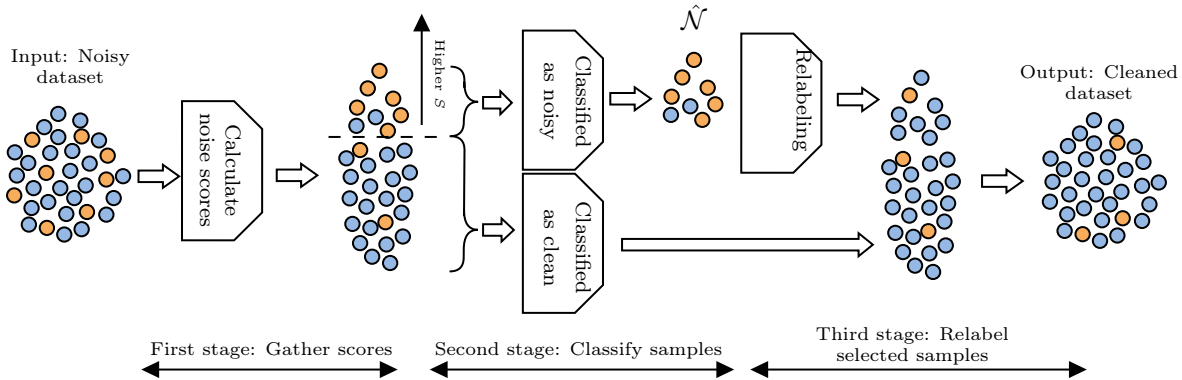}
    \caption{Illustration of the dataset cleaning process: Starting from a noisy dataset, scores are calculated to identify potentially mislabeled samples. Samples are detected as noisy or clean based on a threshold, with mislabeled samples relabeled to produce a cleaned dataset for downstream tasks.}
    \label{fig:cleaning_diagram}
\end{figure}


\new{To address this gap, we perform a comprehensive benchmark comparing noisy label detection methods. Figure \ref{fig:cleaning_diagram} shows the data cleaning process adopted in our benchmark. While detection can be paired with various approaches to treat detected mislabeled samples (as discussed previously, sample removal, label removal or label correction), we assume perfect relabeling in order to remove confounding factors and isolate detection performance. This naturally leads to evaluating detection at a fixed budget, i.e., detecting a fixed fraction of the dataset. Under this assumption, the quality of the cleaned dataset depends solely on which samples are selected for relabeling, ensuring a fair comparison across methods that would otherwise operate at different thresholds. The main contributions of this paper are summarized as follows:}
\begin{itemize}
    \item \textbf{A comprehensive benchmark for noisy label detection}: We introduce a unified evaluation framework where noise detection is performed without knowledge of which labels are corrupted and detection is evaluated as a binary classification task using a novel metric.

    \item \textbf{A modular taxonomy of detection methods}: We decompose methods into three dimensions: gathering strategy, label disagreement measure, and aggregation method. This unifies prior approaches under a common framework and enables the systematic derivation of eight new detection methods.
    
    \new{\item \textbf{No single method dominates, but one of them is consistently strong}: Detection performance varies with noise type, number of classes, model architecture, and training duration, and no single method achieves the best results in all scenarios. Nevertheless, our proposed combination of Logit Margin disagreement measure \cite{pleiss_identifying_nodate} with Mean Probabilities aggregation (over all training epochs) consistently ranks near the top across all evaluated settings, emerging as the most reliable choice when prior knowledge about the noise structure is unavailable.

    \item \textbf{In-sample vs.\ out-of-sample gathering}: We find that out-of-sample gathering does not consistently improve detection performance; in particular, for most methods beyond Confident Learning, in-sample strategies are preferable.}
\end{itemize}

%% file: 02_background.tex
\section{Learning with Noisy Labels}

Although this work focuses on detecting noisy labels, many detection methods are developed in the context of robust training and are often integrated into such approaches. Given this connection, we first formalize the problem of learning with label noise, which is also useful to establish some basic notation.

\label{sec:learning_problem_statement}
Consider a $\numberofclasses$-class classification task, with feature space denoted by $\featurespace$ and label space by $\labelspace = \{1, \dots, \numberofclasses\}$. Let $\dataset = \{\samplepair{\vectorindex{i}}\}^\numberofsamples_{i=1}$ be a set of examples, where the pairs $\samplepair{\vectorindex{i}} \in \featurespace \times \labelspace$ are independent random variables drawn from an unknown joint distribution $\cleanjoint$. In the canonical supervised learning setting, the objective is to learn a function $\classifier_\modelparameters: \featurespace \to \labelspace$, parameterized by $\modelparameters$, that correctly reproduces the mapping of the feature space $\featurespace$ to the label space $\labelspace$ given by $\cleanjoint$. 
\new{We assume the classifier $\classifier_\modelparameters$ can be decomposed as $\classifier_\modelparameters(x) = \argmax_{k \in \labelspace} z_k$, where $\modellogits = (\modellogits_1,\ldots,\modellogits_K) = \modelfunction_\modelparameters(x)$ and $\modelfunction_\modelparameters: \featurespace \to \RR^K$ is a scoring function. The output of $\modelfunction_\modelparameters$, the vector $\modellogits$, is referred to as the vector of \textit{logits}. It is understood that, for $k=1,\ldots,K$, an estimate $\modelprobas_k$ of the posterior probability $P(y=k|x)$ can be obtained from the logits as
\begin{equation}
\modelprobas = (\modelprobas_1,\ldots,\modelprobas_K) = \softmax(\modellogits)
\end{equation}
where
\begin{equation}
\softmax(\modellogits)_k = \frac{\exp(\modellogits_k)}{\sum_{j=1}^K\exp(\modellogits_j)}.
\end{equation}}

To find $\modelfunction_\modelparameters$, usually an optimization of the parameters $\modelparameters$ is performed to minimize the empirical risk
\begin{align}
    \empiricalrisk_{\dataset}(\modelfunction_\modelparameters)
    &= \frac{1}{|\dataset|} \sum_{i=1}^{|\dataset|} \lossfunction(\modelfunction_\modelparameters(\samplefeature\vectorindex{i}), \samplelabel\vectorindex{i})
\end{align}
where $\lossfunction: \RR^K \times \labelspace \to \RR$ is an appropriate loss function.

In the context of label noise learning, however, we observe the set $\noisydataset = \{\samplepairnoisy{\vectorindex{i}}\}^\numberofsamples_{i=1}$, where the labels $\samplelabelnoisy\vectorindex{i} \in \labelspace$ are drawn from a conditional distribution $\probabilityop(\samplelabelnoisy | \samplefeature, \samplelabel)$, referred to as the noise transition distribution. This means that the risk being minimized is actually
\begin{align}
    \empiricalrisk_{\noisydataset}(\modelfunction_\modelparameters) 
    &= \frac{1}{|\noisydataset|} \sum_{i=1}^{|\noisydataset|} \lossfunction(\modelfunction_\modelparameters(\samplefeature\vectorindex{i}), \samplelabelnoisy\vectorindex{i}).
\end{align}
The objective, then, is to learn a good model despite only having access to the noisy empirical risk. This can be a hard task, and multiple different methods have been proposed to train robust models in these conditions.

%% file: 03_detecting.tex
\section{Detecting label noise}

While much of the literature on learning with label noise focuses on training robust models in the presence of corrupted labels, a complementary and increasingly relevant line of work aims to identify which samples are likely mislabeled. The ability to detect noisy labels can enable more effective training strategies, such as sample selection, relabeling, semi-supervised or self-supervised robust training methods. In this work, we focus on the detection problem, which underpins the success of many noise-robust training methods.

\subsection{Problem statement}

To formalize the detection task, recall the definitions from Section \ref{sec:learning_problem_statement}. We assume that the observed dataset $\noisydataset$, which is the dataset we intend to perform detection on, can be partitioned into two disjoint subsets: a clean set $\cleansubset = \{\samplepairnoisy{\vectorindex{i}} : \samplelabelnoisy\vectorindex{i} = \samplelabel\vectorindex{i}, \foreachsample{i}\}$ and a noisy set 
$\noisysubset = \noisydataset \setminus \cleansubset$. We define the detection ground truth as the vector that indicates whether a sample from $\noisydataset$ belongs to $\noisysubset$ or not. This indication, however, is unknown in practice, and our goal is to infer it. That is, a detection method is a function that predicts whether each sample from $\noisydataset$ belongs to $\noisysubset$, represented by
\begin{equation}
    \detectionfunction : (\featurespace \times \labelspace)^N \to \{0, 1\}^N,
\end{equation}
where $\detectionfunction(\noisydataset)\vectorindex{i} = 1$ indicates that the $i$th sample $\samplepairnoisy{\vectorindex{i}}$ is detected as noisy.\footnote{\new{In principle, we could have defined $\detectionfunction$ as a function $\detectionfunction(\cdot; \noisydataset): \featurespace \times \labelspace \to \{0,1\}$, corresponding to a conventional binary classifier. However, this definition is too restrictive, as it requires $\detectionfunction(x,\tilde{y}; \noisydataset)$ to be applicable to \textit{any} pair $(x,y) \in \featurespace \times \labelspace$, which is unnecessary for noisy label detection on $\noisydataset$.}}
\new{
Without loss of generality, we assume $\detectionfunction$ can be decomposed as 
\begin{equation}
    \detectionfunction(\noisydataset)_i = \indicatorfunction\left( \noisyscore_i \geq \detectionthreshold \right), \quad i=1,\ldots,N
\end{equation}
where $\noisyscore = (\noisyscore_1,\ldots,\noisyscore_N) = \scorefunction(\noisydataset)$ is a vector whose elements are the noise scores for each $\samplepairnoisy{\vectorindex{i}}$ pair (the higher the score, the higher the predicted likelihood of a label error),
\begin{equation}
    \scorefunction : (\featurespace \times \labelspace)^N \to \RR^N,
\end{equation}
is a noise scoring function, and $\detectionthreshold \in \realnumbers$ is a decision threshold that can be adjusted to balance sensitivity and specificity.

In this work, we restrict attention to detection methods that can be expressed as
\begin{equation}\label{eq:no-further-training}
    \scorefunction(\noisydataset) = \scorefunction'(\detectionparameters(\noisydataset))
\end{equation}
where $\detectionparameters(\noisydataset)$ denotes a collection of values acquired during conventional training of a classification model $\modelfunction_\modelparameters$ on $\noisydataset$, which we refer to as the \textit{training dynamics}, while $\scorefunction'$ denotes a procedure that does \textit{not} require further training. Almost all of the methods evaluated in this paper consider
\begin{equation}
    \detectionparameters(\noisydataset) = \left\{\left(\modellogits\vi{i}\epochindex{1},\ldots,\modellogits\vi{i}\epochindex{E}, \samplelabelnoisy\vi{i} \right)\right\}_{i=1}^N 
\end{equation}
where
\begin{equation}\label{eq:logits-epoch}
\modellogits\vi{i}\epochindex{t} = \modelfunction_{\modelparameters\epochindex{t}}(\samplefeature\vi{i})  
\end{equation}
denotes the logit vector corresponding to input $\samplefeature\vi{i}$ collected at the $t$-th training epoch and $E \geq 1$ denotes the last training epoch (or desired stopping epoch when using checkpointing). However, for greater generality, we also allow 
\begin{equation}\label{eq:weight-averaging-dynamics}
    \detectionparameters(\noisydataset) = \left\{\left(\samplefeature\vi{i}, \modellogits\vi{i}\epochindex{1},\ldots,\modellogits\vi{i}\epochindex{E}, \samplelabelnoisy\vi{i} \right)\right\}_{i=1}^N \cup \left\{A\big(\modelparameters\epochindex{1}, \ldots, \modelparameters\epochindex{E}\big) \right\}
\end{equation}
where $\modelparameters\epochindex{t}$ denotes the model parameters obtained at epoch $t$ and $A(\cdot)$ denotes some averaging function, since such a parameter averaging may already be present in the training method.


Our restriction to \eqref{eq:no-further-training} is motivated by three main considerations: First, it ensures that no data from outside the problem at hand is used, allowing us to strictly control the experimental variables and perform a fair, self-contained benchmark of the detection algorithms. Second, it makes the detection methods universally applicable to scenarios where foundation models are unavailable or in very specialized domains. Third, it enables the methods to be incorporated into conventional training, requiring no further development effort. Methods that require specialized training procedures focused on noisy label detection are therefore outside the scope of this work.

Because the quality of the training dynamics $\detectionparameters(\noisydataset)$ is highly correlated with the performance of the underlying classification model $\modelfunction_\modelparameters$, optimizing the training hyperparameters becomes crucial for accurate detection. In practice, a noisy validation set drawn from the same distribution as $\noisydataset$ is typically the only available option for this optimization, since obtaining a clean validation set would itself require manual labeling effort. Therefore, we adopt a noisy validation set in our experimental setup to tune the classification models.


Analogous to training classification models, most detection methods also include additional hyperparameters other than the decision threshold itself. Tuning these parameters poses a significant challenge, given that the detection task is unsupervised by nature. In this work, we primarily use the default settings recommended by the authors, but we also briefly explore optimizing these parameters using a small validation subset containing both noisy and corrected labels for each sample.
}

\subsection{A Taxonomy of Detection Methods}
\new{
As discussed above, in this paper, a detection method is any algorithm that produces noise scores $\noisyscore \in \RR^N$ from the training dynamics $\detectionparameters(\noisydataset)$ without retraining. 
Rather than viewing detection methods as monolithic and isolated algorithms, we propose a taxonomy that can represent many common detection methods in the literature by defining three axes:

\begin{enumerate}
    \item \textbf{Gathering strategy:} Dictates whether the model producing the training dynamics for a given sample is trained on data containing that sample (in-sample) or without ever seeing it (out-of-sample).
    \item \textbf{Label disagreement measure (LDM):} 
    Measures the extent to which a logit vector disagrees with a given label. It is exemplified by and generalizes the loss function $\ell$ used for training the classifier.

    \item \textbf{Aggregation method:} Specifies how the training dynamics, combined with an LDM, is summarized into a single scalar noise score $\noisyscore_i$ for each $\samplepairnoisy{\vectorindex{i}}$ sample.

\end{enumerate}
For example, the classic small-loss selection heuristic used in Co-teaching~\cite{han_co-teaching_nodate} can be represented as an \textit{in-sample} gathering strategy, 
with a \textit{cross-entropy} loss function as LDM, evaluated at the \textit{last epoch} as aggregation method (effectively no aggregation). 
In the following, we first describe all the possible choices considered in this paper for each of these axis, before specifying how existing methods proposed in the literature fit into this taxonomy.

}

\subsection{Gathering Strategies}
\label{sec:gathering_protocols}

\definecolor{darkblue}{HTML}{1f77b4}
\definecolor{myorange}{HTML}{ff7f0e}
\definecolor{mygreen}{HTML}{2ca02c}
\begin{figure}
\centering
\begin{tikzpicture}[
  circle node input/.style={circle, draw=black!70, fill=gray!30, minimum size=6mm, inner sep=0pt},
  circle node train/.style={circle, draw=black!70, fill=myorange!60, minimum size=6mm, inner sep=0pt},
  circle node gather/.style={circle, draw=black!70, fill=darkblue!60, minimum size=6mm, inner sep=0pt},
  circle node gather and train/.style={circle, draw=black!70, fill=mygreen!60, minimum size=6mm, inner sep=0pt},
  rectscore/.style={rectangle, draw=black!70, fill=gray!20, minimum size=4mm, inner sep=0pt},
  >=Latex
]

\node[anchor=west] (A) at (0,0) {\textbf{A) In-sample}};
\matrix (poolA) [matrix of nodes, nodes in empty cells,
  nodes={circle node input}, column sep=3mm, row sep=0mm,
  below=4mm of A] { & & & & & & & \\ };
\matrix (matA) [matrix of nodes, nodes in empty cells,
  nodes={circle node gather and train}, column sep=3mm, row sep=0mm,
  below=6mm of poolA] { & & & & & & & \\ };
\draw[->, thick] (poolA.south) -- ++(0,-3mm) -| (matA.north);

\matrix (scoresA) [matrix of nodes, nodes in empty cells,
  nodes={rectscore}, column sep=4mm, row sep=1mm,
  below=6mm of matA] { & & & & & & & \\ };
\draw[->, thick] (matA.south) -- ++(0,-3mm) -- (scoresA.north);

\draw[decorate,decoration={brace,mirror,amplitude=4pt}]
  (scoresA.south west) -- (scoresA.south east)
  node[midway,yshift=-12pt,font=\footnotesize] {Collected scores};

\node[anchor=west] (B) at (7.5,0) {\textbf{B) Out-of-sample}};
\matrix (poolB) [matrix of nodes, nodes in empty cells,
  nodes={circle node input}, column sep=3mm, row sep=0mm,
  below=4mm of B] { & & & & & & & \\ };
\matrix (folds) [matrix of nodes, nodes in empty cells,
  nodes={circle node train}, column sep=3mm, row sep=1mm,
  below=6mm of poolB] {
  & & & & & & & \\
  & & & & & & & \\
  & & & & & & & \\
  & & & & & & & \\
};
\foreach \r/\c in {1/1,1/2,2/3,2/4,3/5,3/6,4/7,4/8} {
  \node[circle node gather] at (folds-\r-\c) {};
}
\draw[->, thick] (poolB.south) -- ++(0,-4mm) -| (folds.north);

\matrix (out) [matrix of nodes, nodes in empty cells,
  nodes={circle node gather}, column sep=3mm, row sep=0mm,
  below=6mm of folds] { & & & & & & & \\ };
\draw[->, thick] (folds.south) -- ++(0,-3mm) -| (out.north);
\draw[decorate,decoration={brace,amplitude=4pt}]
  (folds.north east) -- (folds.south east)
  node[midway,xshift=24pt,font=\footnotesize] {4-fold};

\matrix (scoresB) [matrix of nodes, nodes in empty cells,
  nodes={rectscore}, column sep=4mm, row sep=1mm,
  below=6mm of out] { & & & & & & & \\ };
\draw[->, thick] (out.south) -- ++(0,-3mm) -- (scoresB.north);

\draw[decorate,decoration={brace,mirror,amplitude=4pt}]
  (scoresB.south west) -- (scoresB.south east)
  node[midway,yshift=-12pt,font=\footnotesize] {Collected scores};

\matrix [matrix of nodes, nodes={anchor=west}, column sep=4mm, row sep=2mm, below=12mm of scoresA] (legend) {
  \node[circle, draw=black!70, fill=myorange!60, minimum size=5mm, inner sep=0pt] {}; & Training sample \\
  \node[circle, draw=black!70, fill=darkblue!60, minimum size=5mm, inner sep=0pt] {}; & Prediction gathering sample \\ 
  \node[circle, draw=black!70, fill=mygreen!60, minimum size=5mm, inner sep=0pt] {}; & Training and prediction sample \\ 
};

\end{tikzpicture}
\caption{Data gathering representation for in-sample and out-of-sample methods. Blue circles indicate samples for which predictions are stored, orange circles indicate samples used only for training, and green samples indicate samples in which both training and gathering of predictions are made.}
\label{fig:data_gathering_diagram}
\end{figure}
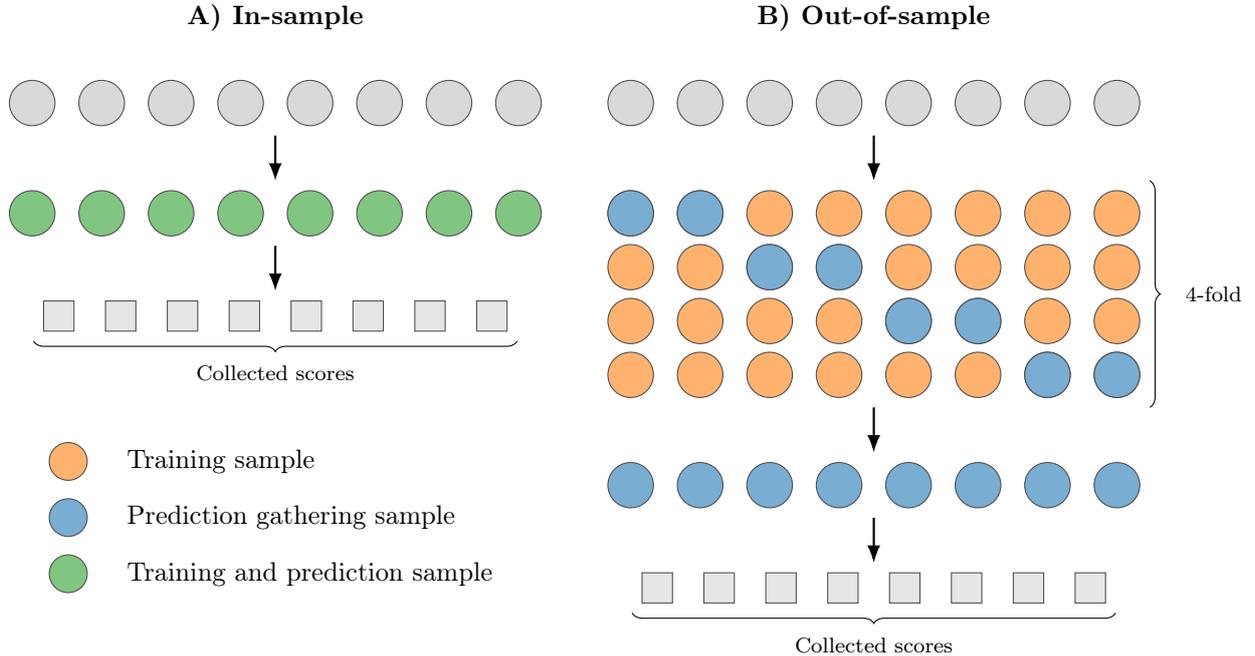

\new{
The gathering strategy determines how samples are partitioned during training dynamics acquisition to avoid or exploit memorization effects (see Figure \ref{fig:data_gathering_diagram}).

\begin{description}
\item[In-sample:] Training dynamics for each sample are obtained from a model trained on the entire dataset, including the sample itself. While computationally efficient (requiring only a single training run), it risks models overfitting and memorizing noisy labels in later epochs. In-sample methods rely on the early learning phenomenon \cite{arpit_closer_2017} to capture clean signals before memorization occurs.

\item[Out-of-sample:] To prevent memorization, this strategy utilizes cross-validation. The dataset is split into $K$ folds and $K$ separate models are trained, each leaving out one fold. The training dynamics for each sample are then collected from the model that did not see that sample during training.

\end{description}

It is important to note that the choice of the gathering strategy also impacts exactly \textit{when} the training dynamics are recorded during training, since, for in-sample methods, the logits at any epoch are computed while the model parameters are still updating, before the end of the epoch (in-epoch), while, for out-of-sample methods, the logits are computed necessarily after the epoch has ended (post-epoch). Strictly speaking, \eqref{eq:logits-epoch} is more precisely expressed as
\begin{equation}\label{eq:logits-epoch-precise}
\modellogits\vi{i}\epochindex{t} = \modelfunction_{\modelparameters\epochindex{i,t}}(\samplefeature\vi{i})  
\end{equation}
where $\modelparameters\epochindex{i,t}$ denotes the parameters of the model assigned to evaluate sample $i$ at epoch $t$. In particular, for in-sample methods, $\modelparameters\epochindex{i,t}=\modelparameters\epochindex{j,t}$ if and only if samples $i$ and $j$ are in the same batch, while, for out-of-sample methods, $\modelparameters\epochindex{i,t}=\modelparameters\epochindex{j,t}$ if and only if samples $i$ and $j$ are in in the same fold. An exception are in-sample methods using weight averaging \eqref{eq:weight-averaging-dynamics}, which must necessarily be evaluated post-epoch (in fact, post-training).
}




\subsection{Label Disagreement Measures}
An LDM is a function $L: \RR^K \times \labelspace \to \RR$ that measures the extent to which a logit vector $\modellogits \in \RR^K$ disagrees with the assigned label $\samplelabelnoisy \in \labelspace$. In the following, let $\modelprobas = \softmax(\modellogits)$ and $\hat{\samplelabel} = \arg\max_{k} \modellogits\vi{k}$.

\begin{description}
\item[Cross-Entropy (CE) loss:] The negative log of the estimated probability of the assigned class:
\begin{equation}
    \agreementf(\modellogits, \samplelabelnoisy) = -\log \modelprobas\vi{\samplelabelnoisy}.
\end{equation}
Naturally, it is strictly monotonic with respect to the estimated probability of the assigned class $\modelprobas\vi{\samplelabelnoisy}$ (also known as self-confidence).

\item[Jensen-Shannon divergence (JSD):] The JSD is a symmetric measure of the dissimilarity between two probability distributions. For $p,q \in \Delta_{K-1}$, it is defined as
\begin{equation}
    \text{JSD}(p,q) = \frac{1}{2} D\left(p \left\| \frac{p+q}{2}\right.\right) + \frac{1}{2} D\left(q \left\| \frac{p+q}{2}\right.\right)
\end{equation}
where $D(p \| q) = \sum_{k=1}^K p_k \log(p_k/q_k)$ denotes the Kullback–Leibler divergence. When used as an LDM, the JSD is computed for the estimated distribution $\modelprobas$ with respect to a one-hot encoding $\sampleonehotlabelnoisy$ of the noisy label $\samplelabelnoisy$, namely,
\begin{equation}
    \agreementf(\modellogits, \samplelabelnoisy) = \text{JSD}(\modelprobas,\sampleonehotlabelnoisy)
\end{equation}
where $\sampleonehotlabelnoisy = (\indicatorfunction(\samplelabelnoisy=1),\ldots,\indicatorfunction(\samplelabelnoisy=K))$. It can be shown that, for the same $\samplelabelnoisy$, the JSD and the CE loss are related by a strictly increasing transformation.


\item[Logit Margin (LM):] Measures the difference between the largest competing class logit and the assigned class logit:
\begin{equation}
    \agreementf(\modellogits, \samplelabelnoisy) = \max_{k\neq \samplelabelnoisy} \modellogits\vi{k} - \modellogits\vi{ \samplelabelnoisy}.
    \label{eq:logit_margin}
\end{equation}
Intuitively, when LM is negative, the model attributes the most confidence to the assigned class. 
Conversely, when LM is positive, the most confident class disagrees with the assigned label. Note that, to reflect disagreement, our definition is the negative of the original definition in \cite{pleiss_identifying_nodate}.

\new{
\item[Class Prediction Disagreement (CPD):] Evaluates whether the predicted class differs from the observed label:
\begin{equation}
    \agreementf(\modellogits, \samplelabelnoisy) = \indicatorfunction\left(\hat{\samplelabel} \neq \samplelabelnoisy \right).
\end{equation}
Note that it is equivalent to the 0-1 loss.
}
\end{description}

\subsection{Aggregation Methods}

The aggregation method specifies how the training dynamics $\detectionparameters(\noisydataset)$ are processed and turned into a single noise score for each sample. In the following, let $\noisyscore_i$ denote the noise score for the $i$th sample and let $L$ denote some LDM. Evaluating an LDM on a particular logit-label pair produces a label disagreement (LD) score for that pair.

\begin{description}
\item[Last epoch:] Uses the LD score measured at the final training epoch $\epochs$:
\begin{equation}
    \noisyscore\vi{i} = \agreementf(\modellogits\vi{i}\epochindex{\epochs}, \samplelabelnoisy\vi{i}).
    \label{eq:last_agg}
\end{equation}

\item[Mean:] Averages the LD scores of a sample over all training epochs:
\begin{equation}
    \noisyscore\vi{i} = \frac{1}{\epochs}\sum_{t = 1}^{ \epochs}\agreementf(\modellogits\vi{i}\epochindex{t}, \samplelabelnoisy\vi{i}).
    \label{eq:mean_agg}
\end{equation}

\item[Mean Probability:] Inspired by Snapshot Ensembles \cite{huang_snapshot_2017} and Dataset Cartography \cite{swayamdipta_dataset_cartography_2020}, this method first aggregates probability outputs over all epochs:
\begin{equation}
    \bar{\modelprobas}\vi{i} = \frac{1}{\epochs}\sum_{t=1}^{\epochs} \softmax(\modellogits\vi{i}\epochindex{t}).
    \label{eq:mean_probas}
\end{equation}
We generalize it by mapping the mean probability vector back to logit space (setting $C=0$):
\begin{equation}
    \label{eq:logits_to_probas}
    \bar{\modellogits}\vi{i} = \log{\bar{\modelprobas}\vi{i}} + C.
\end{equation}
Finally, the noise score of the $i$th sample is computed as
\begin{equation}
    \label{eq:score_from_mean_logits}
    \noisyscore\vi{i} = \agreementf(\bar{\modellogits}\vi{i}, \samplelabelnoisy\vi{i}).
\end{equation}

\item[Stochastic weight averaging (SWA) \cite{izmailov_averaging_2018}:] 
Computes $\noisyscore_i$ using \eqref{eq:score_from_mean_logits}, where
\begin{equation}
\bar{\modellogits}\vi{i} = \modelfunction_{\bar{\modelparameters}}\left(\samplefeature\vi{i}\right), \label{eq:swa_agg}
\end{equation}
and
\begin{equation}
\bar{\modelparameters} = A(\modelparameters\epochindex{1},\ldots,\modelparameters\epochindex{E})
\end{equation}
denotes the average of model parameters across all epochs.
In our experiments, we use
\begin{equation}
A(\modelparameters\epochindex{1},\ldots,\modelparameters\epochindex{E}) = \frac{1}{\epochs}\sum_{t=1}^{\epochs}\modelparameters\epochindex{t}.
\end{equation}
\new{
\item[Late Stopping:] Calculates how many epochs it takes for the model to predict the assigned label (or, more generally, to achieve a sufficiently small LD score) for $k$ consecutive epochs \cite{late_stopping_paper}:
\begin{equation}
\noisyscore\vi{i} = 
\min \left(
\left\{\, t \in \{k,\dots,\epochs\} \;:\; L\left(\modellogits\vi{i}\epochindex{t-j}, \samplelabelnoisy\vi{i}\right) \leq \delta,\; j=0,\ldots,k-1 \,\right\}
\;\cup\; \{\epochs+1\}
\right)
\end{equation}
where $\delta \in \RR$. In our experiments, we consider only the case of CPD with $0 \leq \delta < 1$ (such that $L(\modellogits\vi{i}\epochindex{t-j}, \samplelabelnoisy\vi{i}) \leq \delta \iff L(\modellogits\vi{i}\epochindex{t-j}, \samplelabelnoisy\vi{i}) = 0$), which corresponds to the original formulation in \cite{late_stopping_paper}. We also set $k=1$ following the authors' recommendation.
}

\new{
\item[CTRL:] Follows a clustering approach to classify noisy and clean samples. For each assigned class subset, training dynamics are divided in to $W$ windows (over training epochs) and, for each window, $K$-means is performed and samples on the $K_s$ clusters with highest loss are voted as noisy. Finally, the total number of votes for each sample is used for ranking, where samples with more votes have higher noise score. In the original formulation, samples are classified as noisy by thresholding the votes. In a best-effort, we instead define the noise score as the number of votes over the windows instead of the binary classification, which would produce a less informed ranking. See Algorithm \ref{alg:ctrl_votes} in \ref{app:ctrl_impl} for more details.
}

\end{description}

\subsubsection{Confident Learning (CL)}

\newcommand{\confidentlabel}{y^*}
\new{
CL identifies label errors by estimating the joint noise distribution $Q_{\samplelabelnoisy, \samplelabel}$ between observed labels $\samplelabelnoisy$ and latent true labels $\samplelabel$ \cite{northcutt_confident_2021}. Let $\noisydataset_{\samplelabelnoisy=k} = \{ \samplepairnoisy{\vi{i}} \in \noisydataset : \samplelabelnoisy\vi{i} = k \}$ denote the subset of samples with observed label $k$. First, a class-specific probability threshold $t_k = \frac{1}{|\noisydataset_{\samplelabelnoisy=k}|} \sum_{\samplepairnoisy{\vi{i}} \in \noisydataset_{\samplelabelnoisy=k}} \modelprobas\vi{i,k}$ is computed for each class $k$. Samples whose predicted probability exceeds the threshold for some class are assigned a confident predicted label $\confidentlabel\vi{i}$ (breaking ties via $\arg\max_k \modelprobas\vi{i,k}$), populating a $\numberofclasses \times \numberofclasses$ count matrix $\confidentjoint_{\samplelabelnoisy, \confidentlabel}$ whose off-diagonal entries ($\samplelabelnoisy \neq \confidentlabel$) count estimated label errors. Calibrating $\confidentjoint_{\samplelabelnoisy, \confidentlabel}$ yields the estimated joint noise distribution, where indices $i$ and $j$ range over $\labelspace$ and denote the observed and confident labels, respectively:
\begin{equation}
    \hat{Q}_{ij} \propto \frac{\confidentjoint_{ij}}{\sum_{j'} \confidentjoint_{ij'}} \cdot |\noisydataset_{\samplelabelnoisy=i}|, \quad \text{s.t.} \sum_{i,j} \hat{Q}_{ij} = 1.
    \label{eq:q_normalization}
\end{equation}
Here, $\numberofsamples \cdot \hat{Q}_{ij}$ estimates the count of label errors in partition $(\samplelabelnoisy=i, \confidentlabel=j)$. Based on these estimates, CL presents a baseline and three variants to detect and rank label errors:

\begin{itemize}
    \item \textbf{Baseline 1 ($C_{\text{confusion}}$ / CC):} Flags any sample whose predicted label disagrees with its observed label ($\arg\max_k \modelprobas\vi{i,k} \neq \samplelabelnoisy\vi{i}$), ranked by ascending self-confidence $\modelprobas\vi{i, \samplelabelnoisy\vi{i}}$.
    \item \textbf{Method 2 ($\confidentjoint_{\samplelabelnoisy, \confidentlabel}$ / CYY):} Flags all samples that fall into any off-diagonal entry of $\confidentjoint_{\samplelabelnoisy, \confidentlabel}$ (i.e., samples assigned $\confidentlabel\vi{i} \neq \samplelabelnoisy\vi{i}$), ranked by ascending self-confidence $\modelprobas\vi{i, \samplelabelnoisy\vi{i}}$.
    \item \textbf{Method 3 (Prune by Class - PBC):} For each observed class $i$, computes an error budget $\numberofsamples \cdot \sum_{j \neq i} \hat{Q}_{ij}$ and selects that many samples from $\noisydataset_{\samplelabelnoisy=i}$ with the lowest self-confidence $\modelprobas\vi{l,i}$.
    \item \textbf{Method 4 (Prune by Noise Rate - PBNR):} For each off-diagonal cell $(i, j)$, $i \neq j$, independently selects $\numberofsamples \cdot \hat{Q}_{ij}$ samples from $\noisydataset_{\samplelabelnoisy=i}$ with the largest probability margin $\modelprobas\vi{l,j} - \modelprobas\vi{l,i}$, taking the union across cells (so a sample may be flagged by multiple cells).
\end{itemize}

Confident Learning is a staple of the noisy label literature and one of the most prominent methods employing out-of-sample (OOS) probability gathering. It does not, however, fit neatly into our taxonomy of temporal aggregation paired with an LDM: the per-class probability calibration and joint matrix estimation introduce dataset-level structure absent from pointwise LDMs, and the method first partitions the dataset into noisy and clean subsets before ranking the flagged samples, a two-stage procedure that cannot be reduced to a single per-sample score. Nonetheless, CL operates on predictions from a single checkpoint (in our experiments, the early-stopping epoch), making its temporal aspect directly comparable to the Last aggregation. Although CL is traditionally paired with K-fold for OOS gathering, this is a design choice rather than a requirement, raising the question of whether OOS gathering is truly necessary for CL and whether it could benefit other detection methods as well.

}
\subsection{Summary of existing methods}

\begin{table}[t]
\centering
\begin{tabular}{lcccccc}
\toprule
 & \multicolumn{6}{c}{\textbf{Aggregation}} \\\cmidrule{2-7}
\textbf{Measure}         & \textbf{Last}       & \makecell[c]{\textbf{Mean}\\\textbf{Probability}} & \textbf{SWA} & \textbf{Mean} & \textbf{CTRL} & \makecell[c]{\textbf{Late}\\\textbf{Stopping}} \\ 
\midrule
\textbf{Cross-Entropy}  & \multirow{3}{*}{\makecell[c]{
    Co-teaching \cite{han_co-teaching_nodate}\\
    DivideMix \cite{li_dividemix_2020}\\
    Unicon \cite{karim_unicon_2022}\\
    Jo-SRC \cite{yao_josrc_2021}
}} 
                        & \multirow{3}{*}{\shortstack{\makecell[c]{Dataset\\Cartography\\\cite{swayamdipta_dataset_cartography_2020}}}}
                        & \multirow{3}{*}{$\circ$} 
                        & \makecell[c]{O2U-net \cite{huang_o2u-net_2019}\\AUL \cite{aul_paper}}
                        & \makecell[c]{CTRL\\\cite{yue_ctrl_2023}}
                        & \multirow{3}{*}{$\times$}
                        \\ 
\cmidrule(lr){1-1} \cmidrule(lr){5-6}
\multirow{2}{*}{\textbf{Jensen-Shannon}} 
                        & \multirow{2}{*}{} 
                        & \multirow{2}{*}{} 
                        & \multirow{2}{*}{} 
                        & \multirow{2}{*}{$\circ$} 
                        & \multirow{2}{*}{$\circ$} 
                        & \multirow{2}{*}{}
                        \\
                        &                     
                        &                          
                        &              
                        &               
                        &               
                        &
                        \\ 
\cmidrule{1-7}
\textbf{Logit Margin}   
        & $\circ$             
        & $\circ$                  
        & $\circ$      
        & AUM \cite{pleiss_identifying_nodate}
        & $\circ$
        & $\times$
        \\
\cmidrule{1-7}
\textbf{\makecell[l]{Class Prediction\\Disagreement}} 
    & $\times$
    & \makecell[c]{SELF \cite{self_paper}}                 
    & $\times$      
    & \makecell[c]{Wrong\\Event} \cite{wrong_event_paper}
    & $\circ$
    & \makecell[c]{Late\\Stopping \cite{late_stopping_paper}}
    \\
\bottomrule
\end{tabular}
\caption{Relation between aggregation-measure combinations and existing in-sample detection methods.
Blank cells with $\circ$ represent combinations evaluated in this work but not previously explored, while $\times$ indicate combinations not evaluated in this work.
Cross-Entropy and Jensen-Shannon measures are equivalent under `Last', `Mean Probability' and `SWA' aggregations.}
\label{tab:in_sample_relations}
\end{table}
\new{
In Table \ref{tab:in_sample_relations}, we show how methods that incorporate detection during training are related to combinations of aggregation and measures. As mentioned previously, Co-teaching \cite{han_co-teaching_nodate} is represented by the Last aggregation with the Cross-Entropy LDM. Specifically, Co-teaching trains two models and during training, for each batch, the CE loss of one model is used to filter samples that the other uses for backpropagation. This filtering approach avoids memorization during training.

DivideMix \cite{li_dividemix_2020} also trains two models that cross-filter samples for each other; however, key differences regarding detection are that selection is performed at the end of each epoch rather than per-batch, and the amount of selected samples is guided by a Gaussian Mixture Model (GMM) fitted on the CE losses instead of a fixed fraction. Furthermore, DivideMix leverages the filtered subset for semi-supervised learning (SSL).

Several robust training methods follow a similar approach to DivideMix by separating the identified mislabeled samples as an unlabeled set and using SSL instead of entirely removing them. Since our focus is primarily on the detection aspect, we do not go into detail on how training is performed in these methods.

Unicon \cite{karim_unicon_2022} also trains two models; however, instead of cross-filtering, it uses the average predictions of both models to calculate the JSD as the LDM. To filter samples at the end of each epoch, the fraction of noisy labels is estimated using a threshold calculated from the average and minimum JSD. Then, for each class, a subset proportional to this estimated noise rate is selected and ranked by JSD. This per-class filtering is proposed to address class imbalance and bias toward easier classes. 

Jo-SRC \cite{yao_josrc_2021} also adopts the JSD between predicted and label distributions as its clean sample selection criterion, exploiting its bounded range $[0,1]$ to define a global threshold rather than selecting a pre-specified fraction of each mini-batch as in Co-teaching; within our taxonomy, this maps to the Last aggregation with the CE/JSD measure.

We note that under aggregations that produce a single probability vector per sample (Last, Mean Probability, SWA), Cross-Entropy and Jensen-Shannon divergence produce identical rankings since they are related by a monotonic function. This is not to say that they are interchangeable in the methods described above, since the transformation is non-linear and will interfere with how both the GMM in DivideMix and the threshold calculation in Unicon behave. For Co-teaching, they are indeed interchangeable because a fixed fraction of the batch is selected and ranking order is kept.

Dataset Cartography proposes using the average model confidence (i.e., the probability of the assigned class) over training epochs as a noisy label indicator, after which the model is retrained on the cleaned dataset. Notably, because the cross-entropy loss is a strictly monotonic transformation of this confidence, this detection criterion is equivalent to using the Mean Probability aggregation with the CE/JS measure.

Following a similar retraining approach, O2U-net \cite{huang_o2u-net_2019} uses the Mean Cross-Entropy loss to detect label noise, while AUM \cite{pleiss_identifying_nodate} employs the Mean Logit Margin over training epochs before retraining the model from scratch on the cleaned dataset.

SELF \cite{self_paper} progressively filters noisy labels during training in an iterative semi-supervised framework. At each epoch, it maintains an exponential moving average (EMA) of the predicted probabilities for each sample and filters a sample if its assigned label disagrees with the class that has the highest average probability. Within our taxonomy, this underlying detection criterion corresponds to the MeanProb-CPD configuration. While SELF uses an EMA instead of a simple arithmetic mean, both methods share the core logic of ensembling predictions over epochs. Furthermore, since class prediction disagreement is mathematically equivalent to a positive Logit Margin on the mean probabilities, MeanProb-CPD corresponds to thresholding the MeanProb-LM configuration at zero. For this reason, we do not evaluate this combination in our benchmark, as a binary disagreement score leads to frequent rank ties and is less informative than the continuous ranking provided by LM. In fact, we also exclude CPD when combined with other aggregations that yield a single prediction vector per sample (namely, Last and SWA) for the same reason.

Wrong Event \cite{wrong_event_paper} utilizes the cumulative number of prediction-label mismatches across training epochs as an instance-level metric. Instead of performing a hard binary classification, it fits a Beta Mixture Model (BMM) to the distribution of these mismatch counts to model sample cleanliness and difficulty. Within our taxonomy, we isolate the underlying wrong event accumulator as a detection method corresponding to the Mean-CPD configuration. Since the BMM is a general probabilistic modeling step, we argue that other LDMs could be used in place of the disagreement count; alternative measures and aggregations that provide better separation between noisy and clean samples could therefore be substituted to improve downstream training.

Beyond providing a structured overview of existing work, our taxonomy reveals 13 combinations that, to the best of our knowledge, have not been previously explored. We discard 5 of these due to lack of methodological motivation or due to yielding strictly weaker rankings than existing alternatives. We experimentally evaluate the remaining 8, contributing a substantial set of novel data cleaning strategies.
}

\subsubsection{Excluded methods}
\new{
Although we aim at investigating simpler approaches that can be directly applied on a standard training pipeline with minor changes, some interesting and promising methods that significantly change the training routine in favor of detection have been proposed in the literature:

\begin{itemize}
    \item DynaCor \cite{kim_dynacor_2024}: Detects label noise by deliberately corrupting an augmented subset of the dataset and using training dynamics to learn a convolutional neural network (CNN) for classifying mislabeled samples. The requirement of extra corrupted samples in the dataset is unusual for standard training pipelines. Furthermore, the training of the detector CNN introduces many hyperparameters to optimize in the usual unsupervised label noise detection setting. For these reasons, it was excluded from our study.
    \item ChronoSelect \cite{chronoselect_paper}: Maintains a four-stage moving average of predicted probabilities (called ``Dynamics Temporal Memory'') and compares predictions under weak and strong augmentations to partition the dataset into clean, boundary, and noisy subsets. The requirement of using weak and strong augmentations during training places this method outside the scope of our study.
    \item INCV \cite{chen_incv_nodate}: Iteratively partitions the dataset, training models on clean samples to identify additional clean samples in held-out portions. At each iteration, samples where the model agrees with the label are marked as clean, while a fraction of high-loss samples is flagged as noisy. While performing iterative detection is an interesting approach, it can be wrapped around any detection method; moreover, improvements in the underlying detection criterion would directly translate to the iterative version.
    \item SELF \cite{self_paper}: Progressively filters noisy labels using self-ensembling (running average predictions across training epochs) in an iterative semi-supervised framework. While its full iterative framework is excluded from our benchmark due to its training complexity, its underlying detection criterion is represented in our taxonomy, as described above.
\end{itemize}

A separate and noteworthy trend leverages feature representations and foundation models for noise detection. For example, training-free methods based on feature-space neighborhood consensus \cite{zhu_detecting_2022} bypass classifier training by assessing local label agreement in feature space. Similarly, CLIPCleaner \cite{feng_clipcleaner_2024} and VLM detectors \cite{wei_vlm_strong_detectors_2024} use pre-trained vision-language models to identify image-text semantic misalignment, while other approaches analyze structural properties of data manifolds \cite{kim_structural_labels_2024} or use generative diffusion models \cite{lee_label_diffusion_2023}. These methods are excluded from our taxonomy as they rely on external pre-trained models, auxiliary feature extractors, or wild data, whereas we focus on self-contained training dynamics derived during standard classifier training.
}

\subsection{Prior evaluation of detection methods}
Most prior work on evaluating detection methods is limited in scope. Studies typically evaluate new algorithms against only one or two baselines to justify the novel method, rather than establishing reproducible, system-wide rankings \cite{huang_o2u-net_2019,pleiss_identifying_nodate,yue_ctrl_2023}. They also suffer from methodological inconsistencies, such as varying hyperparameter tuning levels or using metrics without predefined operating points.

To the best of our knowledge, Chen et al.~\cite{chen_clean_2022} is the only prior study that systematically evaluates multiple detection methods (AUM, Last-Loss, and Dataset Cartography) under a unified framework. However, their work focuses on active data acquisition trade-offs (relabeling vs.\ acquiring new samples) under a fixed budget. In contrast, our work abstracts away the data acquisition trade-off and assumes perfect relabeling to isolate and benchmark detection performance directly.

%% file: 04_methods.tex
\section{Methods}

In this section, we present the primary metric used to compare detection methods, describe the \new{unified benchmark suite constructed from our taxonomy, and outline the optimization procedures, datasets, and training settings.}

\subsection{Evaluating detection methods}

Similar to binary classification, the task of detecting label errors involves identifying which samples belong to one of two classes, namely, the noisy set $\noisysubset$ or the clean set $\cleansubset$. Intending to compare different methods, we measure the remaining noise after performing a cleaning pass. In more detail, we assume that every sample detected as noisy, composing the set $\predictedasnoisy$, will be correctly relabeled by a specialist. This means that the number of noisy samples originally is
\begin{equation}
    |\noisysubset| 
    = \truepositive + \falsenegative,
\end{equation}
where 
\begin{equation}
\truepositive = |\noisysubset \cap \predictedasnoisy|
\end{equation}
denotes the number of true positives (correctly identified mislabeled samples) and 
\begin{equation}
\falsenegative = |\noisysubset \cap (\noisydataset\setminus{\predictedasnoisy})|
\end{equation}
denotes the number of false negatives (mislabeled samples incorrectly identified as clean). We can then define the noise rate as
\begin{equation}
    \noiserate = \frac{|\noisysubset|}{\numberofsamples} = \frac{\truepositive + \falsenegative}{\numberofsamples}.
\end{equation}

After cleaning (relabeling), all true positive samples will be corrected, while the false negatives will remain unchanged. Thus, the number of residual mislabeled samples simply becomes $\falsenegative$. The residual noise rate is then given by 
\begin{equation}
\residualnoise = \frac{\falsenegative}{\numberofsamples}.
\end{equation}
Finally, we define the normalized residual noise rate as
\begin{equation}
    \normalizedresidualnoiserate = \frac{\residualnoise}{\noiserate} = \frac{\falsenegative}{\truepositive + \falsenegative} = \fnr
\end{equation}
where FNR stands for false negative rate.
This metric describes the proportion of noise, with respect to the original noise fraction, that persists after cleaning the samples detected as noisy. Since it coincides exactly with the FNR, this is how we will refer to it. 

Under ideal cleaning, increasing sensitivity always maintains or decreases the FNR, but also increases the number of samples requiring analysis. This creates a trade-off between cleaning effort and residual noise: better detection methods achieve lower FNR for the same effort, or require less effort for the same FNR.

We analyze this trade-off by parameterizing the detection task by a budget $\budget \in [0, 1]$, which represents the fraction of positive predictions. In more detail, let $\noisyscore \in \mathbb{R}^N$ be the score produced by a detection method for each sample, and $\pi : \{1, \ldots, \numberofsamples\} \to \{1, \ldots, \numberofsamples\}$ be a permutation of the indices that would produce a sorted version of $\noisyscore$ in decreasing order, with ties broken in an arbitrary but deterministic way. Then we can define the predicted noisy set
$$\predictedasnoisy = \{\samplepairnoisy{\vectorindex{\pi(i)}} : i \in \{1, \ldots, \lceil \budget\numberofsamples \rceil\}\}.$$
Since tied values are ordered arbitrarily, the selection among samples with equal scores is effectively random, and the detection outcome within this tied region is equivalent to that of a random detector.
Naturally, this also works for methods that produce highly discretized scoring. For methods that produce a direct prediction, we encode $S$ as follows: $1$ for samples predicted as noisy and $0$ for samples predicted as clean. 

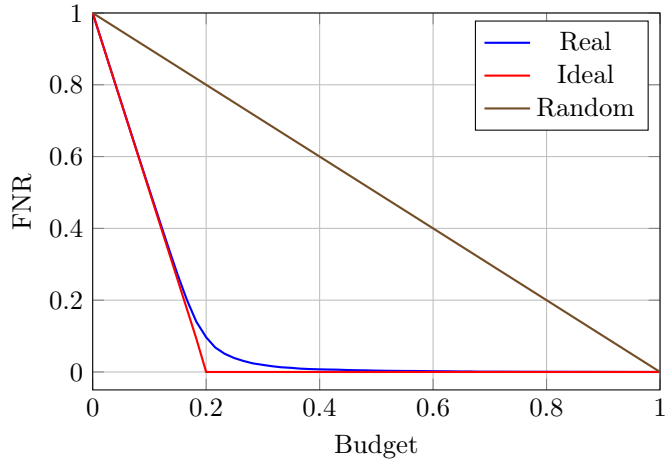
\begin{figure}
    \centering
    \input{fnr_metric_example}
    \caption{Example of the budget x FNR curve for an ideal, random and a real method in a setting with $\noiserate = 0.2$.}
    \label{fig:residual_noise_example}
\end{figure}

For an ideal detection method, the remaining noise fraction for a given budget is expressed as:
\begin{equation}
    \perfectfnr(\budget) = \begin{cases}
        1-\frac{\budget}{\noiserate} & \text{if } \budget < \noiserate \\
        0 & \text{if } \budget \geq \noiserate.
    \end{cases}
\end{equation}
That is, a piece-wise linear function going from $(0, 1)$ to $(\noiserate, 0)$ and then $(1, 0)$. Notably, a random detector is represented by a line going from $(0, 1)$ to $(1, 0)$. Both detectors are represented in Figure \ref{fig:residual_noise_example} together with a real detection method.

An interesting metric to consider is the value of $\remainingfnr(\noiserate)$, that is, the remaining noise after cleaning a fraction of samples equal to the noise rate. The ideal model has $\perfectfnr(\noiserate)=0$ for any noise rate, thus, this measure would indicate how much relative noise is left due to the imperfection of the detection method used. Evaluating at $\noiserate$ also takes into account the natural necessity of increasing the number of samples to investigate in noisier cases. We choose this metric to evaluate detection methods because it summarizes the performance of a method in one single number at a relevant operating point.

\subsection{Detection Methods}

\new{
While it is possible to explore the entire combinatorial space across the three axes of our taxonomy, which would total 46 configurations, we focus on the subset of combinations that address the most meaningful research questions. Our experimental design is guided by two main objectives. First, we pair each aggregation method with its compatible disagreement measures under in-sample, since these represent the most common setting and cover many methods in the literature. Second, we specifically investigate whether out-of-sample estimation, a requirement often assumed necessary for confident learning, is truly indispensable by comparing its in-sample and out-of-sample variants. We also include the OOS variants of some in-sample methods to assess the impact of OOS. This selection of methods allows us to draw actionable conclusions about method and measure interactions and the role of gathering strategy, without diluting statistical power across configurations that are either redundant or lack a clear methodological motivation.

Table \ref{tab:in_sample_relations} shows the combinations of in-sample methods evaluated and their correspondence to existing methods in the literature. In total, 8 new detection methods are explored as well.
Apart from these methods we also include out-of-sample and in-sample version of methods from the Confident Learning (CL) framework \cite{northcutt_confident_2021}: $C_{\text{confusion}}$ (CC), $\confidentjoint_{\samplelabelnoisy, \confidentlabel}$ (CYY), Prune By Class (PBC), and Prune By Noise Rate (PBNR).
}

\subsection{Datasets and Settings}
We evaluate all methods on multiple computer vision and tabular datasets, incorporating both synthetic and real noise. We provide the contingency matrices for each setting on \ref{app:contingency}. In Table \ref{tab:vision_datasets_stats}, we detail the statistics of each dataset.

While many noise types exist in the literature, we focus on symmetric and pairflip noise as they represent complementary extremes of noise sparsity while being easily described solely by noise rate. Symmetric noise affects all classes uniformly (dense), whereas pairflip noise has sparse noise transition. To complement these synthetic noise types, we also evaluate on real-world noise, which captures the complex patterns found in actual datasets.

\begin{table}[htbp]
    \centering
    \begin{tabular}{lccccc}
        \toprule
        \textbf{Dataset} & \textbf{\#} & \textbf{Original features} & \makecell[c]{\textbf{Transformed}\\\textbf{features}} & \textbf{Classes} & \textbf{Train/Test Split} \\
        \midrule
        CIFAR-10      & 60K     & $32\times 32$  & $32\times32$    & 10  & 50K / 10K \\
        CIFAR-100     & 60K     & $32\times 32$  & $32\times32$    & 100 & 50K / 10K \\
        CIFAR-10N \cite{cifar10N}    & 60K     & $32\times 32$  & $32\times32$    & 10  & 50K / 10K \\
        Clothing1M \cite{xiao2015learning}   & 1M+     & \makecell[c]{Varied\\(mostly $256\times 256$)\textsuperscript{1}} & $224\times224$ & 14  & 1M+24K / 10K \\
        Food-101N \cite{lee2017cleannet}    & 310K    & \makecell[c]{Varied\\(320–512 px typical)}\textsuperscript{2} & $128\times128$ & 101 & 285K / 25K \\
        Cardiotocography & 2126 & 21 & 21 & 3 & 1700 / 426 \\
        Credit Fraud & 285K\textsuperscript{3} & 30 & 30 & 2 & 1107 / 369 \\
        \makecell[l]{Human Activity\\Detection} & 10299 & 561 & 30 & 6 & 7352 / 2947 \\
        Letter Recognition & 20K & 16 & 16 & 26 & 15000 / 5000 \\
        Mushroom & 7864 & 22 & 67 & 2 & 5898 / 1966 \\
        Satellite & 6435 & 36 & 36 & 6 & 4435 / 2000 \\
        Sensorless Drive & 59K & 48 & 48 & 11 & 43882 / 14627 \\

        \bottomrule
    \end{tabular}

    \vspace{0.5em}
    \footnotesize
    \textsuperscript{1}\,Images range from $60\times 84$ to $461\times 1399$; Around $80\%$ are $256\times 256$. \\
    \textsuperscript{2}\,Images range from $80\times 70$ to $1511\times 3874$; most common are $426\times 320$, $320\times 426$, $320\times 320$, and $512\times 512$. \\
    \textsuperscript{3}\,Subsampled to balance classes with a ratio of 1:2 of positive to negative examples, following \cite{yue_ctrl_2023}.

    \caption{Dataset statistics}
    \label{tab:vision_datasets_stats}
\end{table}




\begin{table}[htbp]
    \centering
    \begin{tabular}{lcc}
        \toprule
        \textbf{Dataset} & \textbf{Training Transforms} & \textbf{Test Transforms} \\
        \midrule
        \multicolumn{3}{c}{\textbf{Vision Datasets}} \\
        \midrule
        \makecell[c]{
            CIFAR-10 \\ CIFAR-100 \\ CIFAR-10N
            } &
        \makecell[c]{4px padding \\ Random crop ($32\times32$) \\ Horizontal flip (50\%) \\ Normalization} &
        \makecell[c]{Normalization} \\
        \midrule
            Clothing1M &
        \makecell[c]{Resize ($256\times256$) \\ Random crop ($224\times224$) \\ Horizontal flip (50\%) \\ Normalization} &
        \makecell[c]{Resize \\ Center crop ($224\times224$) \\ Normalization} \\
        \midrule
            Food-101N &
        \makecell[c]{Random resize crop with \\ scale [0.08, 1.0] → $128\times128$ \\ Horizontal flip (50\%) \\ Normalization} &
        \makecell[c]{Resize (min edge = 128) \\ Center crop ($128\times128$) \\ Normalization} \\
        \midrule
        \multicolumn{3}{c}{\textbf{Tabular Datasets}$^*$} \\
        \midrule
        Cardiotocography &
        \multicolumn{2}{c}{\makecell[c]{Clip outliers¹ \\ Standard scaling}} \\
        \midrule
        Credit Fraud &
        \multicolumn{2}{c}{\makecell[c]{Clip outliers ($Q_{1\%}-Q_{99\%}$)}} \\
        \midrule
        \makecell[l]{Human Activity\\Detection} &
        \multicolumn{2}{c}{\makecell[c]{Standard scaling \\ PCA (n\_components=0.9) \\ Clip PC features²}} \\
        \midrule
        Letter &
        \multicolumn{2}{c}{\makecell[c]{Clip outliers ($Q_{1\%}-Q_{99\%}$) \\ Standard scaling}} \\
        \midrule
        Mushroom &
        \multicolumn{2}{c}{\makecell[c]{Drop constant features \\ Category consolidation³ \\ One-hot encoding \\ Standard scaling}} \\
        \midrule
        Satellite &
        \multicolumn{2}{c}{\makecell[c]{Clip outliers ($Q_{1\%}-Q_{99\%}$) \\ Standard scaling}} \\
        \midrule
        Sensorless Drive &
        \multicolumn{2}{c}{\makecell[c]{Clip outliers ($Q_{2\%}-Q_{98\%}$) \\ Standard scaling}} \\
        \bottomrule
    \end{tabular}
    
    \footnotesize
    \vspace{0.5em}
    \textsuperscript{*}Same transforms used for training and testing. \\
    \textsuperscript{1}ALTV: $Q_{2\%}$-$Q_{95\%}$, others: $Q_2\%$-$Q_{98}\%$ \\
    \textsuperscript{2}$PC_0$: $Q_1\%$-$Q_{99}\%$, others: $Q_{25}\%$-$Q_{75}\%$ \\
    \textsuperscript{3}Low-frequency categories grouped as 'other'
    \caption{Overview of the transforms used in each dataset}
    \label{tab:datasets_transforms}
\end{table}

For all tabular datasets, we follow the pre-processing methodology described in \cite{yue_ctrl_2023}, with one exception for the Cardiotocography dataset. Instead of using all 33 features as in the original work, we retain only the 21 features explicitly recommended in the dataset documentation, as some of the excluded features are not suitable for inference. Table \ref{tab:datasets_transforms} provides the details of the pre-processing transforms.

\subsubsection{Model Architectures}
For CIFAR datasets, we use the 9-layer Convolutional Neural Network (CNN) architecture proposed in \cite{han_co-teaching_nodate}. For Clothing1M, we employ ResNet-18 (Residual Network), while for Food-101N, we use ResNet-34 (Residual Network). For all tabular datasets, we implement a Multilayer Perceptron (MLP) architecture optimized for each specific setting, following \cite{yue_ctrl_2023}.

\subsubsection{Classification Hyperparameter Optimization}
For each dataset and noise level, we perform a manual optimization of the learning rate, learning rate scheduler, batch size, and weight decay by observing the 10-epoch rolling mean accuracy on a held-out noisy validation set. To determine the optimal number of training epochs, we train the model for a large number of epochs while monitoring validation performance and select the epoch that maximizes the validation metric.

We tailor the optimization strategy to each dataset type. For datasets with synthetic noise, we extensively optimize without noise first (both training and validation sets are clean) and then fine-tune the hyperparameters with noisy labels (both training and validation sets are noisy). For Food-101N and Clothing1M datasets, we optimize with noisy labels only. For tabular datasets, we also search for an MLP architecture following \cite{yue_ctrl_2023} while also optimizing the learning rate in $\{0.0001, 0.001, 0.01\}$ and weight decay in $\{0, 0.001\}$. 

All detection methods use these optimized hyperparameters when training models to collect detection signals. For out-of-sample methods, which require cross-validation, we perform the K-fold procedure within the training set only. All final hyperparameter settings are given in \ref{app:hyperparams}, and the accuracies achieved on the test set are given in \ref{app:classification_performance}.

\subsection{Detection Hyperparameter Optimization}

Outside of the choice of threshold, noisy label detection methods are generally considered hyperparameter-free and independent of the training procedure. It is generally assumed that optimizing the classification task improves the detection performance indirectly. Interestingly, prior work \cite{huang_o2u-net_2019} has demonstrated that modifying training dynamics, such as cyclically varying the learning rate, can significantly enhance detection performance even though classification performance may remain the same.

Optimizing training specifically for detection is beyond the scope of this paper. Instead, we follow a simpler approach and optimize only the aggregation window used by certain methods. These methods (e.g., Mean-CE, MeanProb-LM and Last-CE) rely on aggregating an agreement measure (such as the cross-entropy or logit margin) across multiple training epochs. The choice of which interval of epochs to aggregate over, the aggregation window, may significantly affect detection performance.

Our approach follows a simple yet effective two-step procedure:
\begin{enumerate}
    \item Training: We fully train the model for twice as long as the original training epochs, saving the output logits for all samples at each epoch.
    \item Post-hoc Optimization: Using the saved predictions, we search for the best aggregation window by selecting intervals of epochs (start and end) without the need for retraining.
\end{enumerate}
To find the optimal window, we search over 50 equally spaced epochs for the starting and final aggregation epochs, leading to a total of at most 1250 aggregation windows. Figure \ref{fig:optimization_diagram} shows the processes and acquired data necessary for optimization.

In the case of the CTRL method, which employs clustering over multiple time windows, we optimize its own hyperparameters post-hoc via grid search. Following the original paper, we explore combinations of the number of time windows $\{1, 2, 4\}$, the number of clusters per window $\{2, 3\}$, and the number of selected clusters $\{1, 2\}$.

For selecting these hyperparameters, we propose leveraging a detection tuning set, that is, a small subset of samples with indication of whether the given label is noisy or not. In this experiment, we reserve 10\% of the training dataset as tuning data for the detection task (note that this detection tuning data remains part of the training data). To evaluate the amount of annotation required, we experiment with different proportions of labeled data (10\%, 5\% and 1\%). \new{This range of values was chosen based on preliminary experiments where we observed that increasing the fraction beyond 10\% did not yield significant performance gains while substantially increasing the manual labeling effort. From a practical perspective, the effort required to manually label a larger fraction of the dataset would often be more effectively directed toward correcting the samples actually flagged by the detection method.} We evaluate methods on the remaining 90\% of training data, in contrast to the benchmark evaluation, where methods are assessed on the entire training set. Figure \ref{fig:splits_diagram} shows the splits used in this work.

\begin{figure}[h]
\centering
\begin{subfigure}[b]{0.55\textwidth}
\centering
\begin{tikzpicture}[
  node distance=0.4cm and 0.8cm,
  every node/.style={font=\footnotesize},
  process/.style={rectangle, draw, rounded corners, minimum height=1.1cm, minimum width=3.4cm, align=center, fill=blue!5},
  data/.style={rectangle, draw, minimum height=1.1cm, minimum width=3.4cm, align=center, fill=gray!10},
  arrow/.style={-{Latex[length=3mm]}, thick}
]
\node[data] (trainset) {Full training set};
\node[process, right=of trainset] (train) {\textbf{Train model}\\Save per-epoch logits};
\node[data, below=of train] (signals) {\textbf{Saved logits}\\($\epochs \times \numberofsamples \times \numberofclasses$)};
\node[process, below=0.4cm of signals] (search) {\textbf{Post-hoc search}\\Aggregation window\\or CTRL params};
\node[data, below=of search] (params) {\textbf{Optimized parameters}\\(Aggregation window\\or CTRL params)};
\node[data, left=of search] (tune) {Noisy label indicators\\for detection tuning set\\(e.g., 10\%, 5\%, 1\%)};
\node[process, below=0.4cm of params] (eval) {\textbf{Evaluate final}\\\textbf{detection performance}};
\node[data, left=of eval] (testset) {Remaining 90\%\\of training set};
\draw[arrow] (trainset) -- (train);
\draw[arrow] (train) -- (signals);
\draw[arrow] (signals) -- (search);
\draw[arrow] (tune) -- (search);
\draw[arrow] (search) -- (params);
\draw[arrow] (params) -- (eval);
\draw[arrow] (testset) -- (eval);
\end{tikzpicture}
\caption{Workflow for training and post-hoc detection optimization. The model is trained once, and the logits for each sample are saved across epochs. A small detection tuning set is used to select the best aggregation window or CTRL hyperparameters, which are then evaluated in the remaining dataset.}
\label{fig:optimization_diagram}
\end{subfigure}
\hfill
\begin{subfigure}[b]{0.4\textwidth}
\centering
\begin{adjustbox}{max width=\textwidth, keepaspectratio}
\input{Plots/dataset_splits}
\end{adjustbox}
\caption{The dataset splits used in this work. The dataset is divided into the usual train-validation-test splits for classification. One additional type of split is made for detection, represented by the three tuning splits used for tuning the detection methods.}
\label{fig:splits_diagram}
\end{subfigure}
\caption{Overview of the training/optimization workflow and the corresponding dataset splits.}
\label{fig:combined_diagrams}
\end{figure}

%% file: fnr_metric_example.tex
\begin{tikzpicture}
    \begin{axis}
        [
            xlabel={Budget},
            ylabel={FNR},
            xmin=0,
            xmax=1,
            ymin=-0.05,
            ymax=1,
            ymajorgrids=true,
            xmajorgrids=true,
            width=0.55\textwidth,
            height=0.4\textwidth,
            every axis plot/.append style={thick}
        ]
        \pgfplotstableread[col sep = tab]{fnr_metric_example_data.csv}\data
        \addplot+[mark=none] table[x=Predicted positives fraction, y=Relative residual noise] {\data};
        \addlegendentry{Real}

        \addplot+[mark=none] table[x=Predicted positives fraction, y=Perfect model] {\data};
        \addlegendentry{Ideal}

        \addplot+[mark=none] table[x=Predicted positives fraction, y=Random model] {\data};
        \addlegendentry{Random}
    \end{axis}
\end{tikzpicture}

%% file: Plots/dataset_splits.tex
\begin{tikzpicture}[
    box/.style={
        draw,
        line width=0.75pt,
        align=center,
        inner sep=0pt
    },
    grayBg/.style={fill={rgb,255:red,242; green,242; blue,242}},   
    whiteBg/.style={fill={rgb,255:red,255; green,255; blue,255}},  
    containerBg/.style={fill={rgb,255:red,250; green,250; blue,250}}, 
    dataLeaf/.style={
        box,
        grayBg,
        rounded corners=2pt,
        minimum width=5.0cm,
        minimum height=0.8cm,
        font=\small
    }
]

    \draw[box, containerBg, rounded corners=6pt] 
        (-3.8, 4.4) rectangle (3.8, -4.2);
    \node[font=\small\bfseries] at (0, 4.0) {Dataset};

    \draw[box, whiteBg, rounded corners=4pt] 
        (-3.3, 3.5) rectangle (3.3, -1.6);
    \node[font=\small\bfseries] at (0, 3.1) {Training set};

    \node[dataLeaf, whiteBg, minimum height=0.9cm, minimum width=5.6cm] at (0, 2.2) {90\% Detection test set};

    \draw[box, grayBg, rounded corners=3pt] 
        (-2.8, 1.4) rectangle (2.8, -1.3);
    \node[font=\small] at (0, 1.1) {10\% Tuning};

    \draw[box, whiteBg, rounded corners=2pt] 
        (-2.4, 0.7) rectangle (2.4, -1.1);
    \node[font=\small] at (0, 0.4) {5\% Tuning};

    \node[dataLeaf, grayBg, minimum width=4.0cm, minimum height=0.75cm] at (0, -0.5) {1\% Tuning};

    \node[dataLeaf, whiteBg, minimum width=6.6cm] at (0, -2.3) {Validation};
    \node[dataLeaf, whiteBg, minimum width=6.6cm] at (0, -3.4) {Test};

\end{tikzpicture}

%% file: 05_results.tex
\section{Results and Discussion}

\begin{table}
    \centering
    \begin{adjustbox}{max width=\textwidth, keepaspectratio}
    \input{Tables/fnratnr_cifar_synthetic}
    \end{adjustbox}
    \caption{FNR (\%) measured for a budget equal to the noise rate on CIFAR-10 and CIFAR-100. Best results are shown in bold.}
    \label{tab:cifar_fnr}
\end{table}

\begin{table}
    \centering
    \begin{adjustbox}{max width=\textwidth, keepaspectratio}
    \input{Tables/fnratnr_real}
    \end{adjustbox}
    \caption{FNR (\%) measured for a budget equal to the noise rate on the real-world noise datasets CIFAR-10N, Clothing1M and Food-101N. Best results are shown in bold.}
    \label{tab:real_world_fnr}
\end{table}

\begin{table}
    \centering
    \begin{adjustbox}{max width=\textwidth, keepaspectratio}
    \input{Tables/fnratnr_tabular}
    \end{adjustbox}
    \caption{FNR (\%) measured for a budget equal to the noise rate on tabular datasets. Best results are shown in bold.}
    \label{tab:tabular_fnr}
\end{table}

\subsection{Main Findings}

Tables \ref{tab:cifar_fnr}-\ref{tab:tabular_fnr} present false negative rates (FNR) at a budget equal to the noise rate across datasets with synthetic noise (vision and tabular) and real-world noise (vision only).
As can be seen, 
Mean Probabilities with Logit Margin (MeanProb-LM) emerges as the most consistently effective approach for noisy label detection across diverse noise scenarios, representing a novel contribution that has not been previously explored. Because label noise detection is inherently an unsupervised task where clean ground-truth labels are unavailable for model or metric selection, robustness across varied noise distributions and data modalities is critical for practical deployment. \new{While specific methods such as Mean-LM and Mean-CPD achieve top results in isolated synthetic settings, MeanProb-LM maintains near-optimal performance across all 17 evaluated scenarios spanning synthetic and real-world datasets.}

A potential concern with evaluating methods at a fixed budget equal to the noise rate is whether it inherently favors continuous ranking methods (such as MeanProb-LM) over discrete detectors (such as Confident Learning variants or CTRL) that require tie breaking when forced onto an arbitrary budget fraction. To ensure that our conclusions are not biased by discretization artifacts, Tables \ref{tab:fnr_at_operating_point_synthetic} and \ref{tab:fnr_at_operating_point_real} compare each discrete reference method directly against MeanProb-LM at the reference method's own natural operating point where no tie breaking is required. These results confirm that MeanProb-LM maintains significant advantages over reference methods at their respective operating points across both synthetic vision noise scenarios and real-world noise datasets, demonstrating that its performance is not an artifact of continuous ranking. Note that these results are measured for a budget equal to the operating point of the reference method; since operating points with lower budgets lead to higher FNRs, the absolute values should not be solely interpreted as method performance. Furthermore, operating points differ across each group of rows and columns, making cross-comparison not meaningful.

\begin{table}[ht]
    \centering
    \begin{adjustbox}{max width=\textwidth}
    \input{Tables/fnr_at_operating_point_synthetic.tex}
    \end{adjustbox}
    \caption{FNR (\%) at the operating point of the reference method on CIFAR-10 and CIFAR-100 with synthetic noise.}
    \label{tab:fnr_at_operating_point_synthetic}
\end{table}

\begin{table}[ht]
    \centering
    \begin{adjustbox}{max width=\textwidth}
    \input{Tables/fnr_at_operating_point_real.tex}
    \end{adjustbox}
    \caption{FNR (\%) at the operating point of the reference method on real-world noise datasets.}
    \label{tab:fnr_at_operating_point_real}
\end{table}

\subsection{Choice of LDM}

The choice of label disagreement measure significantly impacts performance across different noise types. For instance, Logit Margin excels in pairflip scenarios, where its ability to capture decision boundary proximity makes it sensitive to this type of noise pattern. \new{Pairflip noise is sparse and structured, affecting only specific class transitions. Logit Margin, by measuring the gap between the assigned class and the most confident competing class, effectively captures samples that are ``confused'' between these two classes.} This advantage also extends to real-world datasets like CIFAR-10N and Clothing1M, \new{suggesting that real-world noise often exhibits similar sparse structures locally in the feature space, where an instance is more likely to be confused between a few related classes rather than being uniformly distributed across all classes (as in symmetric noise)}. In symmetric noise scenarios, while Logit Margin still generally outperforms Jensen-Shannon, the performance gap is considerably smaller than in pairflip settings\new{, as the noise signal is diffused across all classes, making the margin less discriminative}. Jensen-Shannon also demonstrates strong performance on Food-101N, but without knowledge of the noise pattern, the underlying cause remains unclear. \new{CPD also shows good performance specially in pairflip and real world noise settings, which is expected due to its relationship with Logit Margin.} These findings suggest that aligning the label disagreement measure with the underlying noise structure can yield substantial performance improvements, with the choice being most critical in specific noise contexts like pairflip scenarios \cite{oyen_robustness_2022}.

\subsection{Choice of aggregation method}

Equally important is the aggregation method employed. Mean and Mean Probabilities consistently outperform Last aggregation, with Mean Probabilities demonstrating particular strength in challenging scenarios. These improvements show an asymmetric pattern: when Mean Probabilities outperforms other methods, it delivers substantial gains of 5-10\% improvement in FNR, but when it underperforms relative to other methods, the losses are minimal at typically less than 2\% difference. \new{We also observe a significant performance gap when comparing against Late Stopping. Across all benchmarks, LateStopping-CPD is substantially outperformed by Mean-CPD. Because model predictions can fluctuate during training, locking in the noise score at the first epoch of label agreement makes Late Stopping sensitive to transient predictions, whereas averaging disagreement over the entire trajectory provides a more stable and discriminative signal.}

Surprisingly, more sophisticated methods do not consistently justify their complexity. Despite theoretical advantages, neither SWA nor CTRL provided reliable improvements across settings. \new{For SWA, we start aggregating weights from the first epoch, which means the model average may be significantly ``polluted'' by the high-loss, underfit weights from the early stages of training on vision models. Choosing the starting epoch to aggregate weights is a difficult task, which is out of the scope of this paper. Nevertheless, SWA performs noticeably better on tabular datasets (Table \ref{tab:tabular_fnr}), matching and even outperforming other methods in some cases. This is likely because the tabular MLP architecture is significantly smaller in capacity, and training proceeds for a relatively larger number of epochs, allowing the model to spend a greater proportion of training in a stable loss region where full-trajectory weight averaging yields well-behaved predictions.

In contrast to temporal ensembling, CTRL exhibits highly inconsistent performance across settings. While it achieves top results in isolated cases (CTRL-CPD on CIFAR-10 pairflip and CTRL-LM on several tabular datasets), it fails markedly in others, often performing worse than single-checkpoint baselines. This brittleness likely stems from discretization artifacts inherent in its clustering approach: the method's reliance on partitioning training dynamics into discrete temporal windows produces a discrete vote-based ranking that simpler continuous aggregations avoid. Notably, we implement CTRL's optimization by providing the true noise rate in place of the number of selected clusters, representing a best-case scenario that is unavailable in practice, which makes its inconsistency particularly significant. Even when evaluated at the same operating point as other methods (see Tables \ref{tab:fnr_at_operating_point_synthetic} and \ref{tab:fnr_at_operating_point_real}), CTRL does not consistently outperform simpler aggregations like MeanProb-LM, suggesting that explicitly modeling training dynamics through clustering does not reliably improve upon continuous temporal ensembling.}

\subsection{Choice of gathering strategy}

\new{
Confident learning raises a key question: whether computationally
expensive out-of-sample cross-validation is indispensable, or even beneficial, for noise detection. The main argument for OOS gathering is to prevent the model from memorizing corrupted labels, which would suppress the disagreement signal used for detection. However, when comparing IS and OOS variants of CL methods, the results are inconsistent. For CL-CC, the IS variant matches or outperforms OOS on synthetic noise, while for CL-PBC and CL-PBNR, OOS improves performance in most, although not all, settings. This lack of a clear, uniform benefit calls into question whether the additional computational cost of $K$-fold cross-validation is justified within CL itself.

Moreover, regardless of the gathering strategy, CL variants are
outperformed by simpler single-checkpoint baselines (such as Last-LM) in nearly all scenarios, and by even wider margins when compared to temporal ensembling methods (such as MeanProb-LM). To further isolate the effect of gathering strategy beyond CL, we also evaluate OOS versions of Last-LM and Mean-LM. In both cases, the IS counterpart outperforms the OOS variant in most settings, confirming that the overhead of cross-validation does not translate into detection gains even for non-CL methods. 

While we do not investigate the shortfall of OOS in depth, two possible explanations merit discussion. First, $K$-fold cross-validation inherently reduces the training set size for each fold, which may yield weaker models and thus worse training dynamics. Increasing the number of folds would mitigate this at additional computational cost. Second, predictions produced by different fold models may not be directly comparable: if one model produces systematically more confident predictions, for instance, its scores may dominate the final ranking. Normalizing predictions or ranking samples within each fold could address this, though the latter introduces complications since the noise rate naturally varies across folds.

We also observe three datasets, Cardiotocography, Human Activity, and Mushroom, where the trained models have significantly memorized noisy labels, identifiable by the sharp performance degradation of single-checkpoint methods relative to ensemble methods in Figure \ref{fig:varying_end_epoch}. In these settings, OOS gathering does improve detection for single-checkpoint methods by partially avoiding memorization. However, for ensemble methods, the OOS variants still perform worse than their IS counterparts even in these cases. Moreover, our results show that it is generally possible to stop training before significant memorization occurs, suggesting that this particular vulnerability of IS methods can be avoided in practice.
}




\subsection{Impact of hyperparameter optimization on detection performance}

\begin{figure}
    \centering
    \input{Plots/optimization_curves.tex}
    \caption{FNR at noise rate varying the final epoch of the aggregating window on four datasets, measured on the remaining 90\% of data. The vertical black dotted line indicates the original amount of training epochs.}
    \label{fig:varying_end_epoch}
\end{figure}
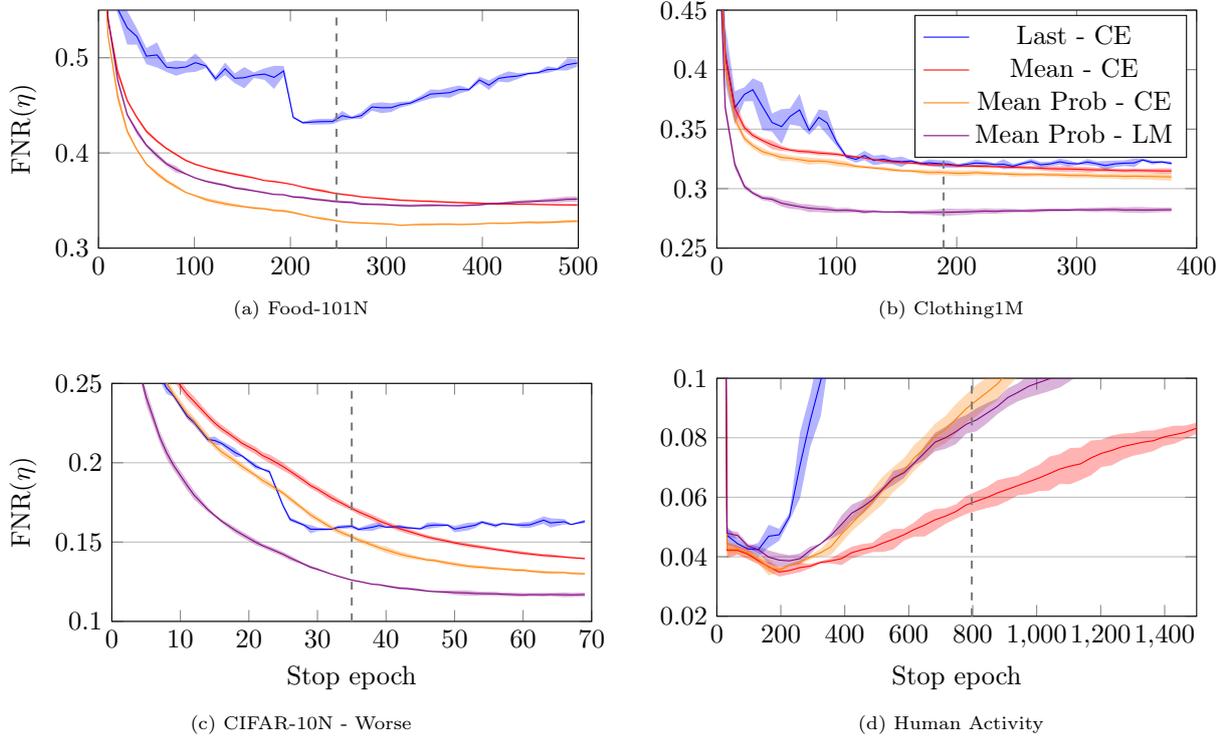

Tables \ref{tab:optimize_cifar_synthetic}, \ref{tab:optimize_real}, and \ref{tab:optimize_tabular} present detection performance after hyperparameter tuning for each method (see \ref{app:detection_hypp} for hyperparameter values). The results demonstrate that hyperparameter optimization has minimal impact on overall performance: across datasets, noise types, and noise rates, tuning rarely alters the relative ranking of methods and produces only marginal improvements in most experimental conditions. The qualitative conclusions drawn from untuned (default) configurations remain valid after optimization.
The few instances of non-negligible improvements are concentrated in Last and CTRL aggregation methods. For Last-CE, we observe that optimization primarily corrects subpar performance in specific datasets by adjusting the final training epoch, suggesting these improvements address artifacts from the early-stopping procedure (see Figure \ref{fig:varying_end_epoch}). \new{For CTRL-CE, optimization yields the most substantial improvements, sometimes halving the FNR (e.g., from 44.52\% to 26.17\% on CIFAR-10N Aggregated). This is partly expected, since other methods were partially optimized by selecting training epochs in our baseline experiments, while CTRL, which depends on additional clustering hyperparameters, benefits disproportionately from tuning. After optimization, CTRL-CE on vision datasets narrows the gap to simpler methods but still does not outperform MeanProb-LM in any vision scenario. On tabular datasets, however, optimized CTRL-CE achieves the best overall result on three of seven datasets (Cardiotocography, Human Activity, and Satellite), confirming the strength already observed in the untuned tabular results. We note that in Cardiotocography and Human Activity the poor performance of untuned Last-CE indicates that the model has suffered significant memorization; this is further confirmed by the fact that optimizing the epoch improves performance drastically. This may indicate that some form of clustering could be even more resilient to memorization than simple ensembling approaches.}

\begin{table}
    \centering
    \begin{adjustbox}{max width=\textwidth, keepaspectratio}
    \input{Tables/optimized_cifar_synthetic}
    \end{adjustbox}
    \caption{Results for optimization on vision datasets with synthetic noise across different held-out fractions for detection method tuning. Every method is evaluated on the remaining 90\% of data. Values in bold indicate the best result for each fraction, while values in both bold and italic represent the best overall across all fractions.}
    \label{tab:optimize_cifar_synthetic}
\end{table}

\begin{table}
    \centering
    \input{Tables/optimized_real}
    \caption{Results for optimization on vision datasets with real noise across different held-out fractions for detection method tuning. Every method is evaluated on the remaining 90\% of data. Values in bold indicate the best result for each fraction, while values in both bold and italic represent the best overall across all fractions.}
    \label{tab:optimize_real}
\end{table}

\begin{table}
    \centering
    \begin{adjustbox}{max width=\textwidth, keepaspectratio}
    \input{Tables/optimized_tabular}
    \end{adjustbox}
    \caption{Results for optimization on tabular datasets with synthetic noise across different held-out fractions for detection method tuning. Every method is evaluated on the remaining 90\% of data. Values in bold indicate the best result for each fraction, while values in both bold and italic represent the best overall across all fractions.}
    \label{tab:optimize_tabular}
\end{table}

\subsection{\new{Practical Considerations and Limitations}}

\new{A key assumption in our benchmark is that detected mislabeled samples are eventually relabeled with their (correct) ground truth. In practice, relabeling is often an imperfect process and may fail to correct errors or even introduce new errors. Nevertheless, our evaluation focuses on the detection task itself motivated by the fact that relabeling quality is bounded by the detection performance and motivating the usage of the FNR as a proxy for the potential cleaning effectiveness.
Assuming that relabeling errors are independent of the detection method, they would affect all methods proportionally, thereby maintaining the relative ranking of the methods established in this benchmark. Consequently, while the absolute gains in downstream classification might vary in real-world applications, we argue that the comparative insights regarding detector performance remain robust.}

%% file: Tables/fnratnr_cifar_synthetic.tex
\begin{tabular}{lcccccc}
\toprule
\textbf{Model} & \multicolumn{6}{c}{\textbf{9-Layer CNN}} \\
\cmidrule(lr){2-7}
\textbf{Dataset} & \multicolumn{3}{c}{\textbf{CIFAR10}} & \multicolumn{3}{c}{\textbf{CIFAR100}} \\
\cmidrule(lr){2-4} \cmidrule(lr){5-7}
\textbf{Noise Type} & \textbf{Pairflip} & \multicolumn{2}{c}{\textbf{Symmetric}} & \textbf{Pairflip} & \multicolumn{2}{c}{\textbf{Symmetric}} \\
\cmidrule(lr){2-2} \cmidrule(lr){3-4} \cmidrule(lr){5-5} \cmidrule(lr){6-7}
\textbf{Noise Rate} & 20\% & 20\% & 50\% & 20\% & 20\% & 50\% \\
\midrule
Last-CE/JS & 13.28 \stdsize{± 0.05} & 9.024 \stdsize{± 0.07} & 8.521 \stdsize{± 0.06} & 49.26 \stdsize{± 0.23} & 12.55 \stdsize{± 0.09} & 9.628 \stdsize{± 0.09} \\
Last-LM & 11.72 \stdsize{± 0.08} & 9.429 \stdsize{± 0.20} & 8.598 \stdsize{± 0.05} & 29.58 \stdsize{± 0.38} & 15.83 \stdsize{± 0.24} & 10.85 \stdsize{± 0.14} \\
Last-LM (OOS) & 29.14 \stdsize{± 0.36} & 10.90 \stdsize{± 0.04} & 9.604 \stdsize{± 0.04} & 47.31 \stdsize{± 0.16} & 18.04 \stdsize{± 0.14} & 12.01 \stdsize{± 0.09} \\
Mean-CE & 15.60 \stdsize{± 0.21} & 7.523 \stdsize{± 0.03} & 5.993 \stdsize{± 0.06} & 47.09 \stdsize{± 0.40} & 9.785 \stdsize{± 0.08} & 8.605 \stdsize{± 0.10} \\
Mean-JS & 10.35 \stdsize{± 0.12} & 7.603 \stdsize{± 0.05} & 5.954 \stdsize{± 0.04} & 43.51 \stdsize{± 0.47} & 9.785 \stdsize{± 0.10} & 8.517 \stdsize{± 0.12} \\
Mean-LM & 8.267 \stdsize{± 0.05} & \bfseries 7.058 \stdsize{± 0.05} & \bfseries 5.639 \stdsize{± 0.05} & 22.15 \stdsize{± 0.29} & \bfseries 8.951 \stdsize{± 0.06} & \bfseries 8.134 \stdsize{± 0.05} \\
Mean-LM (OOS) & 18.74 \stdsize{± 0.08} & 7.882 \stdsize{± 0.02} & 6.321 \stdsize{± 0.03} & 35.46 \stdsize{± 0.07} & 10.64 \stdsize{± 0.06} & 9.790 \stdsize{± 0.04} \\
Mean-CPD & 5.578 \stdsize{± 0.08} & 7.616 \stdsize{± 0.06} & 5.889 \stdsize{± 0.07} & \bfseries 17.66 \stdsize{± 0.33} & 12.53 \stdsize{± 0.04} & 10.52 \stdsize{± 0.06} \\
MeanProb-CE/JS & 8.148 \stdsize{± 0.09} & 7.626 \stdsize{± 0.04} & 5.932 \stdsize{± 0.05} & 41.44 \stdsize{± 0.47} & 9.795 \stdsize{± 0.15} & 8.494 \stdsize{± 0.09} \\
MeanProb-LM & 6.123 \stdsize{± 0.12} & 7.177 \stdsize{± 0.09} & 5.728 \stdsize{± 0.04} & \bfseries 17.67 \stdsize{± 0.29} & 9.459 \stdsize{± 0.11} & 8.751 \stdsize{± 0.04} \\
SWA-CE/JS & 18.59 \stdsize{± 0.25} & 11.11 \stdsize{± 0.03} & 8.137 \stdsize{± 0.08} & 54.92 \stdsize{± 0.00} & 15.11 \stdsize{± 0.28} & 11.77 \stdsize{± 0.03} \\
SWA-LM & 13.53 \stdsize{± 0.11} & 12.13 \stdsize{± 0.07} & 8.423 \stdsize{± 0.07} & 34.85 \stdsize{± 0.56} & 17.14 \stdsize{± 0.21} & 13.45 \stdsize{± 0.07} \\
LateStopping-CPD & 22.01 \stdsize{± 0.15} & 19.15 \stdsize{± 0.23} & 18.95 \stdsize{± 0.07} & 25.83 \stdsize{± 0.28} & 13.26 \stdsize{± 0.23} & 11.86 \stdsize{± 0.06} \\
CTRL-CE & 13.18 \stdsize{± 0.15} & 8.207 \stdsize{± 0.05} & 11.94 \stdsize{± 0.03} & 49.64 \stdsize{± 0.29} & 10.59 \stdsize{± 0.22} & 12.56 \stdsize{± 0.25} \\
CTRL-JS & 7.963 \stdsize{± 0.17} & 8.672 \stdsize{± 0.14} & 6.299 \stdsize{± 0.10} & 56.66 \stdsize{± 0.13} & 52.81 \stdsize{± 0.21} & 12.32 \stdsize{± 0.12} \\
CTRL-LM & 21.97 \stdsize{± 0.09} & 18.82 \stdsize{± 0.02} & 6.791 \stdsize{± 0.04} & 41.79 \stdsize{± 0.32} & 12.18 \stdsize{± 0.24} & 11.06 \stdsize{± 0.10} \\
CTRL-CPD & \bfseries 5.297 \stdsize{± 0.11} & 10.46 \stdsize{± 0.15} & 5.984 \stdsize{± 0.05} & 29.45 \stdsize{± 0.43} & 40.90 \stdsize{± 0.31} & 16.54 \stdsize{± 0.16} \\
CL-CC (IS) & 10.85 \stdsize{± 0.18} & 8.559 \stdsize{± 0.09} & 7.790 \stdsize{± 0.04} & 38.83 \stdsize{± 0.84} & 11.73 \stdsize{± 0.05} & 9.278 \stdsize{± 0.06} \\
CL-CC (OOS) & 33.14 \stdsize{± 0.42} & 9.326 \stdsize{± 0.07} & 9.270 \stdsize{± 0.04} & 63.12 \stdsize{± 0.28} & 13.74 \stdsize{± 0.13} & 10.25 \stdsize{± 0.08} \\
CL-CYY (IS) & 25.92 \stdsize{± 0.24} & 34.77 \stdsize{± 0.40} & 20.49 \stdsize{± 0.29} & 49.62 \stdsize{± 0.45} & 63.75 \stdsize{± 0.55} & 35.63 \stdsize{± 0.48} \\
CL-CYY (OOS) & 23.14 \stdsize{± 0.46} & 23.79 \stdsize{± 0.19} & 12.44 \stdsize{± 0.03} & 44.55 \stdsize{± 0.19} & 39.63 \stdsize{± 0.28} & 18.37 \stdsize{± 0.30} \\
CL-PBC (IS) & 21.09 \stdsize{± 0.21} & 18.28 \stdsize{± 0.26} & 9.867 \stdsize{± 0.10} & 51.43 \stdsize{± 0.10} & 25.30 \stdsize{± 0.85} & 9.710 \stdsize{± 0.18} \\
CL-PBC (OOS) & 33.14 \stdsize{± 0.42} & 9.333 \stdsize{± 0.08} & 9.270 \stdsize{± 0.04} & 63.13 \stdsize{± 0.26} & 13.73 \stdsize{± 0.12} & 10.25 \stdsize{± 0.08} \\
CL-PBNR (IS) & 16.86 \stdsize{± 0.41} & 20.53 \stdsize{± 0.21} & 14.42 \stdsize{± 0.25} & 29.64 \stdsize{± 0.25} & 52.27 \stdsize{± 0.71} & 30.84 \stdsize{± 0.51} \\
CL-PBNR (OOS) & 23.75 \stdsize{± 0.47} & 14.95 \stdsize{± 0.19} & 10.85 \stdsize{± 0.06} & 37.68 \stdsize{± 0.09} & 38.22 \stdsize{± 0.08} & 26.44 \stdsize{± 0.28} \\
\bottomrule
\end{tabular}

%% file: Tables/fnratnr_real.tex
\begin{tabular}{lcccc}
\toprule
\textbf{Model} & \multicolumn{2}{c}{\textbf{9-Layer CNN}} & \textbf{ResNet-18} & \textbf{ResNet-34} \\
\cmidrule(lr){2-3} \cmidrule(lr){4-4} \cmidrule(lr){5-5}
\textbf{Dataset} & \multicolumn{2}{c}{\textbf{CIFAR10N}} & \textbf{Clothing1M} & \textbf{Food-101N} \\
\cmidrule(lr){2-3} \cmidrule(lr){4-4} \cmidrule(lr){5-5}
\textbf{Noise Type} & \textbf{Aggre} & \textbf{Worse} & \textbf{Real} & \textbf{Real} \\
\cmidrule(lr){2-2} \cmidrule(lr){3-3} \cmidrule(lr){4-4} \cmidrule(lr){5-5}
\textbf{Noise Rate} & 9.01\% & 40.21\% & 38.26\% & 18.51\% \\
\midrule
Last-CE/JS & 33.86 \stdsize{± 0.18} & 15.79 \stdsize{± 0.14} & 32.26 \stdsize{± 0.05} & 43.81 \stdsize{± 0.31} \\
Last-LM & 33.40 \stdsize{± 0.13} & 14.99 \stdsize{± 0.08} & 30.52 \stdsize{± 0.04} & 46.16 \stdsize{± 0.50} \\
Last-LM (OOS) & 29.89 \stdsize{± 0.20} & 15.68 \stdsize{± 0.03} & 30.53 \stdsize{± 0.12} & 47.17 \stdsize{± 0.36} \\
Mean-CE & 26.34 \stdsize{± 0.34} & 17.18 \stdsize{± 0.03} & 32.10 \stdsize{± 0.03} & 36.16 \stdsize{± 0.13} \\
Mean-JS & 25.60 \stdsize{± 0.25} & 16.15 \stdsize{± 0.06} & 31.89 \stdsize{± 0.04} & 33.79 \stdsize{± 0.10} \\
Mean-LM & 23.85 \stdsize{± 0.12} & 13.12 \stdsize{± 0.03} & 29.07 \stdsize{± 0.10} & 37.73 \stdsize{± 0.13} \\
Mean-LM (OOS) & 25.68 \stdsize{± 0.08} & 14.86 \stdsize{± 0.11} & 29.43 \stdsize{± 0.06} & 39.18 \stdsize{± 0.04} \\
Mean-CPD & 23.77 \stdsize{± 0.06} & 12.84 \stdsize{± 0.07} & 29.60 \stdsize{± 0.12} & 33.42 \stdsize{± 0.14} \\
MeanProb-CE/JS & 25.16 \stdsize{± 0.20} & 15.39 \stdsize{± 0.02} & 31.61 \stdsize{± 0.06} & \bfseries 33.08 \stdsize{± 0.14} \\
MeanProb-LM & \bfseries 23.39 \stdsize{± 0.24} & \bfseries 12.61 \stdsize{± 0.04} & \bfseries 28.38 \stdsize{± 0.15} & 35.07 \stdsize{± 0.02} \\
SWA-CE/JS & 32.39 \stdsize{± 0.27} & 17.48 \stdsize{± 0.24} & 31.62 \stdsize{± 0.45} & 44.71 \stdsize{± 0.15} \\
SWA-LM & 31.87 \stdsize{± 0.36} & 15.62 \stdsize{± 0.15} & 29.79 \stdsize{± 0.46} & 47.03 \stdsize{± 0.19} \\
LateStopping-CPD & 47.10 \stdsize{± 0.03} & 29.99 \stdsize{± 0.06} & 44.93 \stdsize{± 1.75} & 42.76 \stdsize{± 0.30} \\
CTRL-CE & 44.45 \stdsize{± 0.24} & 38.13 \stdsize{± 0.08} & 44.19 \stdsize{± 0.12} & 40.14 \stdsize{± 0.12} \\
CTRL-JS & 30.06 \stdsize{± 0.97} & 27.83 \stdsize{± 0.11} & 39.43 \stdsize{± 0.20} & 49.24 \stdsize{± 0.11} \\
CTRL-LM & 32.62 \stdsize{± 0.51} & 33.25 \stdsize{± 0.03} & 42.85 \stdsize{± 0.11} & 43.91 \stdsize{± 0.09} \\
CTRL-CPD & 32.54 \stdsize{± 0.63} & 14.65 \stdsize{± 0.06} & 33.02 \stdsize{± 0.11} & 45.36 \stdsize{± 0.21} \\
CL-CC (IS) & 37.99 \stdsize{± 0.31} & 15.86 \stdsize{± 0.40} & 33.99 \stdsize{± 0.10} & 43.96 \stdsize{± 0.04} \\
CL-CC (OOS) & 29.53 \stdsize{± 0.19} & 16.92 \stdsize{± 0.07} & 33.11 \stdsize{± 0.08} & 44.47 \stdsize{± 0.25} \\
CL-CYY (IS) & 80.43 \stdsize{± 0.28} & 34.52 \stdsize{± 0.26} & 42.99 \stdsize{± 0.10} & 70.45 \stdsize{± 0.17} \\
CL-CYY (OOS) & 52.46 \stdsize{± 0.26} & 27.79 \stdsize{± 0.24} & 41.60 \stdsize{± 0.05} & 66.28 \stdsize{± 0.41} \\
CL-PBC (IS) & 76.52 \stdsize{± 0.77} & 23.97 \stdsize{± 0.12} & 39.35 \stdsize{± 0.38} & 50.02 \stdsize{± 0.50} \\
CL-PBC (OOS) & 37.37 \stdsize{± 0.18} & 16.50 \stdsize{± 0.27} & 38.43 \stdsize{± 0.16} & 44.67 \stdsize{± 0.27} \\
CL-PBNR (IS) & 77.03 \stdsize{± 0.67} & 27.07 \stdsize{± 0.21} & 39.50 \stdsize{± 0.12} & 62.44 \stdsize{± 0.45} \\
CL-PBNR (OOS) & 41.59 \stdsize{± 0.26} & 20.51 \stdsize{± 0.26} & 38.31 \stdsize{± 0.06} & 60.10 \stdsize{± 0.25} \\
\bottomrule
\end{tabular}

%% file: Tables/fnratnr_tabular.tex
\begin{tabular}{lccccccc}
\toprule
\textbf{Model} & \multicolumn{7}{c}{\textbf{MLP}} \\
\cmidrule(lr){2-8}
\textbf{Dataset} & \textbf{Cardiotocography} & \textbf{Credit Fraud} & \textbf{Human Activity} & \textbf{Letter} & \textbf{Mushroom} & \textbf{Satellite} & \textbf{Sensorless} \\
\cmidrule(lr){2-2} \cmidrule(lr){3-3} \cmidrule(lr){4-4} \cmidrule(lr){5-5} \cmidrule(lr){6-6} \cmidrule(lr){7-7} \cmidrule(lr){8-8}
\textbf{Noise Type} & \textbf{Symmetric} & \textbf{Symmetric} & \textbf{Symmetric} & \textbf{Symmetric} & \textbf{Symmetric} & \textbf{Symmetric} & \textbf{Symmetric} \\
\cmidrule(lr){2-2} \cmidrule(lr){3-3} \cmidrule(lr){4-4} \cmidrule(lr){5-5} \cmidrule(lr){6-6} \cmidrule(lr){7-7} \cmidrule(lr){8-8}
\textbf{Noise Rate} & 20\% & 20\% & 20\% & 20\% & 20\% & 20\% & 20\% \\
\midrule
Last-CE/JS & 45.35 \stdsize{± 3.85} & 13.57 \stdsize{± 1.11} & 33.87 \stdsize{± 0.59} & 9.133 \stdsize{± 0.28} & 2.707 \stdsize{± 0.33} & 10.47 \stdsize{± 0.66} & 0.6902 \stdsize{± 0.016} \\
Last-LM & 45.71 \stdsize{± 4.85} & 13.57 \stdsize{± 1.11} & 31.80 \stdsize{± 0.67} & 8.331 \stdsize{± 0.17} & 2.707 \stdsize{± 0.33} & 10.43 \stdsize{± 0.30} & 0.6487 \stdsize{± 0.019} \\
Last-LM (OOS) & 42.37 \stdsize{± 0.34} & 35.90 \stdsize{± 5.38} & 25.71 \stdsize{± 2.65} & 15.98 \stdsize{± 0.46} & 0.9415 \stdsize{± 0.181} & 11.82 \stdsize{± 0.38} & 18.96 \stdsize{± 1.21} \\
Mean-CE & 32.07 \stdsize{± 1.35} & 11.61 \stdsize{± 0.56} & 5.484 \stdsize{± 0.24} & 3.411 \stdsize{± 0.23} & 0.1765 \stdsize{± 0.072} & 10.32 \stdsize{± 0.05} & 0.6374 \stdsize{± 0.005} \\
Mean-JS & 31.71 \stdsize{± 1.81} & 11.76 \stdsize{± 0.74} & 6.906 \stdsize{± 0.51} & 3.978 \stdsize{± 0.14} & 0.1765 \stdsize{± 0.072} & 10.32 \stdsize{± 0.05} & 0.6487 \stdsize{± 0.005} \\
Mean-LM & 31.07 \stdsize{± 0.26} & \bfseries 11.46 \stdsize{± 0.43} & 7.741 \stdsize{± 0.23} & 3.155 \stdsize{± 0.20} & 0.2059 \stdsize{± 0.042} & 10.66 \stdsize{± 0.47} & 0.5997 \stdsize{± 0.009} \\
Mean-LM (OOS) & 33.06 \stdsize{± 1.23} & 25.34 \stdsize{± 6.57} & 14.51 \stdsize{± 1.71} & 7.005 \stdsize{± 0.24} & 1.736 \stdsize{± 0.58} & 10.24 \stdsize{± 0.24} & 4.703 \stdsize{± 0.27} \\
Mean-CPD & 29.45 \stdsize{± 0.71} & 11.76 \stdsize{± 0.37} & 6.274 \stdsize{± 0.41} & 4.213 \stdsize{± 0.21} & 0.1765 \stdsize{± 0.125} & 15.28 \stdsize{± 0.71} & 0.6374 \stdsize{± 0.005} \\
MeanProb-CE/JS & 31.62 \stdsize{± 2.01} & \bfseries 11.46 \stdsize{± 0.43} & 8.914 \stdsize{± 0.39} & 4.278 \stdsize{± 0.17} & 0.2059 \stdsize{± 0.042} & 10.32 \stdsize{± 0.05} & 0.6525 \stdsize{± 0.005} \\
MeanProb-LM & 30.62 \stdsize{± 0.66} & \bfseries 11.46 \stdsize{± 0.43} & 8.283 \stdsize{± 0.52} & 3.668 \stdsize{± 0.08} & 0.2059 \stdsize{± 0.042} & 10.70 \stdsize{± 0.39} & \bfseries 0.5959 \stdsize{± 0.019} \\
SWA-CE/JS & 28.46 \stdsize{± 1.23} & \bfseries 11.46 \stdsize{± 0.43} & 5.168 \stdsize{± 0.19} & 3.262 \stdsize{± 0.17} & 0.1765 \stdsize{± 0.072} & 11.43 \stdsize{± 0.52} & 0.8260 \stdsize{± 0.061} \\
SWA-LM & 28.91 \stdsize{± 0.68} & \bfseries 11.46 \stdsize{± 0.43} & 5.484 \stdsize{± 0.15} & \bfseries 3.080 \stdsize{± 0.09} & 0.1765 \stdsize{± 0.072} & 11.59 \stdsize{± 0.29} & 0.7581 \stdsize{± 0.042} \\
LateStopping-CPD & 71.64 \stdsize{± 13.07} & 46.30 \stdsize{± 17.29} & 22.68 \stdsize{± 2.39} & 8.384 \stdsize{± 0.08} & 43.72 \stdsize{± 5.73} & 27.75 \stdsize{± 3.23} & 7.321 \stdsize{± 0.01} \\
CTRL-CE & 23.49 \stdsize{± 1.11} & 12.67 \stdsize{± 0.74} & \bfseries 3.949 \stdsize{± 0.03} & 5.411 \stdsize{± 0.08} & 0.5296 \stdsize{± 0.072} & 9.584 \stdsize{± 0.25} & 0.9542 \stdsize{± 0.042} \\
CTRL-JS & 23.49 \stdsize{± 4.11} & 13.73 \stdsize{± 1.13} & 5.710 \stdsize{± 0.22} & 4.438 \stdsize{± 0.24} & 0.4707 \stdsize{± 0.042} & 9.931 \stdsize{± 0.25} & 1.256 \stdsize{± 0.02} \\
CTRL-LM & \bfseries 20.87 \stdsize{± 1.81} & 12.07 \stdsize{± 0.21} & 7.425 \stdsize{± 0.61} & 3.251 \stdsize{± 0.24} & \bfseries 0.1471 \stdsize{± 0.083} & \bfseries 9.469 \stdsize{± 0.34} & 1.343 \stdsize{± 0.04} \\
CTRL-CPD & 29.81 \stdsize{± 1.01} & 12.07 \stdsize{± 0.21} & 6.658 \stdsize{± 0.61} & 4.770 \stdsize{± 0.27} & 0.2648 \stdsize{± 0.072} & 23.90 \stdsize{± 0.52} & 2.591 \stdsize{± 0.03} \\
CL-CC (IS) & 43.81 \stdsize{± 3.73} & 13.12 \stdsize{± 1.11} & 57.39 \stdsize{± 2.38} & 9.967 \stdsize{± 0.22} & 5.119 \stdsize{± 0.31} & 10.59 \stdsize{± 0.48} & 0.6412 \stdsize{± 0.014} \\
CL-CC (OOS) & 40.65 \stdsize{± 0.44} & 35.90 \stdsize{± 5.38} & 25.57 \stdsize{± 2.66} & 15.81 \stdsize{± 0.33} & 0.9415 \stdsize{± 0.181} & 11.74 \stdsize{± 0.68} & 19.14 \stdsize{± 1.20} \\
CL-CYY (IS) & 55.92 \stdsize{± 5.98} & 28.51 \stdsize{± 7.63} & 78.31 \stdsize{± 1.44} & 51.34 \stdsize{± 0.32} & 41.98 \stdsize{± 0.22} & 20.55 \stdsize{± 0.84} & 10.25 \stdsize{± 0.44} \\
CL-CYY (OOS) & 41.01 \stdsize{± 0.68} & 36.65 \stdsize{± 6.28} & 26.47 \stdsize{± 2.31} & 20.48 \stdsize{± 0.23} & 4.737 \stdsize{± 0.71} & 17.13 \stdsize{± 1.14} & 17.87 \stdsize{± 0.60} \\
CL-PBC (IS) & 45.89 \stdsize{± 5.43} & 18.85 \stdsize{± 7.26} & 77.77 \stdsize{± 0.33} & 40.95 \stdsize{± 0.90} & 32.33 \stdsize{± 0.27} & 11.16 \stdsize{± 0.43} & 3.398 \stdsize{± 0.26} \\
CL-PBC (OOS) & 40.65 \stdsize{± 0.44} & 35.90 \stdsize{± 5.38} & 25.57 \stdsize{± 2.66} & 15.81 \stdsize{± 0.33} & 1.442 \stdsize{± 0.11} & 11.74 \stdsize{± 0.68} & 19.14 \stdsize{± 1.20} \\
CL-PBNR (IS) & 47.15 \stdsize{± 5.65} & 18.85 \stdsize{± 7.26} & 77.70 \stdsize{± 0.83} & 41.39 \stdsize{± 0.85} & 32.33 \stdsize{± 0.27} & 12.78 \stdsize{± 0.14} & 4.669 \stdsize{± 0.26} \\
CL-PBNR (OOS) & 44.17 \stdsize{± 1.97} & 35.90 \stdsize{± 5.38} & 29.68 \stdsize{± 1.52} & 27.99 \stdsize{± 0.69} & 1.442 \stdsize{± 0.11} & 14.51 \stdsize{± 0.43} & 20.12 \stdsize{± 1.00} \\
\bottomrule
\end{tabular}

%% file: Tables/fnr_at_operating_point_synthetic.tex
\begin{tabular}{lcccccc}

\toprule

\textbf{Model} & \multicolumn{6}{c}{\textbf{9-Layer CNN}} \\
\cmidrule(lr){2-7}
\textbf{Dataset} & \multicolumn{3}{c}{\textbf{CIFAR-10}} & \multicolumn{3}{c}{\textbf{CIFAR-100}} \\
\cmidrule(lr){2-4} \cmidrule(lr){5-7}
\textbf{Noise type} & \textbf{Pairflip} & \multicolumn{2}{c}{\textbf{Symmetric}} & \textbf{Pairflip} & \multicolumn{2}{c}{\textbf{Symmetric}} \\
\cmidrule(lr){2-2} \cmidrule(lr){3-4} \cmidrule(lr){5-5} \cmidrule(lr){6-7}
\textbf{Noise rate} & 20.00\% & 20.00\% & 50.00\% & 20.00\% & 20.00\% & 50.00\% \\
\midrule
CTRL-CE & 5.48 \stdsize{± 0.19} & 8.56 \stdsize{± 0.0094} & 13.13 \stdsize{± 0.056} & 51.73 \stdsize{± 0.51} & 8.96 \stdsize{± 0.11} & \bfseries 13.70 \stdsize{± 0.34} \\
MeanProb-LM & \bfseries 3.37 \stdsize{± 0.16} & \bfseries 7.69 \stdsize{± 0.11} & \bfseries 13.10 \stdsize{± 0.12} & \bfseries 20.93 \stdsize{± 0.52} & \bfseries 8.72 \stdsize{± 0.13} & 14.42 \stdsize{± 0.26} \\
\midrule
CL-CYY (OOS) & 15.56 \stdsize{± 0.36} & 24.48 \stdsize{± 0.19} & 11.58 \stdsize{± 0.084} & 40.17 \stdsize{± 0.059} & 39.29 \stdsize{± 0.32} & 16.35 \stdsize{± 0.35} \\
MeanProb-LM & \bfseries 3.25 \stdsize{± 0.16} & \bfseries 14.05 \stdsize{± 0.46} & \bfseries 3.85 \stdsize{± 0.14} & \bfseries 11.68 \stdsize{± 0.54} & \bfseries 3.39 \stdsize{± 0.029} & \bfseries 2.05 \stdsize{± 0.085} \\
\midrule
CL-PBC (OOS) & 13.63 \stdsize{± 0.26} & 4.68 \stdsize{± 0.2} & 3.46 \stdsize{± 0.025} & 30.95 \stdsize{± 0.41} & 1.63 \stdsize{± 0.055} & 0.97 \stdsize{± 0.018} \\
MeanProb-LM & \bfseries 1.54 \stdsize{± 0.066} & \bfseries 2.75 \stdsize{± 0.08} & \bfseries 1.47 \stdsize{± 0.046} & \bfseries 1.70 \stdsize{± 0.15} & \bfseries 0.93 \stdsize{± 0.033} & \bfseries 0.65 \stdsize{± 0.0094} \\
\midrule
CL-PBNR (OOS) & 7.29 \stdsize{± 0.26} & 8.70 \stdsize{± 0.08} & 9.57 \stdsize{± 0.086} & 16.70 \stdsize{± 0.31} & 25.22 \stdsize{± 0.1} & 26.89 \stdsize{± 0.34} \\
MeanProb-LM & \bfseries 1.81 \stdsize{± 0.061} & \bfseries 3.23 \stdsize{± 0.034} & \bfseries 4.53 \stdsize{± 0.072} & \bfseries 2.90 \stdsize{± 0.15} & \bfseries 1.84 \stdsize{± 0.069} & \bfseries 9.71 \stdsize{± 0.14} \\
\bottomrule
\end{tabular}

%% file: Tables/fnr_at_operating_point_real.tex
\begin{tabular}{lcccc}

\toprule

\textbf{Model} & \multicolumn{2}{c}{\textbf{9-Layer CNN}} & \textbf{ResNet-18} & \textbf{ResNet-34} \\
\cmidrule(lr){2-3} \cmidrule(lr){4-4} \cmidrule(lr){5-5} 
\textbf{Dataset} & \multicolumn{2}{c}{\textbf{CIFAR-10N}} & \textbf{Clothing1M} & \textbf{Food-101N} \\
\cmidrule(lr){2-3} \cmidrule(lr){4-4} \cmidrule(lr){5-5} 
\textbf{Noise type} & \textbf{Aggregate} & \textbf{Worse} & \textbf{Real} & \textbf{Real} \\
\cmidrule(lr){2-2} \cmidrule(lr){3-3} \cmidrule(lr){4-4} \cmidrule(lr){5-5} 
\textbf{Noise rate} & 9.01\% & 40.21\% & 38.26\% & 18.51\% \\
\midrule
CTRL-CE & 13.69 \stdsize{± 0.46} & 50.21 \stdsize{± 0.084} & 57.44 \stdsize{± 0.11} & 29.71 \stdsize{± 0.099} \\
MeanProb-LM & \bfseries 11.06 \stdsize{± 0.2} & \bfseries 48.39 \stdsize{± 0.14} & \bfseries 54.39 \stdsize{± 0.14} & \bfseries 29.04 \stdsize{± 0.12} \\
\midrule
CL-CYY (OOS) & 54.61 \stdsize{± 0.33} & 31.43 \stdsize{± 0.33} & 52.06 \stdsize{± 0.15} & 68.13 \stdsize{± 0.37} \\
MeanProb-LM & \bfseries 44.58 \stdsize{± 0.19} & \bfseries 24.85 \stdsize{± 0.44} & \bfseries 50.05 \stdsize{± 0.097} & \bfseries 40.49 \stdsize{± 0.16} \\
\midrule
CL-PBC (OOS) & 37.85 \stdsize{± 0.18} & 16.75 \stdsize{± 0.31} & 44.54 \stdsize{± 0.1} & 33.62 \stdsize{± 0.35} \\
MeanProb-LM & \bfseries 30.59 \stdsize{± 0.34} & \bfseries 13.75 \stdsize{± 0.26} & \bfseries 41.67 \stdsize{± 0.17} & \bfseries 23.91 \stdsize{± 0.073} \\
\midrule
CL-PBNR (OOS) & 42.18 \stdsize{± 0.27} & 21.80 \stdsize{± 0.25} & 45.74 \stdsize{± 0.18} & 52.33 \stdsize{± 0.46} \\
MeanProb-LM & \bfseries 30.60 \stdsize{± 0.33} & \bfseries 17.94 \stdsize{± 0.26} & \bfseries 44.35 \stdsize{± 0.13} & \bfseries 27.28 \stdsize{± 0.074} \\
\bottomrule
\end{tabular}

%% file: Tables/optimized_cifar_synthetic.tex
\begin{tabular}{llcccccc}

\toprule

 & \textbf{Model} & \multicolumn{6}{c}{\textbf{9-Layer CNN}} \\
\cmidrule(lr){2-2} \cmidrule(lr){3-8} 
 & \textbf{Dataset} & \multicolumn{3}{c}{\textbf{CIFAR-10}} & \multicolumn{3}{c}{\textbf{CIFAR-100}} \\
\cmidrule(lr){2-2} \cmidrule(lr){3-5} \cmidrule(lr){6-8} 
 & \textbf{Noise type} & \textbf{Pairflip} & \multicolumn{2}{c}{\textbf{Symmetric}} & \textbf{Pairflip} & \multicolumn{2}{c}{\textbf{Symmetric}} \\
\cmidrule(lr){2-2} \cmidrule(lr){3-3} \cmidrule(lr){4-5} \cmidrule(lr){6-6} \cmidrule(lr){7-8} 
 & \textbf{Noise rate} & \textbf{20.00\%} & \textbf{20.00\%} & \textbf{50.00\%} & \textbf{20.00\%} & \textbf{20.00\%} & \textbf{50.00\%} \\
\cmidrule(lr){2-2} \cmidrule(lr){3-3} \cmidrule(lr){4-4} \cmidrule(lr){5-5} \cmidrule(lr){6-6} \cmidrule(lr){7-7} \cmidrule(lr){8-8} 
Fraction & Method &  &  &  &  &  &  \\
\midrule
\multirow[c]{4}{*}{0\%} & Last-CE/JS & 13.25 ± \stdsize{0.18} & 9.25 ± \stdsize{0.19} & 8.51 ± \stdsize{0.05} & 49.17 ± \stdsize{0.12} & 12.49 ± \stdsize{0.17} & 9.63 ± \stdsize{0.08} \\
 & Mean-CE & 15.90 ± \stdsize{0.25} & 7.57 ± \stdsize{0.04} & 5.97 ± \stdsize{0.06} & 46.93 ± \stdsize{0.33} & 9.80 ± \stdsize{0.09} & \bfseries 8.58 ± \stdsize{0.11} \\
 & MeanProb-LM & \bfseries 6.10 ± \stdsize{0.08} & \bfseries 7.21 ± \stdsize{0.06} & \bfseries 5.71 ± \stdsize{0.04} & \bfseries 17.60 ± \stdsize{0.24} & \bfseries 9.47 ± \stdsize{0.11} & 8.74 ± \stdsize{0.07} \\
 & CTRL-CE & 13.20 ± \stdsize{0.13} & 8.16 ± \stdsize{0.07} & 12.02 ± \stdsize{0.02} & 49.41 ± \stdsize{0.30} & 10.56 ± \stdsize{0.22} & 12.53 ± \stdsize{0.27} \\
\midrule 
\multirow[c]{4}{*}{1\%} & Last-CE/JS & 13.24 ± \stdsize{0.12} & 9.60 ± \stdsize{0.29} & 8.27 ± \stdsize{0.04} & 47.93 ± \stdsize{0.35} & 12.49 ± \stdsize{0.17} & 9.63 ± \stdsize{0.08} \\
 & Mean-CE & 7.77 ± \stdsize{0.23} & 6.99 ± \stdsize{0.09} & 5.45 ± \stdsize{0.02} & 41.63 ± \stdsize{0.46} & 10.09 ± \stdsize{0.04} & \bfseries 7.71 ± \stdsize{0.03} \\
 & MeanProb-LM & \underline{\bfseries 5.12 ± \stdsize{0.14}} & \bfseries 5.86 ± \stdsize{0.09} & \bfseries 5.37 ± \stdsize{0.02} & \bfseries 18.26 ± \stdsize{0.32} & \bfseries 9.90 ± \stdsize{0.01} & 8.44 ± \stdsize{0.07} \\
 & CTRL-CE & 13.20 ± \stdsize{0.13} & 8.16 ± \stdsize{0.07} & 6.34 ± \stdsize{0.02} & 39.28 ± \stdsize{0.36} & 10.56 ± \stdsize{0.22} & 9.58 ± \stdsize{0.07} \\
\midrule 
\multirow[c]{4}{*}{5\%} & Last-CE/JS & 13.24 ± \stdsize{0.12} & 9.25 ± \stdsize{0.19} & 8.32 ± \stdsize{0.04} & 48.06 ± \stdsize{0.34} & 12.73 ± \stdsize{0.16} & 9.48 ± \stdsize{0.02} \\
 & Mean-CE & 7.48 ± \stdsize{0.27} & 6.35 ± \stdsize{0.11} & 5.43 ± \stdsize{0.05} & 40.86 ± \stdsize{0.43} & \bfseries 9.40 ± \stdsize{0.11} & \bfseries 7.74 ± \stdsize{0.03} \\
 & MeanProb-LM & \bfseries 5.16 ± \stdsize{0.15} & \underline{\bfseries 5.85 ± \stdsize{0.07}} & \bfseries 5.30 ± \stdsize{0.03} & \bfseries 16.70 ± \stdsize{0.31} & 9.64 ± \stdsize{0.14} & 8.16 ± \stdsize{0.04} \\
 & CTRL-CE & 13.20 ± \stdsize{0.13} & 8.17 ± \stdsize{0.02} & 6.34 ± \stdsize{0.02} & 39.28 ± \stdsize{0.36} & 10.22 ± \stdsize{0.16} & 9.58 ± \stdsize{0.07} \\
\midrule 
\multirow[c]{4}{*}{10\%} & Last-CE/JS & 13.24 ± \stdsize{0.12} & 8.98 ± \stdsize{0.12} & 8.32 ± \stdsize{0.04} & 48.32 ± \stdsize{0.33} & 12.49 ± \stdsize{0.13} & 9.48 ± \stdsize{0.02} \\
 & Mean-CE & 7.55 ± \stdsize{0.23} & 6.21 ± \stdsize{0.08} & 5.39 ± \stdsize{0.04} & 40.33 ± \stdsize{0.50} & \underline{\bfseries 9.36 ± \stdsize{0.16}} & \underline{\bfseries 7.69 ± \stdsize{0.03}} \\
 & MeanProb-LM & \bfseries 5.19 ± \stdsize{0.15} & \bfseries 5.87 ± \stdsize{0.09} & \underline{\bfseries 5.27 ± \stdsize{0.04}} & \underline{\bfseries 16.59 ± \stdsize{0.26}} & 9.44 ± \stdsize{0.15} & 8.12 ± \stdsize{0.07} \\
 & CTRL-CE & 13.20 ± \stdsize{0.13} & 8.17 ± \stdsize{0.02} & 6.34 ± \stdsize{0.02} & 39.28 ± \stdsize{0.36} & 10.22 ± \stdsize{0.16} & 9.58 ± \stdsize{0.07} \\
\bottomrule
\end{tabular}

%% file: Tables/optimized_real.tex
\begin{tabular}{llcccc}

\toprule

 & \textbf{Model} & \multicolumn{2}{c}{\textbf{9-Layer CNN}} & \textbf{ResNet-18} & \textbf{ResNet-34} \\
\cmidrule(lr){2-2} \cmidrule(lr){3-4} \cmidrule(lr){5-5} \cmidrule(lr){6-6} 
 & \textbf{Dataset} & \multicolumn{2}{c}{\textbf{CIFAR10N}} & \textbf{Clothing1M} & \textbf{Food-101N} \\
\cmidrule(lr){2-2} \cmidrule(lr){3-4} \cmidrule(lr){5-5} \cmidrule(lr){6-6} 
 & \textbf{Noise type} & \textbf{Aggregate} & \textbf{Worse} & \textbf{Real} & \textbf{Real} \\
\cmidrule(lr){2-2} \cmidrule(lr){3-3} \cmidrule(lr){4-4} \cmidrule(lr){5-5} \cmidrule(lr){6-6} 
 & \textbf{Noise rate} & \textbf{9.01\%} & \textbf{40.21\%} & \textbf{38.26\%} & \textbf{18.51\%} \\
\cmidrule(lr){2-2} \cmidrule(lr){3-3} \cmidrule(lr){4-4} \cmidrule(lr){5-5} \cmidrule(lr){6-6} 
Fraction & Method &  &  &  &  \\
\midrule
\multirow[c]{4}{*}{0\%} & Last-CE/JS & 33.92 ± \stdsize{0.46} & 15.77 ± \stdsize{0.15} & 32.47 ± \stdsize{0.01} & 44.25 ± \stdsize{0.31} \\
 & Mean-CE & 26.24 ± \stdsize{0.23} & 17.12 ± \stdsize{0.04} & 32.24 ± \stdsize{0.07} & 36.07 ± \stdsize{0.14} \\
 & MeanProb-LM & \bfseries 23.34 ± \stdsize{0.12} & \bfseries 12.55 ± \stdsize{0.04} & \bfseries 28.50 ± \stdsize{0.15} & \bfseries 35.06 ± \stdsize{0.06} \\
 & CTRL-CE & 44.52 ± \stdsize{0.31} & 38.22 ± \stdsize{0.13} & 44.01 ± \stdsize{0.15} & 40.19 ± \stdsize{0.14} \\
\midrule 
\multirow[c]{4}{*}{1\%} & Last-CE/JS & 33.19 ± \stdsize{0.13} & 15.98 ± \stdsize{0.09} & 32.85 ± \stdsize{0.09} & 44.70 ± \stdsize{0.22} \\
 & Mean-CE & 25.07 ± \stdsize{0.20} & 12.93 ± \stdsize{0.09} & 32.23 ± \stdsize{0.12} & 37.24 ± \stdsize{0.28} \\
 & MeanProb-LM & \bfseries 22.92 ± \stdsize{0.06} & \bfseries 12.15 ± \stdsize{0.07} & \bfseries 28.94 ± \stdsize{0.05} & \bfseries 35.85 ± \stdsize{0.09} \\
 & CTRL-CE & 26.17 ± \stdsize{0.39} & 15.98 ± \stdsize{0.08} & 34.69 ± \stdsize{0.27} & 38.29 ± \stdsize{0.16} \\
\midrule 
\multirow[c]{4}{*}{5\%} & Last-CE/JS & 29.73 ± \stdsize{0.15} & 15.85 ± \stdsize{0.16} & 32.37 ± \stdsize{0.08} & 44.25 ± \stdsize{0.31} \\
 & Mean-CE & 24.10 ± \stdsize{0.15} & 12.83 ± \stdsize{0.05} & 32.39 ± \stdsize{0.08} & \underline{\bfseries 34.63 ± \stdsize{0.17}} \\
 & MeanProb-LM & \underline{\bfseries 22.30 ± \stdsize{0.13}} & \underline{\bfseries 11.36 ± \stdsize{0.03}} & \bfseries 28.71 ± \stdsize{0.05} & 34.88 ± \stdsize{0.10} \\
 & CTRL-CE & 26.17 ± \stdsize{0.39} & 15.98 ± \stdsize{0.08} & 34.69 ± \stdsize{0.27} & 38.29 ± \stdsize{0.16} \\
\midrule 
\multirow[c]{4}{*}{10\%} & Last-CE/JS & 30.41 ± \stdsize{0.09} & 15.88 ± \stdsize{0.11} & 32.49 ± \stdsize{0.18} & 44.73 ± \stdsize{0.48} \\
 & Mean-CE & 23.96 ± \stdsize{0.18} & 12.65 ± \stdsize{0.10} & 32.09 ± \stdsize{0.06} & \bfseries 34.98 ± \stdsize{0.11} \\
 & MeanProb-LM & \underline{\bfseries 22.30 ± \stdsize{0.13}} & \bfseries 11.48 ± \stdsize{0.03} & \underline{\bfseries 28.44 ± \stdsize{0.10}} & 35.20 ± \stdsize{0.21} \\
 & CTRL-CE & 26.17 ± \stdsize{0.39} & 15.93 ± \stdsize{0.18} & 34.69 ± \stdsize{0.27} & 38.21 ± \stdsize{0.21} \\
\bottomrule
\end{tabular}

%% file: Tables/optimized_tabular.tex
\begin{tabular}{llccccccc}

\toprule

 & \textbf{Model} & \multicolumn{7}{c}{\textbf{MLP}} \\
\cmidrule(lr){2-2} \cmidrule(lr){3-9} 
 & \textbf{Dataset} & \textbf{\shortstack{Cardioto-\\cography}} & \textbf{\shortstack{Credit\\Fraud}} & \textbf{\shortstack{Human\\Activity}} & \textbf{Letter} & \textbf{Mushroom} & \textbf{Satellite} & \textbf{\shortstack{Sensorless\\Drive}} \\
\cmidrule(lr){2-2} \cmidrule(lr){3-3} \cmidrule(lr){4-4} \cmidrule(lr){5-5} \cmidrule(lr){6-6} \cmidrule(lr){7-7} \cmidrule(lr){8-8} \cmidrule(lr){9-9} 
 & \textbf{Noise type} & \textbf{Symmetric} & \textbf{Symmetric} & \textbf{Symmetric} & \textbf{Symmetric} & \textbf{Symmetric} & \textbf{Symmetric} & \textbf{Symmetric} \\
\cmidrule(lr){2-2} \cmidrule(lr){3-3} \cmidrule(lr){4-4} \cmidrule(lr){5-5} \cmidrule(lr){6-6} \cmidrule(lr){7-7} \cmidrule(lr){8-8} \cmidrule(lr){9-9} 
 & \textbf{Noise rate} & \textbf{20.00\%} & \textbf{20.00\%} & \textbf{20.00\%} & \textbf{20.00\%} & \textbf{20.00\%} & \textbf{20.00\%} & \textbf{20.00\%} \\
\cmidrule(lr){2-2} \cmidrule(lr){3-3} \cmidrule(lr){4-4} \cmidrule(lr){5-5} \cmidrule(lr){6-6} \cmidrule(lr){7-7} \cmidrule(lr){8-8} \cmidrule(lr){9-9} 
Fraction & Method &  &  &  &  &  &  &  \\
\midrule
\multirow[c]{4}{*}{0\%} & Last-CE/JS & 47.47 ± \stdsize{6.00} & 17.87 ± \stdsize{4.60} & 32.73 ± \stdsize{0.78} & 9.24 ± \stdsize{0.14} & 2.45 ± \stdsize{0.17} & 10.56 ± \stdsize{0.30} & 0.77 ± \stdsize{0.06} \\
 & Mean-CE & 31.31 ± \stdsize{1.20} & \bfseries 10.82 ± \stdsize{0.42} & 5.42 ± \stdsize{0.14} & \bfseries 3.42 ± \stdsize{0.24} & \underline{\bfseries 0.20 ± \stdsize{0.08}} & 9.79 ± \stdsize{0.06} & 0.60 ± \stdsize{0.01} \\
 & MeanProb-LM & 30.91 ± \stdsize{0.49} & 11.00 ± \stdsize{0.64} & 8.35 ± \stdsize{0.57} & 3.64 ± \stdsize{0.11} & 0.23 ± \stdsize{0.05} & 10.22 ± \stdsize{0.38} & \bfseries 0.57 ± \stdsize{0.02} \\
 & CTRL-CE & \underline{\bfseries 23.23 ± \stdsize{1.00}} & 11.86 ± \stdsize{0.42} & \bfseries 3.73 ± \stdsize{0.04} & 5.41 ± \stdsize{0.12} & 0.60 ± \stdsize{0.08} & \underline{\bfseries 9.10 ± \stdsize{0.22}} & 0.89 ± \stdsize{0.04} \\
\midrule 
\multirow[c]{4}{*}{1\%} & Last-CE/JS & 28.28 ± \stdsize{3.50} & 13.23 ± \stdsize{1.10} & 9.71 ± \stdsize{0.46} & 6.42 ± \stdsize{0.31} & 9.70 ± \stdsize{0.42} & 9.96 ± \stdsize{0.28} & 0.87 ± \stdsize{0.06} \\
 & Mean-CE & 27.88 ± \stdsize{3.00} & 22.68 ± \stdsize{12.00} & \bfseries 3.93 ± \stdsize{0.27} & 5.10 ± \stdsize{0.19} & 12.28 ± \stdsize{0.74} & 10.82 ± \stdsize{0.27} & \bfseries 0.71 ± \stdsize{0.02} \\
 & MeanProb-LM & 29.09 ± \stdsize{3.00} & \bfseries 13.06 ± \stdsize{0.64} & 5.40 ± \stdsize{0.40} & 4.09 ± \stdsize{0.15} & 12.35 ± \stdsize{0.69} & 11.34 ± \stdsize{0.06} & 0.73 ± \stdsize{0.01} \\
 & CTRL-CE & \bfseries 26.16 ± \stdsize{1.80} & 16.84 ± \stdsize{0.64} & 5.65 ± \stdsize{0.53} & \bfseries 3.47 ± \stdsize{0.14} & \bfseries 1.03 ± \stdsize{0.12} & \underline{\bfseries 9.10 ± \stdsize{0.22}} & 1.07 ± \stdsize{0.02} \\
\midrule 
\multirow[c]{4}{*}{5\%} & Last-CE/JS & \bfseries 25.15 ± \stdsize{1.90} & 13.23 ± \stdsize{1.10} & 4.72 ± \stdsize{0.13} & 3.94 ± \stdsize{0.23} & 0.76 ± \stdsize{0.25} & 9.66 ± \stdsize{0.50} & 0.68 ± \stdsize{0.02} \\
 & Mean-CE & 27.98 ± \stdsize{3.40} & \bfseries 13.06 ± \stdsize{0.64} & 4.11 ± \stdsize{0.09} & 3.47 ± \stdsize{0.13} & \bfseries 0.26 ± \stdsize{0.17} & 10.44 ± \stdsize{0.06} & 0.54 ± \stdsize{0.01} \\
 & MeanProb-LM & 27.17 ± \stdsize{2.20} & \bfseries 13.06 ± \stdsize{0.64} & 3.96 ± \stdsize{0.16} & \bfseries 3.24 ± \stdsize{0.10} & 0.46 ± \stdsize{0.17} & 10.48 ± \stdsize{0.21} & \underline{\bfseries 0.49 ± \stdsize{0.01}} \\
 & CTRL-CE & 26.16 ± \stdsize{1.80} & 20.45 ± \stdsize{0.24} & \underline{\bfseries 3.61 ± \stdsize{0.20}} & 3.47 ± \stdsize{0.14} & 1.03 ± \stdsize{0.12} & \underline{\bfseries 9.10 ± \stdsize{0.22}} & 0.89 ± \stdsize{0.04} \\
\midrule 
\multirow[c]{4}{*}{10\%} & Last-CE/JS & 32.83 ± \stdsize{1.40} & 11.68 ± \stdsize{1.10} & 3.91 ± \stdsize{0.19} & 3.88 ± \stdsize{0.19} & \bfseries 0.76 ± \stdsize{0.25} & 10.95 ± \stdsize{0.65} & 0.71 ± \stdsize{0.04} \\
 & Mean-CE & 28.79 ± \stdsize{3.00} & \underline{\bfseries 10.65 ± \stdsize{0.64}} & 3.68 ± \stdsize{0.13} & 3.23 ± \stdsize{0.12} & 1.22 ± \stdsize{0.28} & 10.78 ± \stdsize{0.24} & 0.55 ± \stdsize{0.01} \\
 & MeanProb-LM & 31.52 ± \stdsize{0.43} & \underline{\bfseries 10.65 ± \stdsize{0.88}} & 4.16 ± \stdsize{0.00} & \underline{\bfseries 3.00 ± \stdsize{0.13}} & 0.89 ± \stdsize{0.29} & 11.25 ± \stdsize{0.28} & \bfseries 0.50 ± \stdsize{0.01} \\
 & CTRL-CE & \bfseries 25.56 ± \stdsize{4.50} & 12.20 ± \stdsize{1.20} & \underline{\bfseries 3.61 ± \stdsize{0.20}} & 3.47 ± \stdsize{0.14} & 1.03 ± \stdsize{0.12} & \bfseries 9.87 ± \stdsize{0.34} & 0.89 ± \stdsize{0.04} \\
\bottomrule
\end{tabular}

%% file: Plots/optimization_curves.tex
\begin{subfigure}[b]{0.48\textwidth}
    \begin{tikzpicture}
        \begin{axis}[
            ylabel={$\text{FNR}(\eta)$},
            ymajorgrids=true,
            xticklabel style={/pgf/number format/fixed},
            width=\textwidth,
            height=0.6\textwidth,
            xmin=0,
            xmax=500,
            ymin=0.3,
            ymax=0.55,
            legend style={/tikz/opacity=0},
        ]
        \pgfplotstableread{Plots/fnr_curve_Food-101N.csv}\plotdata
        \plotwithconfidenceinterval[blue]{\plotdata}{epoch}{mean_Last_CE}{std_Last_CE}
        \addlegendentry{Last-CE}
        \plotwithconfidenceinterval[red]{\plotdata}{epoch}{mean_Mean_CE}{std_Mean_CE}
        \addlegendentry{Mean-CE}
        \plotwithconfidenceinterval[orange]{\plotdata}{epoch}{mean_MeanProb_CE}{std_MeanProb_CE}
        \addlegendentry{MeanProb-CE}
        \plotwithconfidenceinterval[violet]{\plotdata}{epoch}{mean_MeanProb_LM}{std_MeanProb_LM}
        \addlegendentry{MeanProb-LM}
        \addplot [dashed, thick, gray, ycomb] coordinates {(248, 0.55)};
        \end{axis}
    \end{tikzpicture}
    \caption{Food-101N}
\end{subfigure}
\hfill
\begin{subfigure}[b]{0.48\textwidth}
    \begin{tikzpicture}
        \begin{axis}[
            ymajorgrids=true,
            xticklabel style={/pgf/number format/fixed},
            width=\textwidth,
            height=0.6\textwidth,
            xmin=0,
            xmax=400,
            ymin=0.25,
            ymax=0.45,
        ]
        \pgfplotstableread{Plots/fnr_curve_Clothing1M.csv}\plotdata
        \plotwithconfidenceinterval[blue]{\plotdata}{epoch}{mean_Last_CE}{std_Last_CE}
        \addlegendentry{Last-CE}
        \plotwithconfidenceinterval[red]{\plotdata}{epoch}{mean_Mean_CE}{std_Mean_CE}
        \addlegendentry{Mean-CE}
        \plotwithconfidenceinterval[orange]{\plotdata}{epoch}{mean_MeanProb_CE}{std_MeanProb_CE}
        \addlegendentry{MeanProb-CE}
        \plotwithconfidenceinterval[violet]{\plotdata}{epoch}{mean_MeanProb_LM}{std_MeanProb_LM}
        \addlegendentry{MeanProb-LM}
        \addplot [dashed, thick, gray, ycomb] coordinates {(189, 0.55)};
        \end{axis}
    \end{tikzpicture}
    \caption{Clothing1M}
\end{subfigure}

\vspace{0.5cm}

\begin{subfigure}[b]{0.48\textwidth}
    \begin{tikzpicture}
        \begin{axis}[
            xlabel={Stop epoch},
            ylabel={$\text{FNR}(\eta)$},
            ymajorgrids=true,
            xticklabel style={/pgf/number format/fixed},
            width=\textwidth,
            height=0.6\textwidth,
            xmin=0,
            xmax=70,
            ymin=0.1,
            ymax=0.25,
            legend style={/tikz/opacity=0},
        ]
        \pgfplotstableread{Plots/fnr_curve_CIFAR-10N.csv}\plotdata
        \plotwithconfidenceinterval[blue]{\plotdata}{epoch}{mean_Last_CE}{std_Last_CE}
        \addlegendentry{Last-CE}
        \plotwithconfidenceinterval[red]{\plotdata}{epoch}{mean_Mean_CE}{std_Mean_CE}
        \addlegendentry{Mean-CE}
        \plotwithconfidenceinterval[orange]{\plotdata}{epoch}{mean_MeanProb_CE}{std_MeanProb_CE}
        \addlegendentry{MeanProb-CE}
        \plotwithconfidenceinterval[violet]{\plotdata}{epoch}{mean_MeanProb_LM}{std_MeanProb_LM}
        \addlegendentry{MeanProb-LM}
        \addplot [dashed, thick, gray, ycomb] coordinates {(35, 0.55)};
        \end{axis}
    \end{tikzpicture}
    \caption{CIFAR-10N - Worse}
\end{subfigure}
\hfill
\begin{subfigure}[b]{0.48\textwidth}
    \begin{tikzpicture}
        \begin{axis}[
            xlabel={Stop epoch},
            ymajorgrids=true,
            xticklabel style={/pgf/number format/fixed},
            width=\textwidth,
            height=0.6\textwidth,
            xmin=0,
            xmax=1500,
            ymin=0.02,
            ymax=0.1,
            legend style={/tikz/opacity=0},
            yticklabel style={
                /pgf/number format/fixed,
                /pgf/number format/precision=5
            },
            scaled y ticks=false,
        ]
        \pgfplotstableread{Plots/fnr_curve_Human_Activity.csv}\plotdata
        \plotwithconfidenceinterval[blue]{\plotdata}{epoch}{mean_Last_CE}{std_Last_CE}
        \addlegendentry{Last-CE}
        \plotwithconfidenceinterval[red]{\plotdata}{epoch}{mean_Mean_CE}{std_Mean_CE}
        \addlegendentry{Mean-CE}
        \plotwithconfidenceinterval[orange]{\plotdata}{epoch}{mean_MeanProb_CE}{std_MeanProb_CE}
        \addlegendentry{MeanProb-CE}
        \plotwithconfidenceinterval[violet]{\plotdata}{epoch}{mean_MeanProb_LM}{std_MeanProb_LM}
        \addlegendentry{MeanProb-LM}
        \addplot [dashed, thick, gray, ycomb] coordinates {(797, 0.55)};
        \end{axis}
    \end{tikzpicture}
    \caption{Human Activity}
\end{subfigure}

%% file: 06_conclusion.tex

\section{Conclusion}

In this work, we presented a comprehensive benchmark for noisy label detection, introducing a unified evaluation framework and a modular taxonomy that decomposes detection methods into three core components: gathering strategy, label disagreement measure, and aggregation method. Our key findings show that detection performance varies with noise structure, data modality, and model architecture, so that no single method dominates in every scenario. For instance, Mean-CPD achieves top results in specific synthetic settings, while MeanProb-CE/JS performs particularly well on datasets like Food-101N. Nevertheless, the proposed combination of Mean Probabilities aggregation with the Logit Margin disagreement measure (MeanProb-LM) consistently ranks near the top across all evaluated synthetic and real-world settings, emerging as the most reliable default choice when prior knowledge about the underlying noise distribution is unavailable.

Our evaluation also provides practical guidance regarding gathering strategies and temporal dynamics modeling. First, we find that out-of-sample ($K$-fold cross-validation) gathering does not consistently improve detection performance over in-sample gathering, demonstrating that the computational overhead of cross-validation is generally unnecessary and simpler in-sample strategies are preferable. Second, methods exploiting training dynamics via clustering (such as CTRL) can achieve top performance in specific settings, such as pairflip noise or scenarios with severe label memorization, where clustering proves particularly resilient. However, the method remains inconsistent across noise conditions due to its discrete voting mechanism. Moreover, this performance gap persists even after accounting for discretization effects. In contrast, continuous temporal ensembling (MeanProb-LM) avoids these discretization artifacts, providing a simpler and more reliably robust detection signal.

\new{While ensemble methods show clear advantages, the performance of Stochastic Weight Averaging (SWA) depends heavily on the model architecture and training duration. On vision models, SWA performs worse than probability-level ensembling, primarily because aggregating weights from early training epochs, when models are underfit, severely pollutes parameter averaging. In contrast, on tabular datasets with smaller architectures trained over many epochs, SWA performs noticeably better, matching or exceeding other methods in several cases. This suggests that while SWA can be effective when the model spends a large fraction of training in a stable loss region, probability-level ensembling remains a more consistently robust strategy without requiring careful tuning of the weight averaging window.}

We also explored the sensitivity of detection methods to hyperparameter tuning. In particular, we found that MeanProb-LM demonstrates low sensitivity to hyperparameter changes. This is a valuable property, as hyperparameter optimization often requires relabeling a significant fraction of samples to evaluate detection performance. Given the minimal performance gains observed, such optimization efforts may not be justified, especially when resources could be more efficiently allocated by using the unoptimized method directly.

Although the post-hoc optimization performed appears to offer limited benefit, optimizing the training process itself for detection may unlock further improvements in noise detection. For instance, prior work demonstrates that using a cyclic learning rate schedule \cite{huang_o2u-net_2019}, possibly with checkpointing \cite{kamassury}, can enhance detection performance. However, optimizing training for detection presents significant challenges, such as the increased computational cost due to multiple training rounds. Moreover, data-efficient strategies need to be developed to allow optimization on small tuning sets without risking overfitting.

While we have explored vision and tabular datasets, a natural extension for this work is to consider other dataset domains and tasks such as text, audio, time series, and other classification tasks. We use a range of noise types from symmetric noise to real-world noise; however, other noise types may be challenging and reveal weaknesses and strengths of each method, such as asymmetric noise and instance noise. Exploring how different model architectures, such as transformer-based models, affect detection would be valuable as well. Incorporating different label noise learning techniques during the training phases for detection may also yield significant improvements.  


\section*{Acknowledgements}

This work was supported in part by the Brazilian National Council for Scientific and Technological Development (CNPq-Brazil) under Grants 304619/2022-1 and 402516/2024-9, and in part by Fundacao de Amparo à Pesquisa e Inovacao do Estado de Santa Catarina (FAPESC) under Grant 2024TR000090.

%% file: 09_appendix.tex



\section{Contingency Tables}
\label{app:contingency}
\input{Appendices/contingency}

\section{Hyperparameters}
\label{app:hyperparams}
\input{Appendices/hyperparams}

\section{Classification Performances}
\label{app:classification_performance}
\begin{table}[H]
    \centering
    \begin{adjustbox}{max width=\textwidth, keepaspectratio}
    \input{Tables/classification_performances}
    \end{adjustbox}
    \caption{Classification performance achieved on each setting as measured on the clean test set.}
\end{table}

\section{Detection Hyperparameters}
\label{app:detection_hypp}
\input{Appendices/detection_hyperparams}

\section{CTRL implementation}
\label{app:ctrl_impl}

\begin{algorithm}[H]
\caption{CTRL Algorithm}
\label{alg:ctrl_votes}
\begin{algorithmic}[1]
\Require Loss matrix $\mathbf{L} \in \mathbb{R}^{N \times T}$, Noisy labels $\mathbf{y} \in \{1, \dots, C\}^N$
\Require Number of windows $W$, Smooth window size $S$, Number of clusters $K$, Selected clusters $K_s$
\Ensure Noisiness scores $\mathbf{v} \in \mathbb{R}^N$

\State $C_{\text{unique}}, \mathbf{c}_{\text{counts}} \gets \Call{Unique}{\mathbf{y}}$ \Comment{Get unique classes and their counts}
\State $L_{\text{thresh}} \gets 2 \log(|C_{\text{unique}}|)$
\State $\mathbf{L} \gets \Call{Clip}{\mathbf{L}, \text{upper}=L_{\text{thresh}}}$
\State $\mathbf{L} \gets \Call{RollingMean}{\mathbf{L}, \text{window\_size}=S}$ \Comment{Smooth loss trajectories}

\State Divide $T$ epochs evenly into $W$ windows: $\mathcal{W} = \{w_1, w_2, \dots, w_W\}$
\State Initialize mask votes $\mathbf{M} \in \mathbf{1}^{N \times W}$

\For{each window $w \in \mathcal{W}$ with boundaries $[t_{\text{start}}, t_{\text{end}}]$}
    \For{each class $c \in C_{\text{unique}}$}
        \If{number of samples in class $c < 2$}
            \State \textbf{continue}
        \EndIf
        
        \State $\mathbf{L}_{c, w} \gets \mathbf{L}[i, t_{\text{start}}:t_{\text{end}}]$ for all $i$ where $\mathbf{y}_i = c$ \Comment{Gather losses in window $w$ and class $c$}
        
        \State $\mathcal{C}, \mathbf{A} \gets \Call{KMeans}{\mathbf{L}_{c, w}, \text{n\_clusters}=K}$ \Comment{$\mathcal{C}$: cluster centers, $\mathbf{A}$: assignments}
        
        \State Compute area under loss curve for each center: $\mathbf{a} \gets \sum_{\tau} \mathcal{C}[:, \tau]$
        \State Sort clusters by $\mathbf{a}$ in descending order to get sorted indices $\pi$
        
        \For{$k = 1$ \textbf{to} $K_s$} \Comment{Assign samples in top $K_s$ loss clusters as noisy for this window}
            \State $\mathbf{M}[i, w] \gets 0$ for all $i$ where $\mathbf{y}_i = c$ and $\mathbf{A}_i = \pi[k]$
        \EndFor
    \EndFor
\EndFor

\State Aggregate votes across all windows: $\mathbf{u} \gets \sum_{w=1}^W \mathbf{M}[:, w]$
\State \Return $\mathbf{v} \gets \max(\mathbf{u}) - \mathbf{u}$ \Comment{Invert so highest vote is the noisiest sample}
\end{algorithmic}
\end{algorithm}

\new{
\subsection{CTRL Hyperparameter Optimization and Scoring Metric}
\label{app:ctrl_opt}

Following Yue and Jha \cite{yue_ctrl_2023}, the optimal hyperparameter candidate tuple $(K, K_s, W)$ is selected by evaluating candidate clean masks $m \in \{0, 1\}^N$ against the scoring function:
\begin{equation}
\text{score}(m) = S \times \left[ A \times R \right]^\alpha,
\label{eq:ctrl_opt_score}
\end{equation}
where the components are defined as follows:
\begin{enumerate}
    \item \textbf{Silhouette Score ($S$)}: Measures the trajectory separation between clean ($m_i = 0$) and noisy ($m_i = 1$) classified samples in the smoothed loss space $\mathbf{L}_{\text{smoothed}}$:
    \begin{equation}
        S = \text{Silhouette}\left(\mathbf{L}_{\text{smoothed}}, m\right).
    \end{equation}
    
    \item \textbf{Masked Training Accuracy ($A$)}: Evaluates the classification accuracy of the model's final-epoch predictions $\hat{y}_i$ against the given noisy labels $\tilde{y}_i$ restricted to samples flagged as clean ($m_i = 0$):
    \begin{equation}
        A = \frac{\sum_{i: m_i = 0} \mathbb{I}(\hat{y}_i = \tilde{y}_i)}{\sum_{i=1}^N m_i}.
    \end{equation}

    \item \textbf{Loss Ratio ($R$)}: Measures the ratio of the mean final-epoch smoothed loss between noisy-designated and clean-designated samples:
    \begin{equation}
        R = \frac{\frac{1}{\sum_{i} m_i} \sum_{i: m_i = 1} \mathbf{L}_{\text{smoothed}}[i, T]}{\left(\frac{1}{\sum_{i} (1-m_i)} \sum_{i: m_i = 0} \mathbf{L}_{\text{smoothed}}[i, T]\right)}.
    \end{equation}
\end{enumerate}

The trade-off parameter $\alpha$ balances the clustering silhouette against training accuracy and loss ratio. Following the recommended paper settings, we specify $\alpha$ per dataset:
\begin{equation}
\alpha = \begin{cases}
1.0, & \text{for CIFAR-10 and CIFAR-10N}, \\
0.25, & \text{for CIFAR-100}, \\
0.5, & \text{for Clothing1M and Food-101N}, \\
0.0, & \text{for Tabular Datasets and Logit Margin (LM) loss measure}.
\end{cases}
\end{equation}
Where the values for CIFAR and tabular datasets are provided by the authors. Since it is not possible to optimize the parameter $\alpha$ itself, we choose the intermediate value $0.5$ for the Food-101N and Clothing1M datasets. Furthermore, since LM is a signed loss it uncharacterizes the loss ratio measure and so we set $\alpha=0$.

Notably, while the original optimization includes the vote threshold, we force selecting a fraction of samples equal to the noise rate, meaning we already have the optimal threshold for our evaluation metric. With this, $m$ is generated by assigning $m_i = 1$ to the $B = \lceil\budget \numberofsamples\rceil$ highest-ranked (noisiest) samples returned by Algorithm~\ref{alg:ctrl_votes}. To isolate hyperparameter tuning from single-run stochasticity, candidate scores are averaged across random seeds:
\begin{equation}
\bar{\text{score}}(K, K_s, W) = \frac{1}{|\mathcal{S}|} \sum_{s \in \mathcal{S}} \text{score}_s(K, K_s, W),
\end{equation}
and the candidate tuple $(K, K_s, W)$ maximizing $\bar{\text{score}}$ is selected as the single optimal hyperparameter configuration for the benchmark.
}

%% file: Appendices/contingency.tex
\begin{figure}
    \centering

    \begin{subfigure}[t]{0.48\textwidth}
        \centering
        \includegraphics[width=\textwidth]{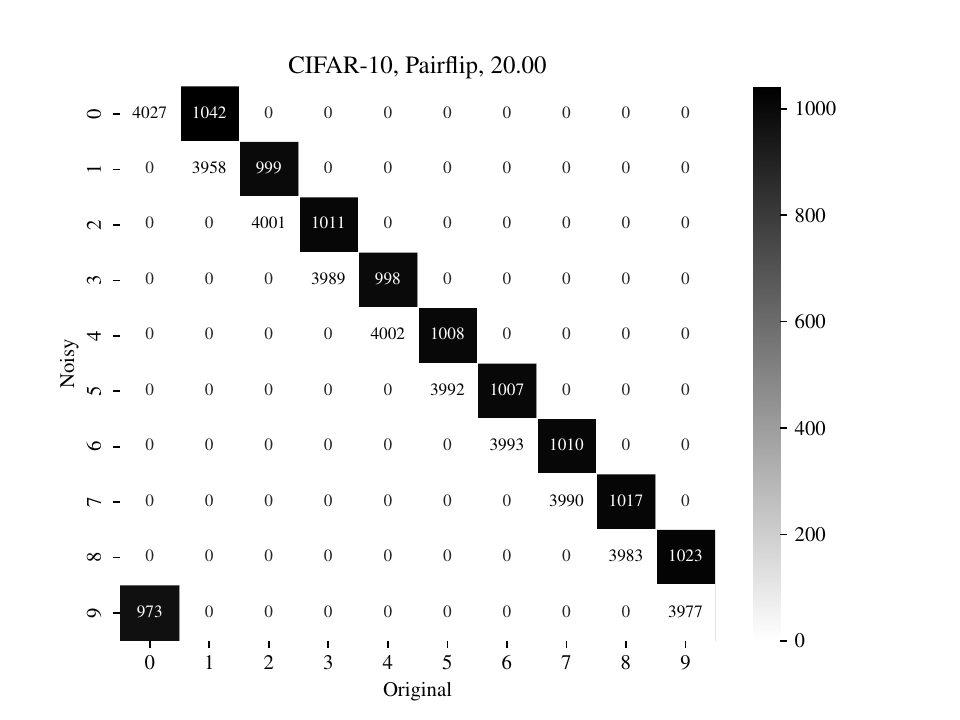}
    \end{subfigure}
    \begin{subfigure}[t]{0.48\textwidth}
        \centering
        \includegraphics[width=\textwidth]{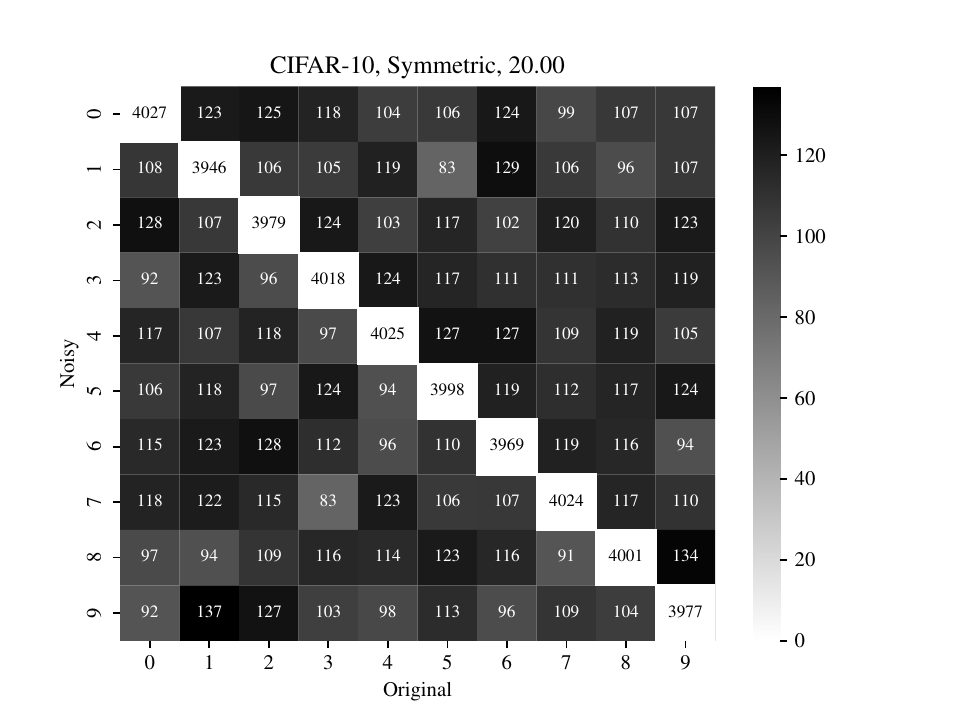}
    \end{subfigure}

    \begin{subfigure}[t]{0.48\textwidth}
        \centering
        \includegraphics[width=\textwidth]{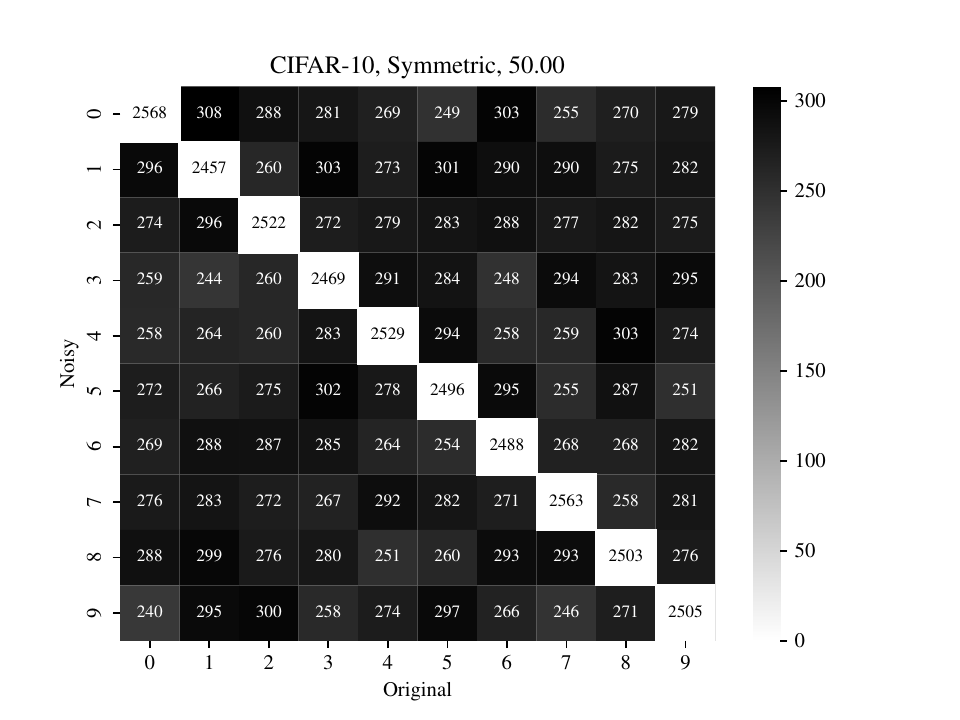}
    \end{subfigure}
    \begin{subfigure}[t]{0.48\textwidth}
        \centering
        \includegraphics[width=\textwidth]{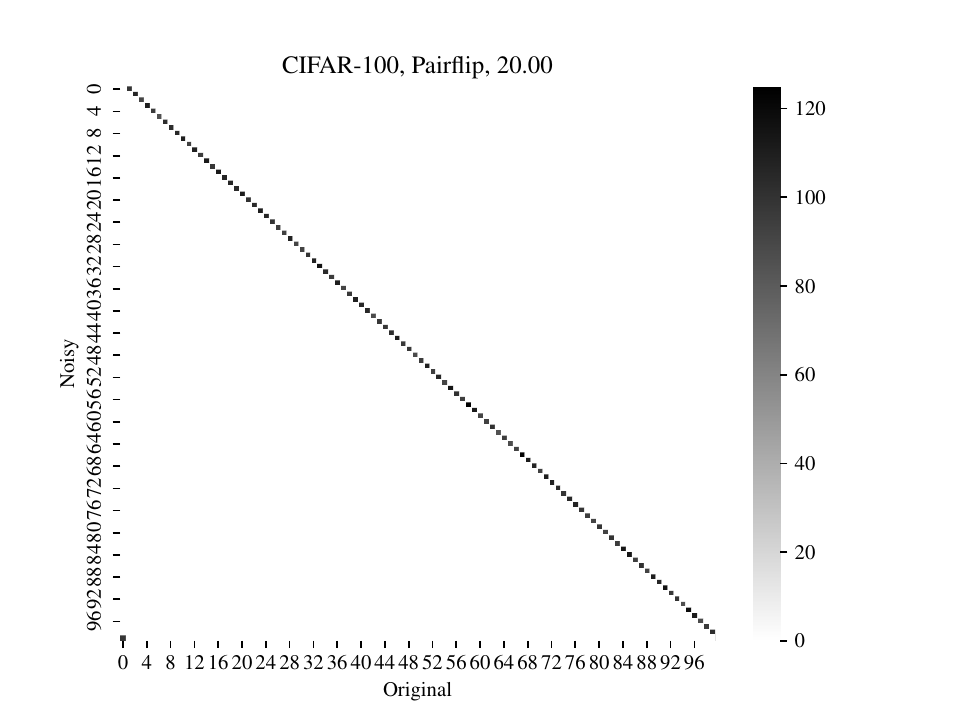}
    \end{subfigure}

    \begin{subfigure}[t]{0.48\textwidth}
        \centering
        \includegraphics[width=\textwidth]{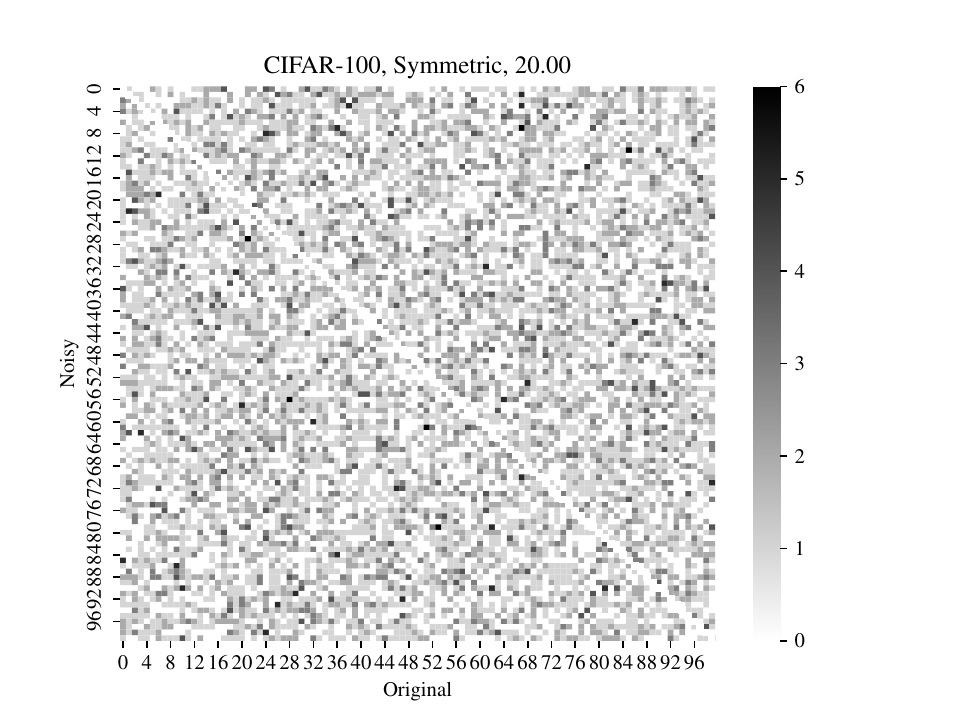}
    \end{subfigure}
    \begin{subfigure}[t]{0.48\textwidth}
        \centering
        \includegraphics[width=\textwidth]{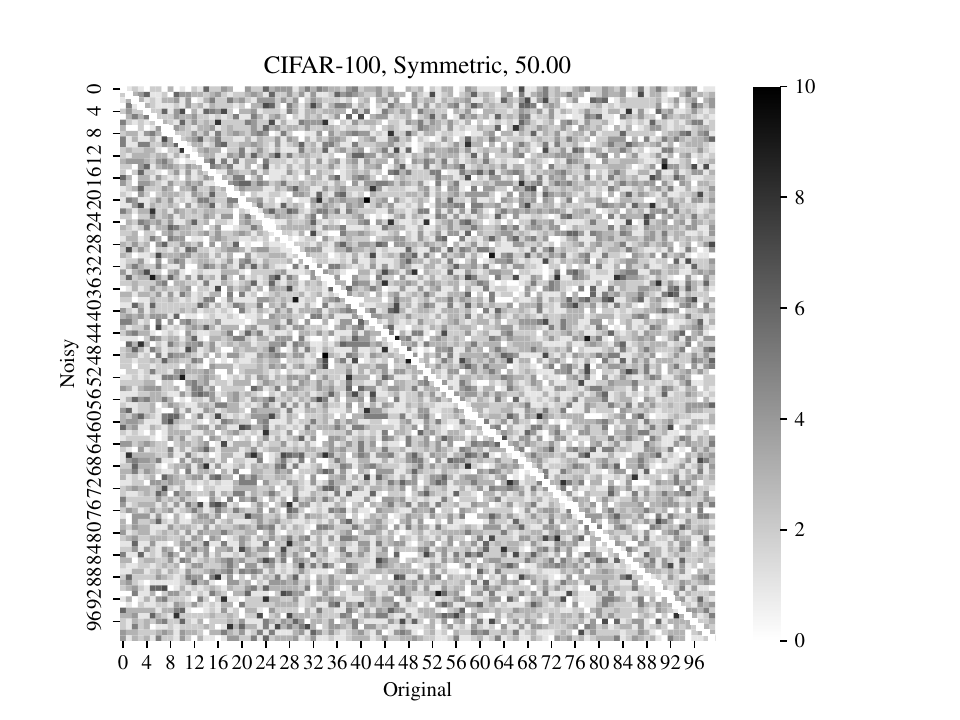}
    \end{subfigure}

    \caption{Contingency tables for synthetic noise vision datasets.}
    \label{fig:noise_contingency_matrices_vision}
\end{figure}
\begin{figure}[htbp]
    \centering

    \begin{subfigure}[t]{0.48\textwidth}
        \centering
        \includegraphics[width=\textwidth]{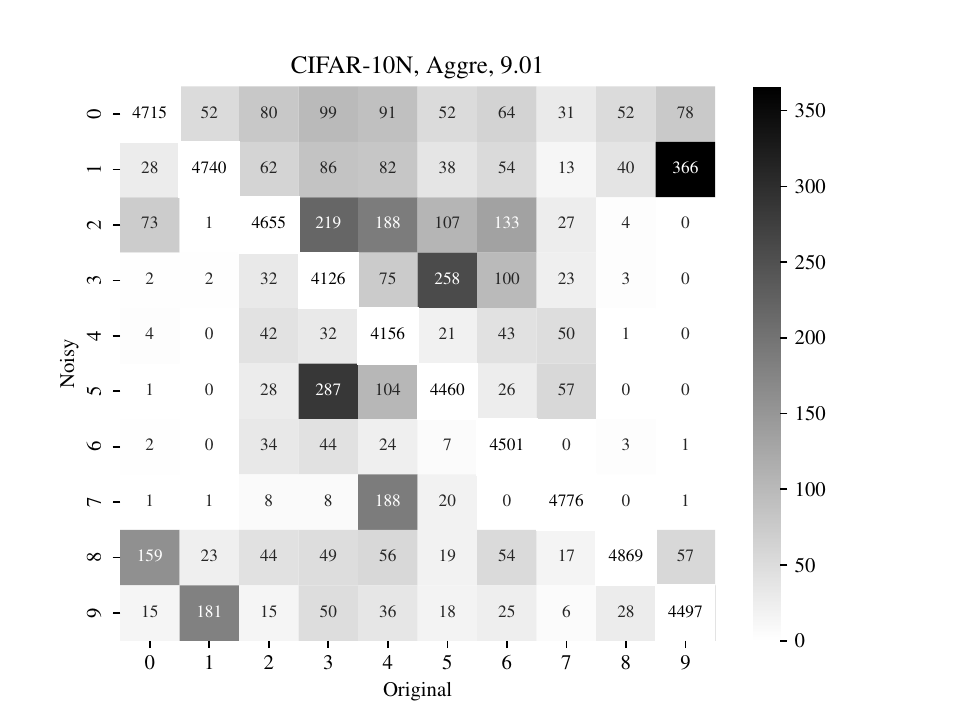}
    \end{subfigure}
    \begin{subfigure}[t]{0.48\textwidth}
        \centering
        \includegraphics[width=\textwidth]{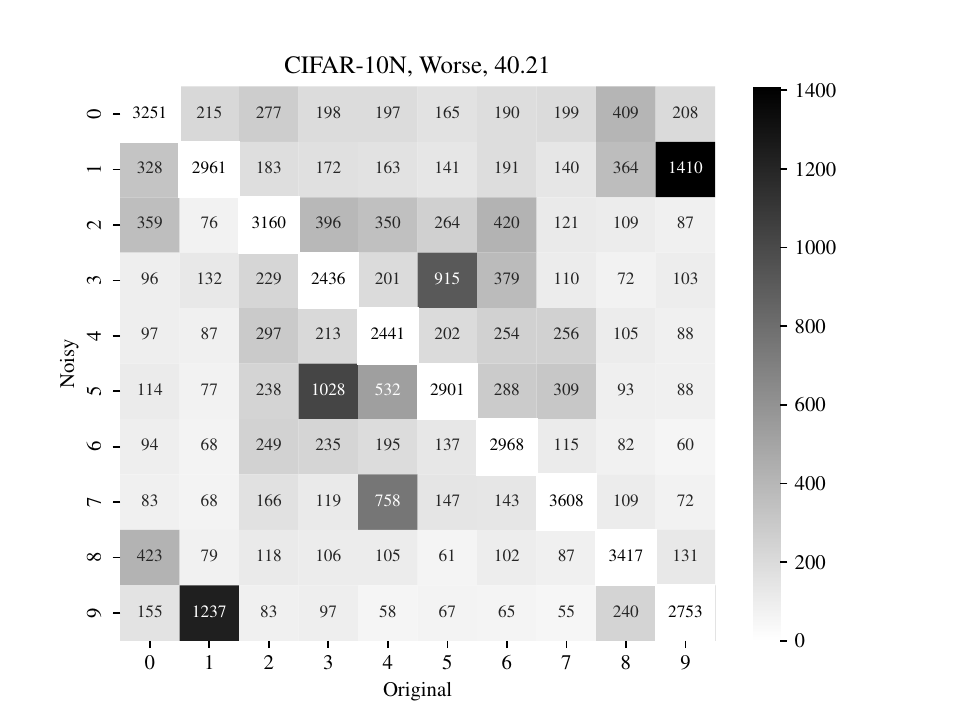}
    \end{subfigure}

    \begin{subfigure}[t]{0.48\textwidth}
        \centering
        \includegraphics[width=\textwidth]{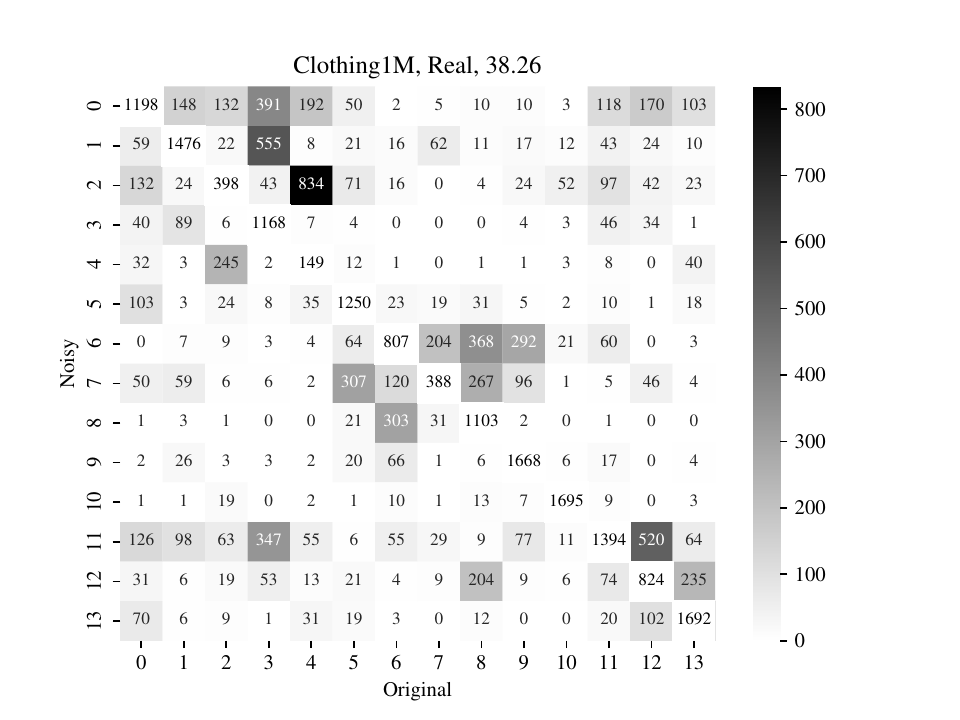}
    \end{subfigure}

    \caption{Contingency tables for real noise vision datasets.}
    \label{fig:noise_contingency_matrices_real}
\end{figure}
\begin{figure}[htbp]
    \centering

    \begin{subfigure}[t]{0.48\textwidth}
        \centering
        \includegraphics[width=\textwidth]{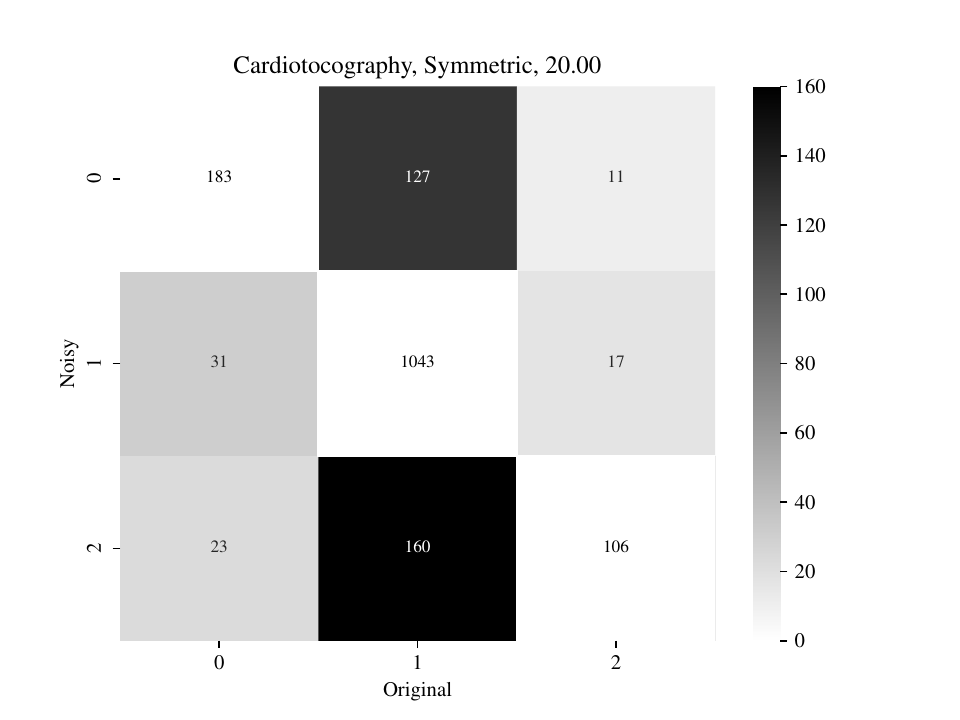}
    \end{subfigure}
    \begin{subfigure}[t]{0.48\textwidth}
        \centering
        \includegraphics[width=\textwidth]{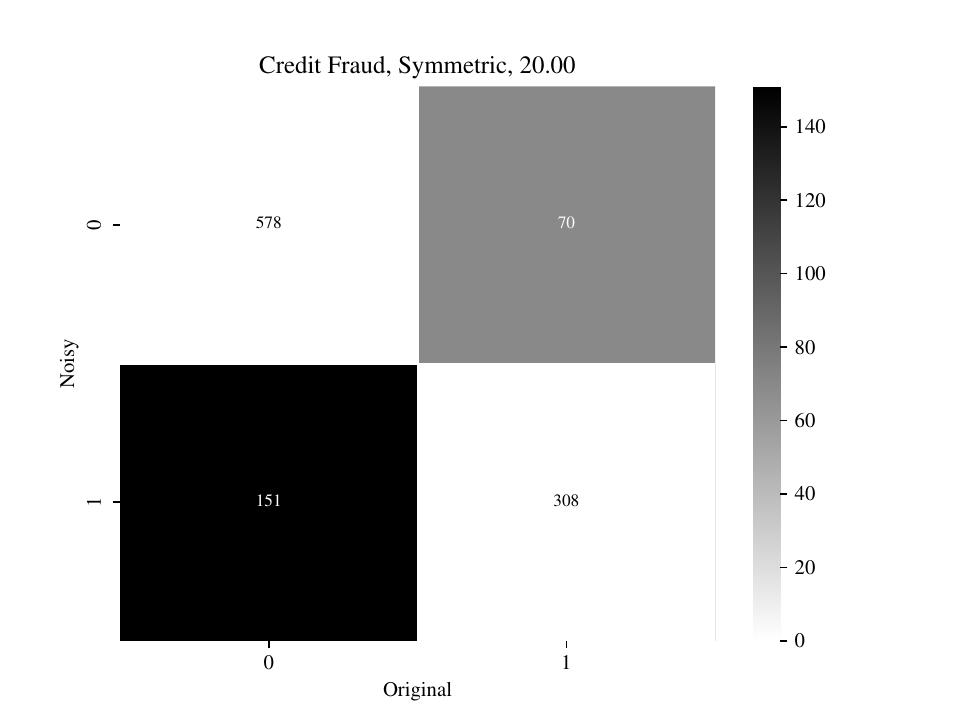}
    \end{subfigure}

    \begin{subfigure}[t]{0.48\textwidth}
        \centering
        \includegraphics[width=\textwidth]{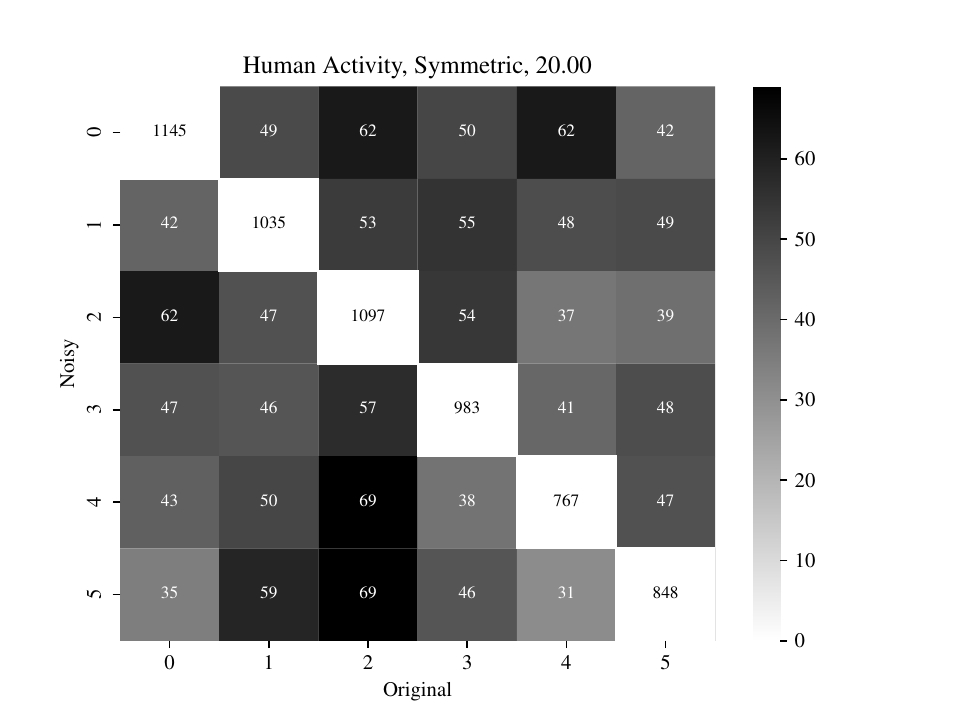}
    \end{subfigure}
    \begin{subfigure}[t]{0.48\textwidth}
        \centering
        \includegraphics[width=\textwidth]{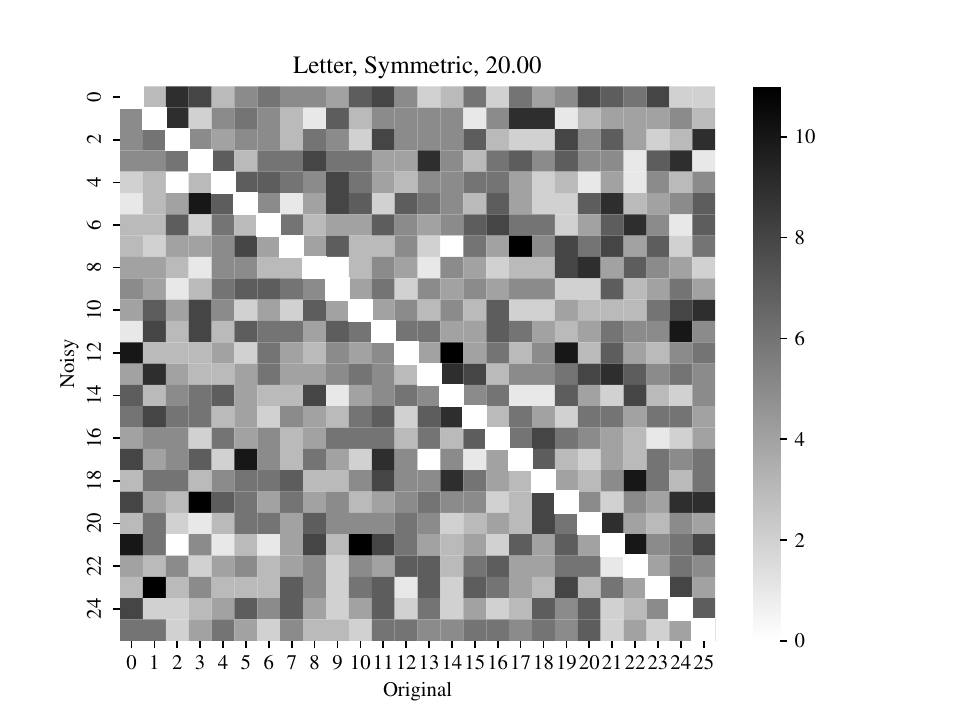}
    \end{subfigure}

    \begin{subfigure}[t]{0.48\textwidth}
        \centering
        \includegraphics[width=\textwidth]{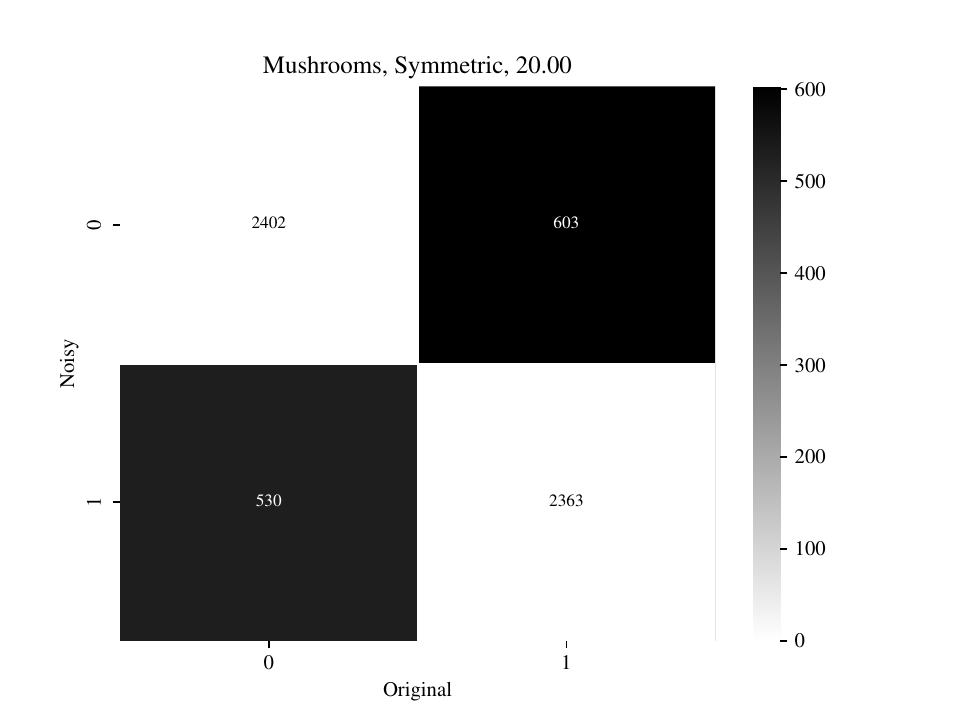}
    \end{subfigure}
    \begin{subfigure}[t]{0.48\textwidth}
        \centering
        \includegraphics[width=\textwidth]{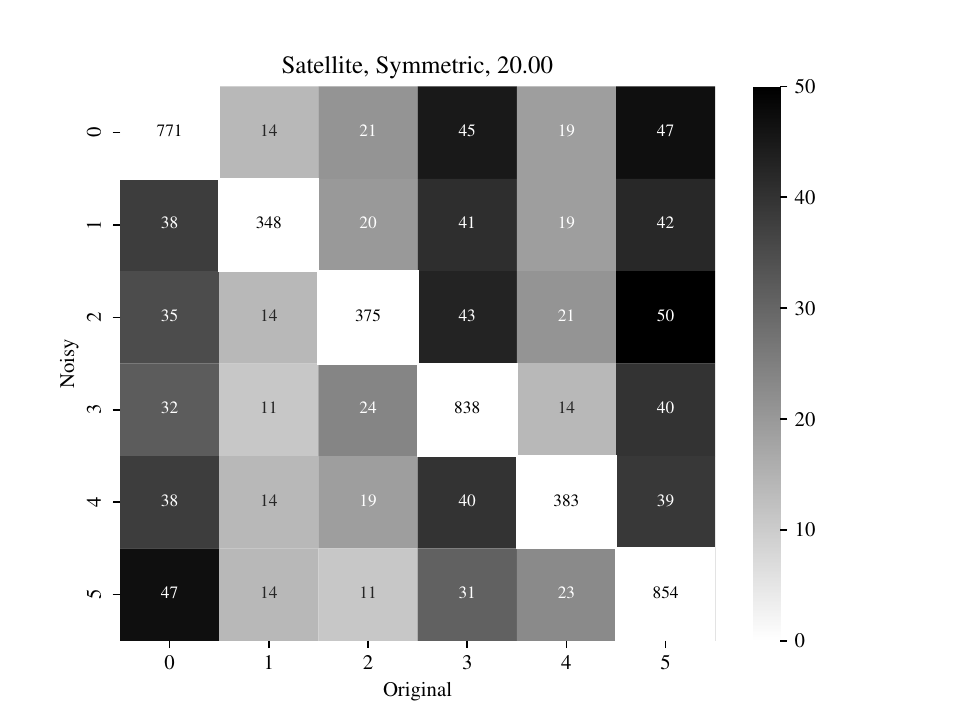}
    \end{subfigure}

    \begin{subfigure}[t]{0.48\textwidth}
        \centering
        \includegraphics[width=\textwidth]{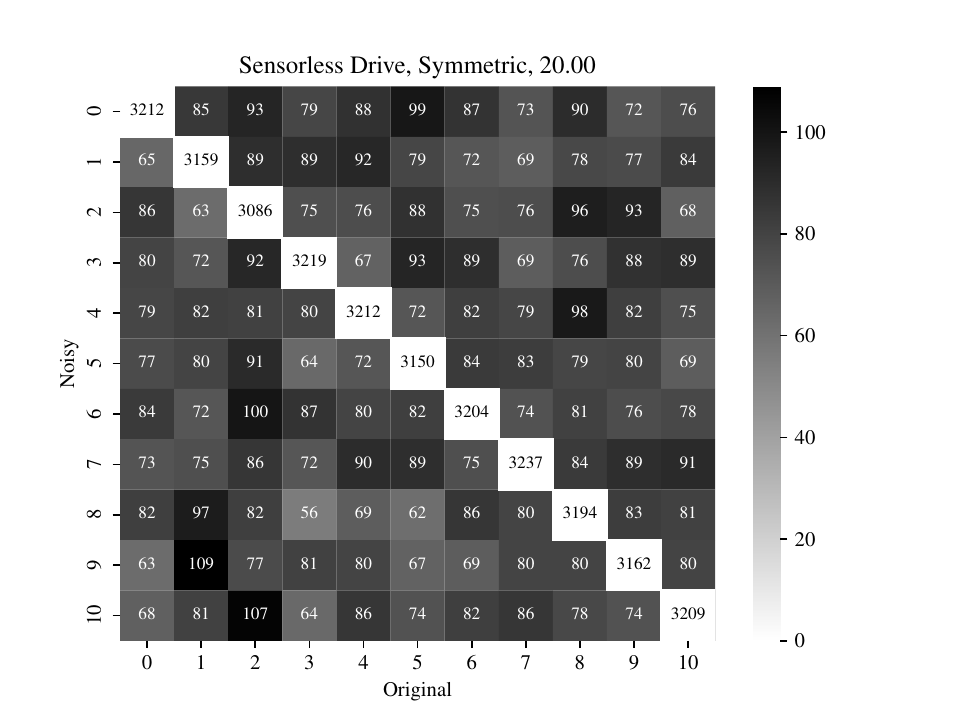}
    \end{subfigure}

    \caption{Contingency tables for synthetic noise tabular datasets.}
    \label{fig:noise_contingency_matrices_tabular}
\end{figure}

Figures \ref{fig:noise_contingency_matrices_vision}-\ref{fig:noise_contingency_matrices_tabular} provide the contingency matrices for each setting. We present them in a heatmap where, for easier visualization, the counts of the correct labels are not included in the color bar.

Since Food-101N only provides a value indicating if a sample is clean or noisy, it is not possible to compute a contingency matrix. We provide instead, the noise rate for each class on Figure \ref{fig:food_101_noise}.

For the Clothing1M dataset we use only the subset of data in the training set that has both clean and noisy labels. We note that on Clothing1M the 4th class is plagued with much more noise than correct labels, with 149 correct labels and 1185 wrong labels. We have observed that in this class most detection methods fail, presenting random detection.

\begin{figure}[htbp]
\centering
\begin{tikzpicture}
\begin{axis}[
    xbar,
    bar width=4pt,
    width=\textwidth,
    height=0.9\textheight,
    enlarge y limits=0.01,
    xlabel={Noise rate (\%)},
    symbolic y coords={
        Apple Pie,Baby Back Ribs,Baklava,Beef Carpaccio,Beef Tartare,Beet Salad,Beignets,Bibimbap,
        Bread Pudding,Breakfast Burrito,Bruschetta,Caesar Salad,Cannoli,Caprese Salad,Carrot Cake,
        Ceviche,Cheese Plate,Cheesecake,Chicken Curry,Chicken Quesadilla,Chicken Wings,Chocolate Cake,
        Chocolate Mousse,Churros,Clam Chowder,Club Sandwich,Crab Cakes,Creme Brulee,Croque Madame,
        Cup Cakes,Deviled Eggs,Donuts,Dumplings,Edamame,Eggs Benedict,Escargots,Falafel,Filet Mignon,
        Fish And Chips,Foie Gras,French Fries,French Onion Soup,French Toast,Fried Calamari,Fried Rice,
        Frozen Yogurt,Garlic Bread,Gnocchi,Greek Salad,Grilled Cheese Sandwich,Grilled Salmon,
        Guacamole,Gyoza,Hamburger,Hot And Sour Soup,Hot Dog,Huevos Rancheros,Hummus,Ice Cream,Lasagna,
        Lobster Bisque,Lobster Roll Sandwich,Macaroni And Cheese,Macarons,Miso Soup,Mussels,Nachos,
        Omelette,Onion Rings,Oysters,Pad Thai,Paella,Pancakes,Panna Cotta,Peking Duck,Pho,Pizza,
        Pork Chop,Poutine,Prime Rib,Pulled Pork Sandwich,Ramen,Ravioli,Red Velvet Cake,Risotto,
        Samosa,Sashimi,Scallops,Seaweed Salad,Shrimp And Grits,Spaghetti Bolognese,
        Spaghetti Carbonara,Spring Rolls,Steak,Strawberry Shortcake,Sushi,Tacos,Takoyaki,Tiramisu,
        Tuna Tartare,Waffles
    },
    ytick=data,
    ytick style={draw=none},
    xmin=0,
    xmax=50,
    axis y line*=left,
    axis x line*=bottom,
    tick label style={font=\fontsize{7.0}{7.5}\selectfont},
    every node near coord/.append style={font=\tiny, anchor=west, text=black},
    nodes near coords,
]
\addplot+[
    xbar,
    fill=black!30,
    draw=black!90
] table [
    y=Class,
    x=Noise rate,
    col sep=comma,
] {Appendices/noise_rates.csv};
\end{axis}
\end{tikzpicture}
\caption{Noise rate per noisy class, $P(\samplelabel \neq \samplelabelnoisy | \samplelabelnoisy)$, for the Food-101N dataset.}
\label{fig:food_101_noise}
\end{figure}

%% file: Appendices/hyperparams.tex
Table~\ref{tab:hyperparameters_classification} provides the hyperparameters used for training on each setting.

\begin{table}[H]
    \centering
    \begin{adjustbox}{angle=90, max width=\textwidth, max height=0.9\textheight, keepaspectratio}
    \input{Tables/hyperparameters}
    \end{adjustbox}
    \caption{Hyperparameters used for training. The Scheduler column lists the parameters of a step scheduler that adjusts the learning rate relative to the initial learning rate. It shows two lists of numbers separated by a forward slash where the first refers to the epochs and the second refers to the values that the initial learning rate should be multiplied by. The epoch values for Clothing1M and Food-101N are mini-epochs instead, where a mini-epoch is 1/20th and 1/10th of an epoch, respectively.}
    \label{tab:hyperparameters_classification}
\end{table}

%% file: Tables/hyperparameters.tex
\begin{tabular}{llllllll}
\toprule
 &  &  &  & Batch Size & Optimizer & Epochs & Scheduler \\
Dataset & Noise Type & Noise Rate & Model &  &  &  &  \\
\midrule
\multirow[t]{5}{*}{CIFAR-10} & Aggregate & 9.01\% & 9-Layer CNN & 64 & \lstinline|Adam(lr=3e-05, weight_decay=0.03)| & 100 & 0,50,80 / 1.0,0.1,0.01 \\
\cline{2-8} \cline{3-8}
 & Worse & 40.21\% & 9-Layer CNN & 64 & \lstinline|Adam(lr=0.0003, weight_decay=0.0002)| & 35 & 0,25 / 1.0,0.1 \\
\cline{2-8} \cline{3-8}
 & Pairflip & 20.00\% & 9-Layer CNN & 128 & \lstinline|Adam(lr=0.001, weight_decay=0.0001)| & 63 & 0,50 / 1.0,0.1 \\
\cline{2-8} \cline{3-8}
 & \multirow[t]{2}{*}{Symmetric} & 20.00\% & 9-Layer CNN & 64 & \lstinline|Adam(lr=5e-05, weight_decay=0.02)| & 63 & 0,50 / 1.0,0.1 \\
\cline{3-8}
 &  & 50.00\% & 9-Layer CNN & 32 & \lstinline|Adam(lr=0.0001, weight_decay=0.001)| & 60 & 0,50 / 1.0,0.1 \\
\cline{1-8} \cline{2-8} \cline{3-8}
\multirow[t]{3}{*}{CIFAR-100} & Pairflip & 20.00\% & 9-Layer CNN & 64 & \lstinline|Adam(lr=5e-05, weight_decay=0.01)| & 101 & 0,50,75 / 1.0,0.1,0.01 \\
\cline{2-8} \cline{3-8}
 & \multirow[t]{2}{*}{Symmetric} & 20.00\% & 9-Layer CNN & 64 & \lstinline|Adam(lr=0.0001, weight_decay=0.01)| & 92 & 0,50,75 / 1.0,0.1,0.01 \\
\cline{3-8}
 &  & 50.00\% & 9-Layer CNN & 64 & \lstinline|Adam(lr=5e-05, weight_decay=0.005)| & 62 & 0,50 / 1.0,0.1 \\
\cline{1-8} \cline{2-8} \cline{3-8}
\shortstack{Cardioto-\\cography} & Symmetric & 20.00\% & MLP & 1024 & \lstinline|Adam(lr=0.01, weight_decay=0.0)| & 206 & --- \\
\cline{1-8} \cline{2-8} \cline{3-8}
Clothing1M & Real & 38.26\% & ResNet-18 & 64 & \lstinline|Adam(lr=0.001, weight_decay=5e-05)| & $189^*$ & 0,100 / 1.0,0.01 \\
\cline{1-8} \cline{2-8} \cline{3-8}
\shortstack{Credit\\Fraud} & Symmetric & 20.00\% & MLP & 1024 & \lstinline|Adam(lr=0.001, weight_decay=0.001)| & 749 & --- \\
\cline{1-8} \cline{2-8} \cline{3-8}
Food-101N & Real & 18.51\% & ResNet-34 & 64 & \lstinline|SGD(lr=0.04, weight_decay=0.0005, momentum=0.0)| & $248^*$ & 0,200 / 1.0,0.1 \\
\cline{1-8} \cline{2-8} \cline{3-8}
\shortstack{Human\\Activity} & Symmetric & 20.00\% & MLP & 1024 & \lstinline|Adam(lr=0.001, weight_decay=0.001)| & 797 & --- \\
\cline{1-8} \cline{2-8} \cline{3-8}
Letter & Symmetric & 20.00\% & MLP & 1024 & \lstinline|Adam(lr=0.001, weight_decay=0.0)| & 785 & --- \\
\cline{1-8} \cline{2-8} \cline{3-8}
Mushroom & Symmetric & 20.00\% & MLP & 1024 & \lstinline|Adam(lr=0.0001, weight_decay=0.001)| & 344 & --- \\
\cline{1-8} \cline{2-8} \cline{3-8}
Satellite & Symmetric & 20.00\% & MLP & 1024 & \lstinline|Adam(lr=0.001, weight_decay=0.001)| & 722 & --- \\
\cline{1-8} \cline{2-8} \cline{3-8}
\shortstack{Sensorless\\Drive} & Symmetric & 20.00\% & MLP & 1024 & \lstinline|Adam(lr=0.001, weight_decay=0.001)| & 797 & --- \\
\cline{1-8} \cline{2-8} \cline{3-8}
\bottomrule
\end{tabular}

%% file: Tables/classification_performances.tex
\begin{tabular}{llclc}
    \toprule
    \textbf{Dataset} & \textbf{Noise Type} & {\textbf{Noise Rate (\%)}} & \textbf{Model Architecture} & {\textbf{Accuracy (\%)}} \\
    \midrule
    
    CIFAR-10                 & Pairflip    & 20.00 & 9-Layer CNN             & 90.02 \\
                            \cmidrule{2-5}
                            & Symmetric   & 20.00 & 9-Layer CNN             & 89.21 \\
                            \cmidrule{2-5}
                            & Symmetric   & 50.00 & 9-Layer CNN             & 81.77 \\
    \midrule

    CIFAR-10N                & Aggregate       & 9.01  & 9-Layer CNN             & 90.33 \\
                            \cmidrule{2-5}
                            & Worse       & 40.21 & 9-Layer CNN             & 81.57 \\
    \midrule
    
    CIFAR-100                  & Pairflip    & 20.00 & 9-Layer CNN             & 68.78 \\
                            \cmidrule{2-5}
                            & Symmetric   & 20.00 & 9-Layer CNN             & 63.72 \\
                            \cmidrule{2-5}
                            & Symmetric   & 50.00 & 9-Layer CNN             & 56.07 \\
    \midrule
    
    
    Clothing1M                & Real        & 38.26 & ResNet-18           & 68.73 \\
    \midrule
    Food-101N                 & Real        & 18.51 & ResNet-34           & 77.92 \\
    \midrule
    Cardiotocography          & Symmetric   & 20.00 & MLP    & 78.59 \\
    \midrule
    Credit Fraud               & Symmetric   & 20.00 & MLP    & 94.58 \\
    \midrule
    Human Activity  & Symmetric   & 20.00 & MLP    & 76.62 \\
    \midrule
    Letter         & Symmetric   & 20.00 & MLP    & 90.58 \\
    \midrule
    Mushrooms                 & Symmetric   & 20.00 & MLP    & 99.19 \\
    \midrule
    Satellite                 & Symmetric   & 20.00 & MLP    & 88.25 \\
    \midrule
    Sensorless Drive  & Symmetric   & 20.00 & MLP    & 98.55 \\
    
    \bottomrule
\end{tabular}

%% file: Appendices/detection_hyperparams.tex
Table~\ref{tab:hyperparameters_detection} provides the hyperparameters for each method that achieved best performance on a tuning set of varying size. The table on the left provides the parameters for vision datasets, and the table on the right for tabular datasets.

\begin{table}[H]
    \centering
    \begin{minipage}[t]{0.45\textwidth}
        \begin{adjustbox}{angle=90, max width=\textwidth, max height=0.85\textheight, keepaspectratio}
        \input{Tables/detection_hyperparameters_vision}
        \end{adjustbox}
    \end{minipage}
    \hfill
    \begin{minipage}[t]{0.45\textwidth}
        \begin{adjustbox}{angle=90, max width=\textwidth, max height=0.85\textheight, keepaspectratio}
        \input{Tables/detection_hyperparameters_tabular}
        \end{adjustbox}
    \end{minipage}
    \caption{Optimized hyperparameters for each detection method on varying fractions of labeled samples, 0\% size means the default parameters used in our main results. For the Last method the value indicates the single epoch used for calculating the scores. For Mean-CE and MeanProb-LM, it indicates the range of epochs used for calculating scores. For CTRL-CE it provides the parameters for the number of windows, the total number of clusters used in K-means for each window and the number of selected clusters, respectively.}
    \label{tab:hyperparameters_detection}
\end{table}

%% file: Tables/detection_hyperparameters_vision.tex
\begin{tabular}{llcccccccccc}

\toprule

 & \textbf{Model} & \multicolumn{8}{c}{\textbf{9-Layer CNN}} & \textbf{ResNet-18} & \textbf{ResNet-34} \\
\cmidrule(lr){2-2} \cmidrule(lr){3-10} \cmidrule(lr){11-11} \cmidrule(lr){12-12} 
 & \textbf{Dataset} & \multicolumn{5}{c}{\textbf{CIFAR-10}} & \multicolumn{3}{c}{\textbf{CIFAR-100}} & \textbf{Clothing1M} & \textbf{Food-101N} \\
\cmidrule(lr){2-2} \cmidrule(lr){3-7} \cmidrule(lr){8-10} \cmidrule(lr){11-11} \cmidrule(lr){12-12} 
 & \textbf{Noise type} & \textbf{Aggregate} & \textbf{Worse} & \textbf{Pairflip} & \multicolumn{2}{c}{\textbf{Symmetric}} & \textbf{Pairflip} & \multicolumn{2}{c}{\textbf{Symmetric}} & \textbf{Real} & \textbf{Real} \\
\cmidrule(lr){2-2} \cmidrule(lr){3-3} \cmidrule(lr){4-4} \cmidrule(lr){5-5} \cmidrule(lr){6-7} \cmidrule(lr){8-8} \cmidrule(lr){9-10} \cmidrule(lr){11-11} \cmidrule(lr){12-12} 
 & \textbf{Noise rate} & 9.01\% & 40.21\% & 20.00\% & 20.00\% & 50.00\% & 20.00\% & 20.00\% & 50.00\% & 38.26\% & 18.51\% \\
\cmidrule(lr){2-2} \cmidrule(lr){3-3} \cmidrule(lr){4-4} \cmidrule(lr){5-5} \cmidrule(lr){6-6} \cmidrule(lr){7-7} \cmidrule(lr){8-8} \cmidrule(lr){9-9} \cmidrule(lr){10-10} \cmidrule(lr){11-11} \cmidrule(lr){12-12} 
Method & Fraction &  &  &  &  &  &  &  &  &  &  \\
\midrule
\multirow[c]{4}{*}{Last-CE/JS} & 0\% & 55 & 45 & 58 & 56 & 52 & 149 & 115 & 56 & 160 & 213 \\
 & 1\% & 91 & 50 & 60 & 88 & 52 & 145 & 91 & 62 & 125 & 299 \\
 & 5\% & 55 & 33 & 60 & 62 & 54 & 133 & 130 & 56 & 201 & 250 \\
 & 10\% & 60 & 36 & 60 & 60 & 54 & 127 & 145 & 56 & 210 & 268 \\
\multirow[c]{4}{*}{Mean-CE} & 0\% & 50-96 & 30-67 & 52-90 & 54-88 & 28-70 & 124-149 & 15-142 & 38-76 & 183-196 & 152-292 \\
 & 1\% & 45-60 & 42-52 & 48-72 & 88-92 & 4-84 & 88-112 & 85-88 & 30-82 & 107-116 & 85-109 \\
 & 5\% & 55-71 & 39-50 & 56-80 & 56-66 & 18-99 & 91-136 & 63-100 & 50-78 & 8-165 & 201-286 \\
 & 10\% & 45-96 & 30-69 & 52-74 & 66-96 & 28-90 & 112-149 & 36-118 & 42-82 & 116-129 & 250-286 \\
\multirow[c]{4}{*}{MeanProb-LM} & 0\% & 35-111 & 14-61 & 36-74 & 40-96 & 20-82 & 97-149 & 15-103 & 38-72 & 13-116 & 61-286 \\
 & 1\% & 50-60 & 23-29 & 42-66 & 38-84 & 0-76 & 6-85 & 106-118 & 56-70 & 13-53 & 250-256 \\
 & 5\% & 35-86 & 7-67 & 30-84 & 44-88 & 6-92 & 103-145 & 57-79 & 34-86 & 17-84 & 6-280 \\
 & 10\% & 35-86 & 18-47 & 44-64 & 40-92 & 24-72 & 100-149 & 9-85 & 38-64 & 8-116 & 250-292 \\
\multirow[c]{4}{*}{CTRL-CE} & 0\% & (4, 2, 1) & (2, 3, 2) & (4, 2, 1) & (4, 3, 1) & (2, 2, 1) & (4, 2, 1) & (4, 3, 1) & (4, 3, 1) & (4, 3, 2) & (2, 3, 1) \\
 & 1\% & (4, 2, 1) & (4, 3, 2) & (4, 2, 1) & (4, 3, 1) & (2, 2, 1) & (4, 2, 1) & (1, 3, 1) & (4, 3, 1) & (4, 3, 2) & (4, 3, 1) \\
 & 5\% & (4, 2, 1) & (4, 3, 2) & (4, 2, 1) & (2, 3, 1) & (2, 2, 1) & (4, 2, 1) & (4, 3, 1) & (4, 3, 1) & (4, 3, 2) & (4, 3, 1) \\
 & 10\% & (4, 2, 1) & (2, 3, 2) & (4, 2, 1) & (2, 3, 1) & (2, 2, 1) & (4, 2, 1) & (4, 3, 1) & (4, 3, 1) & (4, 3, 2) & (2, 3, 1) \\
\bottomrule
\end{tabular}

%% file: Tables/detection_hyperparameters_tabular.tex
\begin{tabular}{llccccccc}

\toprule

 & \textbf{Model} & \multicolumn{7}{c}{\textbf{MLP}} \\
\cmidrule(lr){2-2} \cmidrule(lr){3-9} 
 & \textbf{Dataset} & \textbf{\shortstack{Cardioto-\\cography}} & \textbf{\shortstack{Credit\\Fraud}} & \textbf{\shortstack{Human\\Activity}} & \textbf{Letter} & \textbf{Mushroom} & \textbf{Satellite} & \textbf{\shortstack{Sensorless\\Drive}} \\
\cmidrule(lr){2-2} \cmidrule(lr){3-3} \cmidrule(lr){4-4} \cmidrule(lr){5-5} \cmidrule(lr){6-6} \cmidrule(lr){7-7} \cmidrule(lr){8-8} \cmidrule(lr){9-9} 
 & \textbf{Noise type} & \textbf{Symmetric} & \textbf{Symmetric} & \textbf{Symmetric} & \textbf{Symmetric} & \textbf{Symmetric} & \textbf{Symmetric} & \textbf{Symmetric} \\
\cmidrule(lr){2-2} \cmidrule(lr){3-3} \cmidrule(lr){4-4} \cmidrule(lr){5-5} \cmidrule(lr){6-6} \cmidrule(lr){7-7} \cmidrule(lr){8-8} \cmidrule(lr){9-9} 
 & \textbf{Noise rate} & 20.00\% & 20.00\% & 20.00\% & 20.00\% & 20.00\% & 20.00\% & 20.00\% \\
\cmidrule(lr){2-2} \cmidrule(lr){3-3} \cmidrule(lr){4-4} \cmidrule(lr){5-5} \cmidrule(lr){6-6} \cmidrule(lr){7-7} \cmidrule(lr){8-8} \cmidrule(lr){9-9} 
Method & Fraction &  &  &  &  &  &  &  \\
\midrule
\multirow[c]{4}{*}{Last-CE/JS} & 0\% & 48 & 213 & 130 & 192 & 146 & 500 & 1495 \\
 & 1\% & 81 & 61 & 325 & 64 & 16 & 559 & 227 \\
 & 5\% & 48 & 61 & 32 & 160 & 81 & 500 & 877 \\
 & 10\% & 130 & 366 & 130 & 224 & 81 & 1236 & 1007 \\
\multirow[c]{4}{*}{Mean-CE} & 0\% & 48-65 & 91-336 & 0-325 & 0-416 & 97-195 & 0-1001 & 422-650 \\
 & 1\% & 65-81 & 0-61 & 0-487 & 0-160 & 0-16 & 58-117 & 0-390 \\
 & 5\% & 32-48 & 0-91 & 0-65 & 32-288 & 0-211 & 1030-1060 & 845-1430 \\
 & 10\% & 65-97 & 61-336 & 0-162 & 0-512 & 0-130 & 1236-1295 & 1105-1137 \\
\multirow[c]{4}{*}{MeanProb-LM} & 0\% & 16-97 & 0-397 & 97-227 & 0-384 & 81-211 & 500-559 & 1430-1462 \\
 & 1\% & 65-81 & 0-91 & 97-422 & 64-128 & 0-16 & 0-58 & 0-260 \\
 & 5\% & 0-114 & 0-91 & 65-162 & 0-256 & 0-179 & 677-706 & 845-910 \\
 & 10\% & 32-195 & 61-336 & 0-130 & 0-448 & 0-146 & 1266-1384 & 1202-1235 \\
\multirow[c]{4}{*}{CTRL-CE} & 0\% & (4, 2, 1) & (4, 3, 1) & (4, 2, 1) & (2, 2, 1) & (4, 2, 1) & (2, 2, 1) & (4, 2, 1) \\
 & 1\% & (1, 2, 1) & (1, 2, 1) & (2, 2, 1) & (2, 2, 1) & (1, 2, 1) & (4, 2, 1) & (1, 2, 1) \\
 & 5\% & (1, 2, 1) & (1, 3, 1) & (4, 2, 1) & (2, 2, 1) & (1, 2, 1) & (4, 2, 1) & (4, 2, 1) \\
 & 10\% & (2, 3, 1) & (4, 2, 1) & (4, 2, 1) & (2, 2, 1) & (1, 2, 1) & (1, 2, 1) & (4, 2, 1) \\
\bottomrule
\end{tabular}